\documentclass[11pt,a4paper]{article}

\usepackage[utf8]{inputenc}
\usepackage[T1]{fontenc}
\usepackage{arxiv}
\usepackage{amsmath,amssymb}
\usepackage{booktabs}
\usepackage{array}
\usepackage{tabularx}
\usepackage{tikz}
\usetikzlibrary{positioning,arrows.meta,shapes.geometric,fit,calc,backgrounds}
\usepackage{graphicx}
\usepackage{caption}
\usepackage[numbers,sort&compress]{natbib}
\usepackage{url}
\usepackage{nicefrac}
\usepackage{microtype}
\usepackage{doi}
\usepackage{hyperref}
\usepackage{cleveref}
\usepackage{xcolor}
\hypersetup{
  colorlinks=true,
  citecolor=black, linkcolor=black, urlcolor=black,
  pdftitle={When Style Similarity Scores Fail: Diagnosing Raw CSD Cosine in Artist-Style Evaluation},
  pdfauthor={J\"org Frochte},
  pdfsubject={cs.LG, cs.CV},
  pdfkeywords={text-to-image style evaluation, contrastive style descriptor, CSD, discrimination gap, CSLS, CSD+, zero-training readout, style similarity, pretrained vision encoders, hubness, validity diagnostic}
}

\newcommand{\gap}{\ensuremath{g_k}}
\newcommand{\wmed}{\ensuremath{w_k}}
\newcommand{\cmed}{\ensuremath{c_k}}

\title{When Style Similarity Scores Fail:\\
       Diagnosing Raw CSD Cosine\\
       in Artist-Style Evaluation}
\renewcommand{\shorttitle}{When Style Similarity Scores Fail}

\author{%
  \href{https://orcid.org/0000-0002-5908-5649}{%
    \includegraphics[scale=0.06]{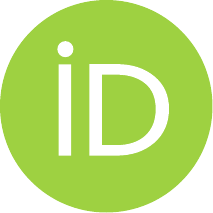}\hspace{1mm}J\"org Frochte}\\
  Bochum University of Applied Sciences\\
  \texttt{joerg.frochte@hs-bochum.de}
}

\date{}

\begin{document}

\maketitle

\begin{abstract}
The cosine similarity of the Contrastive Style Descriptor (CSD) is widely
used as a style score in text-to-image evaluation, and it is usually
averaged over several artists. This presupposes that a high value is
evidence for the style of the artist in question. We introduce the
discrimination gap, a diagnostic that tests a necessary condition for this
on the evaluation corpus itself: the works of an artist have to be more
similar to each other than to the works of any other artist. On an audited
corpus of works in the public domain the condition fails for $20$ of $89$
artists, and two large public corpora show comparable rates at the same
number of artists and works. We separate
three causes. Local density can be corrected by a different readout of the
same embeddings, cross-domain similarity local scaling (CSLS), the influence of the technique can be avoided by the
choice of the reference works, and the overlap between artists of one
tradition remains on all four encoders that we tested. If the readout is
combined with a larger input, for which the positional embeddings of the
encoder are interpolated, the area under the ROC curve of pair verification
rises from $0.891$ to $0.911$. We call the protocol of this readout and the
optional larger input CSD+. It works on the frozen encoder and is not a new
encoder. A
verification test on unseen artists and two tests with generated images
support the diagnosis. In the larger one, which uses four text-to-image
models, the intended artist is ranked first in
$88$ of $95$ cases if the gap is positive and in $1$ of $92$ cases if it is
negative. The diagnostic should therefore be run before cosine values are
averaged over artists.
\end{abstract}

\paragraph{Code.} A reference implementation of the diagnostic and the
CSLS readout is available at
\url{https://github.com/hsbo-ai-ml-dm/csd-plus} (MIT). The analysis
scripts and the result files from which every number in this paper is derived
accompany the submission as supplementary material.

\keywords{text-to-image style evaluation \and contrastive style descriptor (CSD)
\and discrimination gap \and CSLS \and CSD+ \and zero-training readout
\and style similarity \and pretrained vision encoders \and hubness
\and validity diagnostic}

\section{Introduction}
\label{sec:intro}

One of the properties on which text-to-image models are evaluated is how
well they imitate the style of a given artist, school or tradition, or how
reliably they refrain from doing so. Such an evaluation covers thousands of
images and dozens of artists, so it cannot rest on human judgement or on a
description in words. It needs a score that is computed in the same way for
every image, that returns the same value if the computation is repeated,
and that puts images in an order, so that methods can be ranked and the
works closest to a generated image can be retrieved. The same score serves
to ask which artists of the training data a generated image
imitates~\cite{somepalli2024csd}. In
recent work this score is very often the cosine similarity between
embeddings of the Contrastive Style Descriptor (CSD) of Somepalli et
al.~\cite{somepalli2024csd}, a ViT-L that was trained on LAION-Styles with
tags for artists, media and movements.

The evaluations that we review in \S\ref{sec:background} follow a
common scheme. For each target style or artist a small set of reference
images is fixed, which we call the anchor pool. The model generates images
in this style, and the score of a generated image is the mean cosine between
its embedding and the embeddings of the anchors. A function of this kind,
which maps the embeddings of a frozen encoder to a score, is called a
readout in the following. The mean of the raw cosine is the readout of
current practice. The scores are then averaged over the generated images
and over the styles, and two methods are compared by the resulting single
number, often at a difference of two to four percentage points. A difference
of this size is only informative if a higher score means more of the
intended style for every style.

In this paper the styles are those of artists, and this is the case with
which the CSD paper itself begins. Somepalli et al.\ score the images that a
model generates for $96$ artists against the mean embedding of the works of
each artist. They compare the artists by this score, and they suppose that
a value below $0.5$ indicates the absence of the style of an artist and a
value above $0.8$ its presence~\cite{somepalli2024csd}. Such a reading
assumes that a high value for an artist is evidence for the style of this
particular artist, and the CSD paper does not validate this. It
validates the descriptor for retrieval on WikiArt, and there the authors
already observe that artists of the same movement are hard to separate. An
absolute score per image is a stronger
use than retrieval, because retrieval only needs the correct order within
one query.

In this paper we introduce the discrimination gap as a diagnostic for this
assumption. For each artist we compute the median cosine among the anchors
of this artist and the median cosine between these anchors and the anchors
of each other artist. The gap is the difference between the first value and
the largest of the others. If the gap is negative, there is at least one
other artist whose works are, in the median, closer to the pool than the
artist's own works are. In this case a high score of a generated image does
not tell us that the image is close to the style of this particular artist.
The diagnostic needs only the anchor pools that an evaluation uses anyway,
and it needs no threshold and no training. A positive gap is a necessary
condition for this reading, but it is not a sufficient one.

With this diagnostic we ask three questions. At which rate does the reading
as an absolute score fail, and on which properties of a corpus does this
rate depend, so that it can be transferred to other corpora? What are the
causes of a failure, and which of them can be repaired without a new
encoder? And does a failure that is found on authentic works also hold for
generated images, which are what an evaluation finally scores, even for a
generator that was adapted to the artist? The first
two questions are studied on authentic works, because only for them the
artist is known. CSD serves as the example. Since the gap needs only
embeddings and labels, we repeat the analysis on CLIP, SigLIP and DINOv2.

The paper is organised along these questions. \S\ref{sec:background}
reviews how the CSD score is used, and \S\ref{sec:corpus} describes the
three corpora, of which Commons-89 was assembled for this study with $1645$
works in the public domain by $89$ artists. \S\ref{sec:diagnostic}
answers the first question. On Commons-89 the gap is negative for $20$ of
the $89$ artists, and with ten works per artist, the largest
pool among the evaluations we review, the three corpora lie between $19\%$ and $28\%$.
\S\S\ref{sec:density} and~\ref{sec:tradition} separate three causes.
These are the local density, which a readout with Cross-domain Similarity
Local Scaling (CSLS)~\cite{conneau2018word} corrects, the binding of an
artist to a technique, which the choice of the anchors avoids, and the
overlap between artists of one tradition, which none of the corrections we
tested removes. For the density there is a second correction without new
training. The input of the encoder is enlarged, and its positional
embeddings are interpolated to the larger input. We call the protocol that
uses the CSLS readout and, as an option, this larger input CSD+. It is not
a new encoder. \S\ref{sec:verification} tests this separation with pair verification on artists that were not used
for the correction. As the separation predicts, the correction gains most
against the closest stylistic neighbour, and this holds in all twelve
combinations of three corpora and four encoders.
\S\ref{sec:generated} answers the third question with images from four
text-to-image models. For four artists with a positive gap the pool of the
artist has the highest score for $88$ of $95$ images, and for four artists
with a negative gap for $1$ of $92$. For three of the latter the images go
most often to the competitor that the diagnostic names beforehand. A second
test adapts one generator with LoRA to the artists of three pairs. Under the
raw cosine, at least $85\%$ of the images for the artist with the negative
gap are still closer to the pool of the competitor than to the artist's own
pool. \S\ref{sec:discussion} turns the
results into a procedure for evaluations, and \S\ref{sec:conclusion}
returns to the three questions.

\section{Background}
\label{sec:background}

\begin{table}[t]
\centering
\caption{How recent work uses CSD. The entries are taken from the cited
papers; n.s.\ means that the paper does not state the value.}
\label{tab:audit}
\begin{tabular}{lllrr}
\toprule
Work & Role of CSD & Comparison & Anchors & Images \\
\midrule
Magic Insert~\cite{sohn2024magicinsert} & metric & across styles & n.s. & n.s. \\
UnZipLoRA~\cite{castano2024unziplora} & metric & across styles & 1 & 16 \\
Staniszewski et al.~\cite{wodynski2024lowrank} & metric & per style, averaged & 10 & 40 \\
InstantStyle-Plus~\cite{wang2024instantstyleplus} & sampler & -- & -- & -- \\
RB-Modulation~\cite{rout2025rbmodulation} & sampler & -- & -- & -- \\
\bottomrule
\end{tabular}
\end{table}

Neural
style transfer~\cite{gatys2016style} compares Gram matrices of VGG
features, and the evaluation of diffusion models later moved to large
pretrained encoders such as CLIP~\cite{radford2021clip} and
DINOv2~\cite{oquab2024dinov2}. Neither of these was trained with style
labels. CSD~\cite{somepalli2024csd} is a ViT-L that was trained on
LAION-Styles, a set of about half a million images with tags for artists,
media and movements. Somepalli et al.\ evaluate it by
retrieval on WikiArt, and they use it to study how far diffusion models
reproduce the style of their training images.

Later text-to-image work uses the CSD cosine in two different ways. In the
first one it is a guidance signal inside the sampler, as in
InstantStyle-Plus~\cite{wang2024instantstyleplus} and
RB-Modulation~\cite{rout2025rbmodulation}. This use is not our subject. In
the second one it is a column in an evaluation table, next to CLIP and
DINO scores. Magic Insert~\cite{sohn2024magicinsert} reports it as one of
its style-fidelity scores, UnZipLoRA~\cite{castano2024unziplora} uses it to
measure the fidelity of style and subject, and Staniszewski et
al.~\cite{wodynski2024lowrank} use it to measure forgetting in continual
personalisation. We went through these papers to see how the number is
actually read (Tab.~\ref{tab:audit}). Unlike the CSD paper with its two thresholds, none of the three interprets
an absolute threshold, so in all
three the score serves to rank methods. However, all three reduce it to one
number per method over several styles or artists, and such an average is
read as fidelity to the targets only if a higher score means more of the
intended style for each of them. Where
the number of anchors is stated, it is one and ten, so the pools are
small. A comparison of two
models on the same artist with the same anchors and prompts is a different
inference, and our diagnostic does not test it.

The effect that we will see in \S\ref{sec:diagnostic} is related to a
known property of high-dimensional spaces. Radovanovi\'c et
al.~\cite{radovanovic2010hubs} show that in such spaces a few vectors
of a data set appear among the nearest neighbours of very many other
vectors. These vectors are called hubs. CSLS~\cite{conneau2018word} was introduced as a correction for hubs in
the translation of word embeddings. It subtracts from each cosine the mean
cosine of both vectors to their nearest neighbours. In
\S\ref{sec:density} we carry this construction over to the scoring of
an image against the anchor pool of an artist.

\section{Corpora and Embeddings}
\label{sec:corpus}

We use three corpora (Tab.~\ref{tab:corpora}). The first one was built for
this study, because the controls in \S\S\ref{sec:diagnostic}
and~\ref{sec:tradition} need to know for every single work who made it and
in which technique. We call it Commons-89 after its source and its number
of artists. The artist set starts from the list of $400$ artist names
released with the CSD code. After merging different spellings of the same
name, $367$ artists remain. This list is dominated by the Western canon.
Therefore we added fourteen artists from Japan, China, Korea, Persia, India
and Mexico, and a few artists to fill gaps in single traditions. The added artists are
listed in App.~\ref{app:corpus}. For each
artist we fetched works in the public domain from Wikimedia Commons.

The candidates were checked in two stages. In the first one a
language model judges from the description of a file whether the work is by
the artist, and a vision-language model judges from the image whether it
shows the artwork and nothing else. In the second stage every image that
had passed the first one was inspected by a human. In this stage
copies of the same physical work were reduced to one anchor, reproductions
by another hand were labelled as such, and photographs that show a frame or
a museum wall were cropped to the artwork. An artist was kept only if at
least ten works remained after both stages. This leaves $89$ artists with
$1645$ works, with $10$ to $27$ works per artist. The technique of each
work was recorded in this process, and every artist carries the label of a
movement, which we call the tradition of the artist. These labels come
from a vocabulary of $47$ labels that we compiled, and $45$ of them occur in
Commons-89. The details of both stages and the reasons for which files were
rejected are given in App.~\ref{app:corpus}.

The audit also records for every work how its source attributes it. A work
is labelled master if the source assigns it to the hand of the artist, and
otherwise workshop, school, after the artist or attributed. These labels
are used in \S\ref{sec:diagnostic} to repeat the diagnostic on the master
works alone. For a study of style Commons-89 is a small corpus. For this
reason the main results are broken down by tradition and by competitor in
addition to the averages, and the verification in
\S\ref{sec:verification} is repeated on $25$ splits of the artists.

The other two corpora supply the scale that Commons-89 does not have. The
public WikiArt dump~\cite{wikiart} draws on the collection on which CSD
was evaluated, and in ArtBench-10~\cite{liao2022artbench} we count $1639$
artists with at least ten works. We did not audit their images and only removed the works of unknown
artists from the WikiArt dump, and both go through exactly the same code as
Commons-89. They are not independent of each other, because ArtBench
follows the file naming of WikiArt and shares $117$ of the $128$ WikiArt
artists. In addition, we use the version of ArtBench in which the images are
cropped to squares of $256$ pixels. The construction of both corpora and
their overlap at the level of single images are described in
App.~\ref{app:external}. Because of these differences we compare the three corpora only at
the same number of artists and the same number of works per artist.

\begin{table}[t]
\centering
\caption{The three corpora. Works per artist are given as median and
range. The last column states whether the images were audited for this
study.}
\label{tab:corpora}
\begin{tabular}{lrrrl}
\toprule
Corpus & Works & Artists & Works per artist & Audit \\
\midrule
Commons-89  & 1645  & 89   & 19 (10--27)    & per image \\
WikiArt     & 39530 & 128  & 198.5 (42--1889) & none \\
ArtBench-10 & 57800 & 1639 & 24 (10--443)   & none \\
\bottomrule
\end{tabular}
\end{table}

Each image is resized so that its shorter side has $224$ pixels and is then
cropped at the centre to $224 \times 224$ pixels, which is the input of the
public CSD checkpoint. So the encoder sees the whole work at a reduced
resolution, apart from the margins of the longer side. Its output has
$d = 768$ dimensions and is normalised to unit length. This is the pipeline
that is used in the literature, and all results on CSD refer to it with
one exception. The optional larger input of CSD+, which is described in
\S\ref{sec:density}, is used in \S\S\ref{sec:density}
and~\ref{sec:verification}, for the residual in \S\ref{sec:tradition} and
for the LoRA test in \S\ref{sec:generated}. For the
comparison in \S\ref{sec:tradition} the same works are embedded with
CLIP-ViT-L/14~\cite{radford2021clip}, SigLIP-large~\cite{zhai2023siglip}
and DINOv2-Large~\cite{oquab2024dinov2}, each with its own standard
preprocessing.

The following notation is used in the rest of the paper. The encoder maps a
work to a vector $x \in \mathbb{R}^d$ with $\|x\| = 1$, which we call the
embedding of the work. Since all embeddings have unit length, the cosine of
two works is the inner product $x^\top y$ of their embeddings. A corpus has
$K$ artists. For $k = 1, \dots, K$ we write $A_k$ for the set of the
embeddings of all works of artist $k$ in the corpus and call it the pool of
this artist. Its elements are the anchors, and $n_k = |A_k|$ is their
number. The classes are artists throughout. Nothing in the definitions depends on this, however, and a class
could equally be a school or a tradition.

\section{The Discrimination Gap}
\label{sec:diagnostic}

For two artists $k$ and $j$ let
\begin{equation}
  \mu_{kj} = \operatorname{median} \{\, x^\top y \;:\; x \in A_k,\ y \in A_j,\ x \neq y \,\}
\end{equation}
be the median cosine between their pools. For $j = k$ this is the median
over all pairs of different works of artist $k$. From these medians we form
two numbers for each artist $k$, the within value and the cross value,
\begin{equation}
  \wmed = \mu_{kk} , \qquad \cmed = \max_{j \neq k} \mu_{kj} .
\end{equation}
The artist $j$ at which the maximum is reached is called the worst
competitor of $k$. The discrimination gap of artist $k$ is
\begin{equation}
  \gap = \wmed - \cmed .
\end{equation}
If the gap is positive, the works of artist $k$ are in the median more
similar to each other than to the works of any other artist. If it is
negative, there is at least one artist whose works are closer. In this case
the order of the raw cosine is wrong for artist $k$ at the level of the
medians, and a high value cannot be read as evidence for the style of $k$ in particular.
Fig.~\ref{fig:intuition} shows the
two cases.

\begin{figure}[t]
\centering
\begin{tikzpicture}[x=0.9cm, y=0.9cm,
  every node/.style={font=\small},
  axisline/.style={->, >=Latex, line width=0.5pt, gray!70},
  withinden/.style={line width=1.0pt, draw=blue!70!black, fill=blue!18, fill opacity=0.55},
  crossden/.style={line width=1.0pt, draw=orange!75!black, fill=orange!22, fill opacity=0.55, dashed},
  marker/.style={line width=0.5pt, gray!50},
  gaparr/.style={<->, >=Latex, line width=0.7pt, black}]
\begin{scope}[xshift=0cm]
  \draw[axisline] (-0.2,0) -- (5.9,0);
  \node[below, font=\footnotesize] at (5.45,-0.05) {cosine};
  \fill[crossden] plot[domain=0.0:3.0, samples=50] ({\x},{1.5*exp(-(\x-1.5)^2/0.32)}) -- (3.0,0) -- (0.0,0) -- cycle;
  \fill[withinden] plot[domain=2.5:5.5, samples=50] ({\x},{1.5*exp(-(\x-4.0)^2/0.32)}) -- (5.5,0) -- (2.5,0) -- cycle;
  \draw[marker] (1.5,0) -- (1.5,1.5);
  \draw[marker] (4.0,0) -- (4.0,1.5);
  \node[below] at (1.5,-0.05) {$\cmed$};
  \node[below] at (4.0,-0.05) {$\wmed$};
  \node[color=orange!75!black, align=center, font=\footnotesize] at (1.2,2.0) {worst\\ competitor};
  \node[color=blue!70!black, align=center, font=\footnotesize] at (4.3,2.0) {own\\ works};
  \draw[gaparr] (1.5,2.75) -- (4.0,2.75);
  \node[above] at (2.75,2.75) {$\gap > 0$};
  \node at (2.85,-1.05) {(a) positive gap};
\end{scope}
\begin{scope}[xshift=6.5cm]
  \draw[axisline] (-0.2,0) -- (5.9,0);
  \node[below, font=\footnotesize] at (5.45,-0.05) {cosine};
  \fill[crossden] plot[domain=1.7000000000000002:4.7, samples=50] ({\x},{1.5*exp(-(\x-3.2)^2/0.32)}) -- (4.7,0) -- (1.7000000000000002,0) -- cycle;
  \fill[withinden] plot[domain=0.8999999999999999:3.9, samples=50] ({\x},{1.5*exp(-(\x-2.4)^2/0.32)}) -- (3.9,0) -- (0.8999999999999999,0) -- cycle;
  \draw[marker] (3.2,0) -- (3.2,1.5);
  \draw[marker] (2.4,0) -- (2.4,1.5);
  \node[below] at (3.2,-0.05) {$\cmed$};
  \node[below] at (2.4,-0.05) {$\wmed$};
  \node[color=orange!75!black, align=center, font=\footnotesize] at (4.0,2.0) {worst\\ competitor};
  \node[color=blue!70!black, align=center, font=\footnotesize] at (1.6,2.0) {own\\ works};
  \draw[gaparr] (2.4,2.75) -- (3.2,2.75);
  \node[above] at (2.8,2.75) {$\gap < 0$};
  \node at (2.85,-1.05) {(b) negative gap};
\end{scope}
\end{tikzpicture}
\caption{The two cases of the discrimination gap. In (a) the median cosine
among the works of an artist lies above the median cosine to the worst
competitor. In
(b) the two medians are in the wrong order.}
\label{fig:intuition}
\end{figure}

The class of a work is the artist to whom its source attributes it, so the gap
measures how well works of different authors are told apart. It does not
measure whether two works look alike to a person. Two painters of one
school can be very similar in this second sense, and hence a negative gap
is not in itself an error of the metric. It only says that the score cannot
separate these two artists, which is exactly what an evaluation per artist
assumes. A third question is whether a generated image reproduces the
manner of an artist. This is the use in an evaluation, and a diagnostic on
authentic works cannot answer it alone. For this reason
\S\ref{sec:generated} tests generated images separately. Furthermore, the
gap tests the order of two medians. It is not a
test of calibration, and a positive gap does not guarantee that an artist
is well separated.

\begin{figure}[t]
\centering
\includegraphics{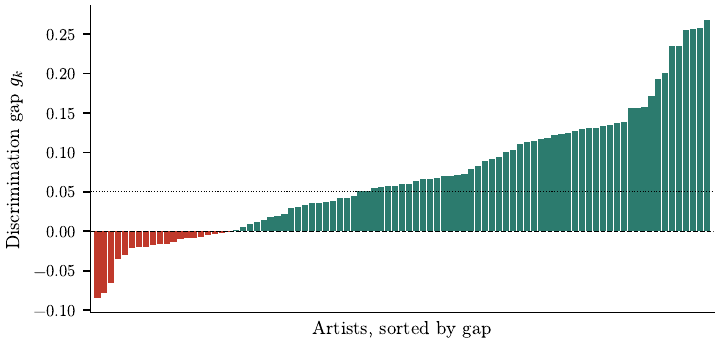}
\caption{Discrimination gap of the $89$ artists of Commons-89 in ascending
order. The bars below the dashed zero line belong to the $20$ artists with
a negative gap, and $38$ artists lie below the dotted line at $0.05$.}
\label{fig:gap-dist}
\end{figure}

On Commons-89 the gap is negative for $20$ of the $89$ artists, which is
$22.5\%$ (Fig.~\ref{fig:gap-dist}). A further $18$ artists have a gap
between zero and $0.05$, and only twelve reach $0.15$ or more. So the margin is narrow also for
many of the artists that are not flagged. The count is
stable if the anchors are resampled. Over $1000$ bootstrap samples it is
$23.7 \pm 2.7$, and it lies above $20$ because a bootstrap sample is in
effect a thinner pool. For single artists the picture is less clear. With
$10$ to $27$ works per artist, the $95\%$ interval of the gap lies
completely below zero for only three of the twenty artists, and the others
should be read as doubtful cases. Not every work in the pools is assigned to the artist without reservation.
Of the $1645$ works, $1482$ are master works, which is $90\%$. The source
marks some of the others as workshop pieces, in which the share of the
artist is open, and some as copies by another hand. We keep these works,
because the pools are meant to be the ones that a practitioner would
assemble from public collections. They explain only a small part of the
count. If only the master works are used, $15$ instead of $18$ of the $80$
artists that remain in both settings are flagged, and the deepest cases are
not affected. App.~\ref{app:controls} gives two further checks of the
attribution.

A negative gap could in principle be a weak effect, in which two medians
are in the wrong order while the two distributions still separate well. To
check this, we computed for each artist the probability that a random pair
of the artist's own works has a higher cosine than a random pair with the
worst competitor. For the $20$ flagged artists this probability is $0.469$
on average and it stays below $0.7$ for every one of them, whereas it is
$0.692$ for the other $69$ artists. So for a flagged artist the separation
from the worst competitor is at the level of chance. For the other artists
it is better, but about three of ten such comparisons still go the wrong
way.

Two further checks show how far a single flag can be trusted. The worst
competitor is selected on the same anchors on which the gap is then
computed, and such a selection can produce negative gaps by chance.
Therefore we selected the competitor on one part of the anchors and
computed the gap on the other part. Eleven of the twenty artists stay
negative in the majority of such splits, and six of the eight most negative artists are
among them. In the same spirit we removed the
worst competitor of an artist from the corpus. For the fifteen most
negative artists this turns six gaps positive, and all six are shallow
cases with a gap of at most $0.02$ in magnitude. Behind each of the deep
cases stands a second competitor. The second check concerns the
median. If it is replaced by the mean, a trimmed mean or a quartile, between
$13$ and $31$ artists are flagged, so the order of magnitude is stable.
However, between a third and nearly half of the flagged artists change with
the statistic. For this reason we regard the rate as the result, and the
pairs that we name below serve as illustrations.

The two public corpora show the same effect. The gap is negative for
$19$ of $128$ artists on WikiArt, which is $14.8\%$, and for $646$ of
$1639$ on ArtBench-10, which is $39.4\%$. Of these, $11$ and $289$ are
negative in at least $95\%$ of the bootstrap samples. The pools of WikiArt hold up to $1889$
works, so small samples cannot be the reason.

At first sight these two rates and the $22.5\%$ of Commons-89 seem to say
that the corpora differ in difficulty. This reading is wrong, and the
reason lies in the definition of $\cmed$. Since $\cmed$ is a maximum over
all competitors, a new artist in the corpus can turn a positive gap into a
negative one, but never the other way round. Hence the count can only grow
with the number of artists, and the rate grows with it as long as the new
artists are not all well separated. One can observe this inside a single
corpus by restricting the maximum to random subsets of the artists. On
WikiArt the rate rises from $4.7\%$ for $32$ artists to $14.8\%$ for all
$128$, and ArtBench has a rate of $11.3\%$ at $89$ artists against $39.4\%$
at $1639$. The number of works per artist has a second and independent
influence, because both terms of the gap are medians over few pairs and
$\cmed$ is the maximum of many such noisy medians.
Tab.~\ref{tab:inventory} holds the number of artists at $89$ and varies
the number of works. With ten works per artist the three corpora lie at
$25.4\%$, $27.5\%$ and $19.4\%$, and with all works WikiArt and ArtBench fall
to $11.9\%$ and $11.1\%$.

\begin{table}[t]
\centering
\caption{Rate of negative gaps for $89$ artists and a limited
number of works per artist, averaged over $60$ random draws of artists and
works. The second column gives the median number of works per artist in
the corpus. A dash means that the corpus does not hold enough works.}
\label{tab:inventory}
\label{tab:inventory-controls}
\setlength{\tabcolsep}{3.5pt}
\begin{tabular}{lrrrrrrrrrrr}
\toprule
& & \multicolumn{10}{c}{Works per artist} \\
\cmidrule(lr){3-12}
Corpus & Median & 10 & 12 & 15 & 19 & 24 & 32 & 48 & 96 & 192 & all \\
\midrule
Commons-89  & 19  & 25.4\% & 23.1\% & 22.2\% & -- & -- & -- & -- & -- & -- & 22.5\% \\
WikiArt     & 198 & 27.5\% & -- & -- & 19.3\% & 15.9\% & 14.7\% & 12.9\% & 12.0\% & 11.2\% & 11.9\% \\
ArtBench-10 & 24  & 19.4\% & -- & -- & 13.6\% & 12.3\% & -- & -- & -- & -- & 11.1\% \\
\bottomrule
\end{tabular}
\end{table}

This has two consequences. First, a count can only grow if further
artists are added. So a count that is reported for a corpus is a lower
bound for every larger corpus that contains it, and a rate is a statement about a
measurement setting and only in the second place about a corpus. It can
only be interpreted together with the number of artists and the number of
anchors per artist, and a rate from one paper cannot be transferred to the
corpus of another. Second, the evaluations of Tab.~\ref{tab:audit} that state the
size of their pools use one and ten anchors. This is the end of
Tab.~\ref{tab:inventory} with the highest rates. Already for pools of ten
anchors one has to expect a negative gap for about a fifth to a quarter of
the artists, and we could not measure smaller pools.

\begin{table}[t]
\centering
\caption{The fifteen artists with the most negative gap on Commons-89; $n$
is the number of works.}
\label{tab:worst-gap}
\begin{tabular}{lrrrlr}
\toprule
Artist & $n$ & $\wmed$ & $\cmed$ & Worst competitor & $\gap$ \\
\midrule
Kawanabe Ky\=osai & 20 & 0.449 & 0.534 & T\=osh\=usai Sharaku & $-0.085$ \\
Hasegawa T\=ohaku & 12 & 0.420 & 0.499 & T\=osh\=usai Sharaku & $-0.078$ \\
Gustave Dor\'e & 23 & 0.263 & 0.330 & Francisco Goya & $-0.066$ \\
Ivan Bilibin & 21 & 0.250 & 0.284 & Jos\'e Guadalupe Posada & $-0.035$ \\
Isaac Levitan & 20 & 0.289 & 0.319 & Ivan Shishkin & $-0.030$ \\
Albert Aublet & 16 & 0.225 & 0.247 & Francisco Goya & $-0.022$ \\
Frederic Leighton & 17 & 0.293 & 0.313 & John William Godward & $-0.020$ \\
Gustav Klimt & 18 & 0.283 & 0.302 & Odilon Redon & $-0.020$ \\
Diego Vel\'azquez & 15 & 0.422 & 0.440 & Georges de La Tour & $-0.018$ \\
Eug\`ene Delacroix & 15 & 0.433 & 0.449 & Francisco Goya & $-0.017$ \\
Albert Bierstadt & 24 & 0.401 & 0.417 & George Inness & $-0.016$ \\
John Singer Sargent & 18 & 0.270 & 0.284 & Edward Hopper & $-0.014$ \\
Raffaello Sanzio & 15 & 0.455 & 0.465 & Andrea Mantegna & $-0.009$ \\
Gaston Bussiere & 21 & 0.266 & 0.274 & John William Waterhouse & $-0.009$ \\
Francisco de Zurbar\'an & 13 & 0.419 & 0.427 & Georges de La Tour & $-0.008$ \\
\bottomrule
\end{tabular}
\end{table}

\begin{figure}[t]
\centering
\includegraphics[height=36.4mm]{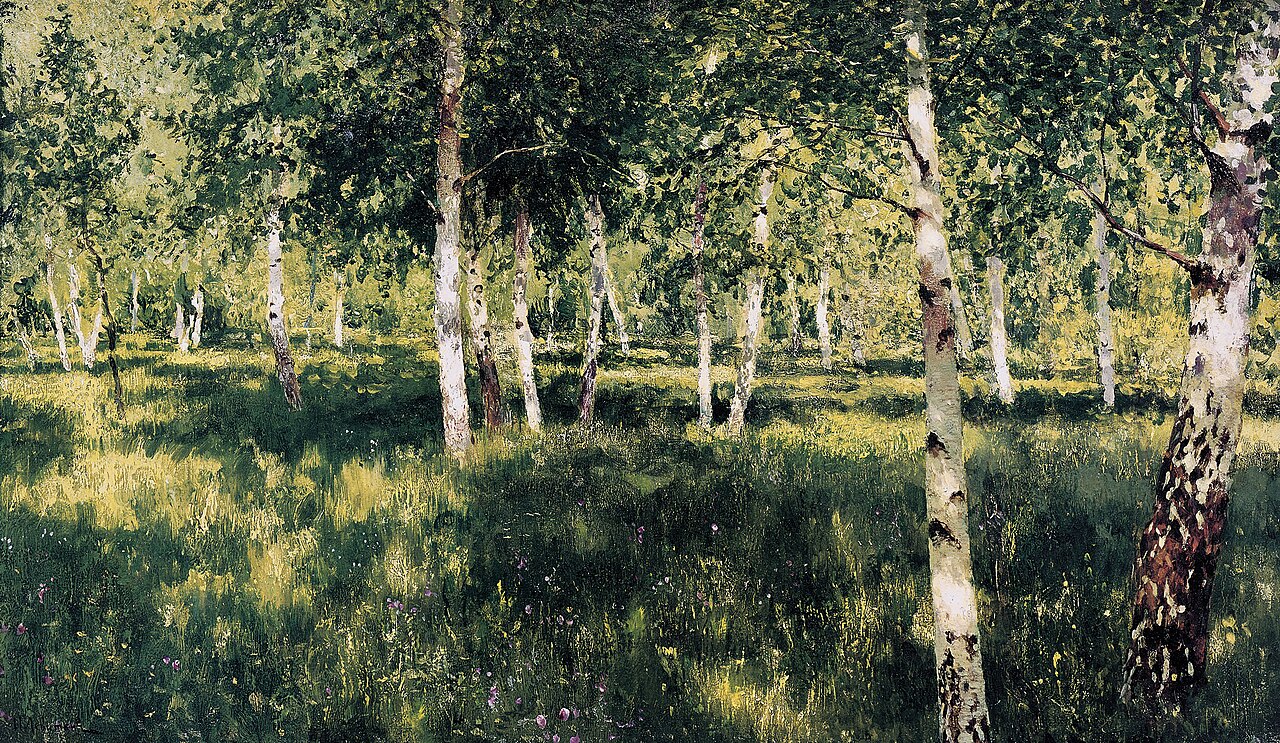}\hspace{4mm}
\includegraphics[height=36.4mm]{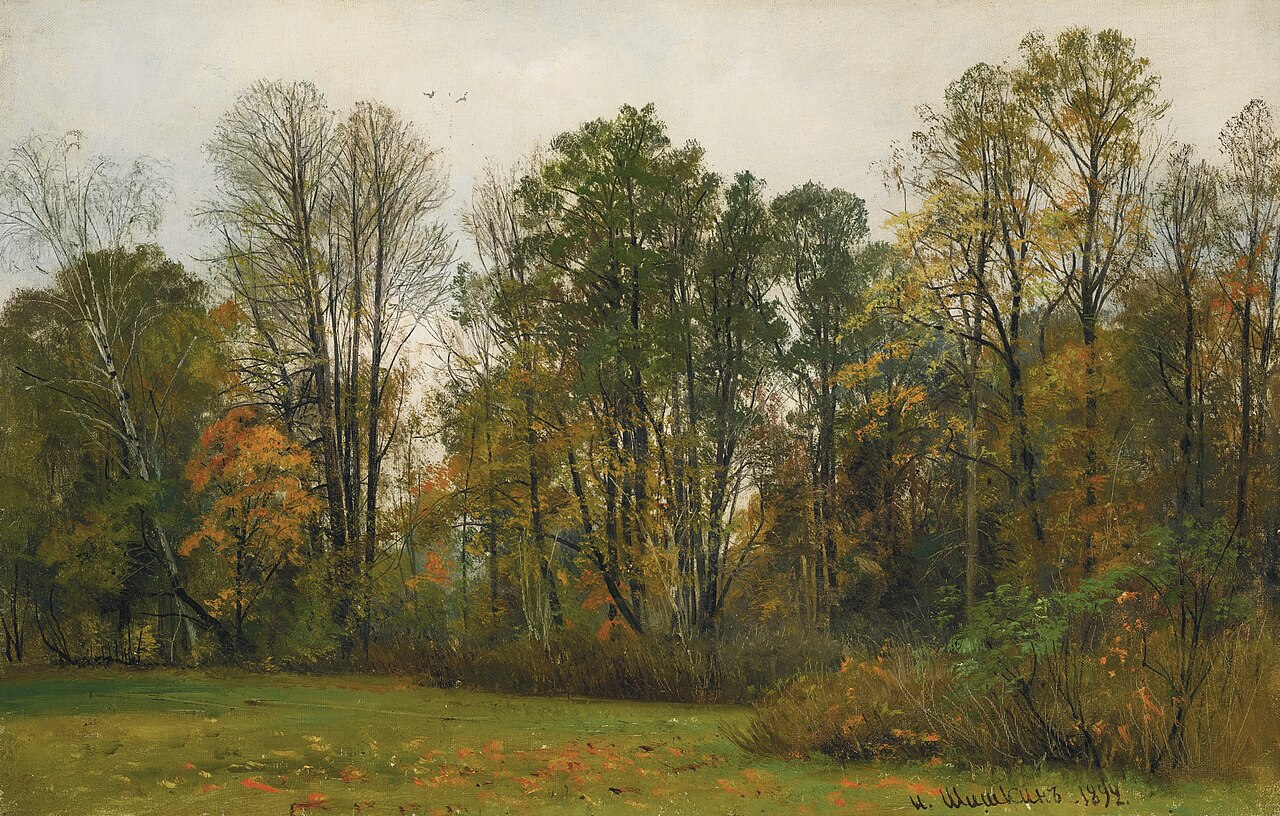}\\[2mm]
\includegraphics[height=40.2mm]{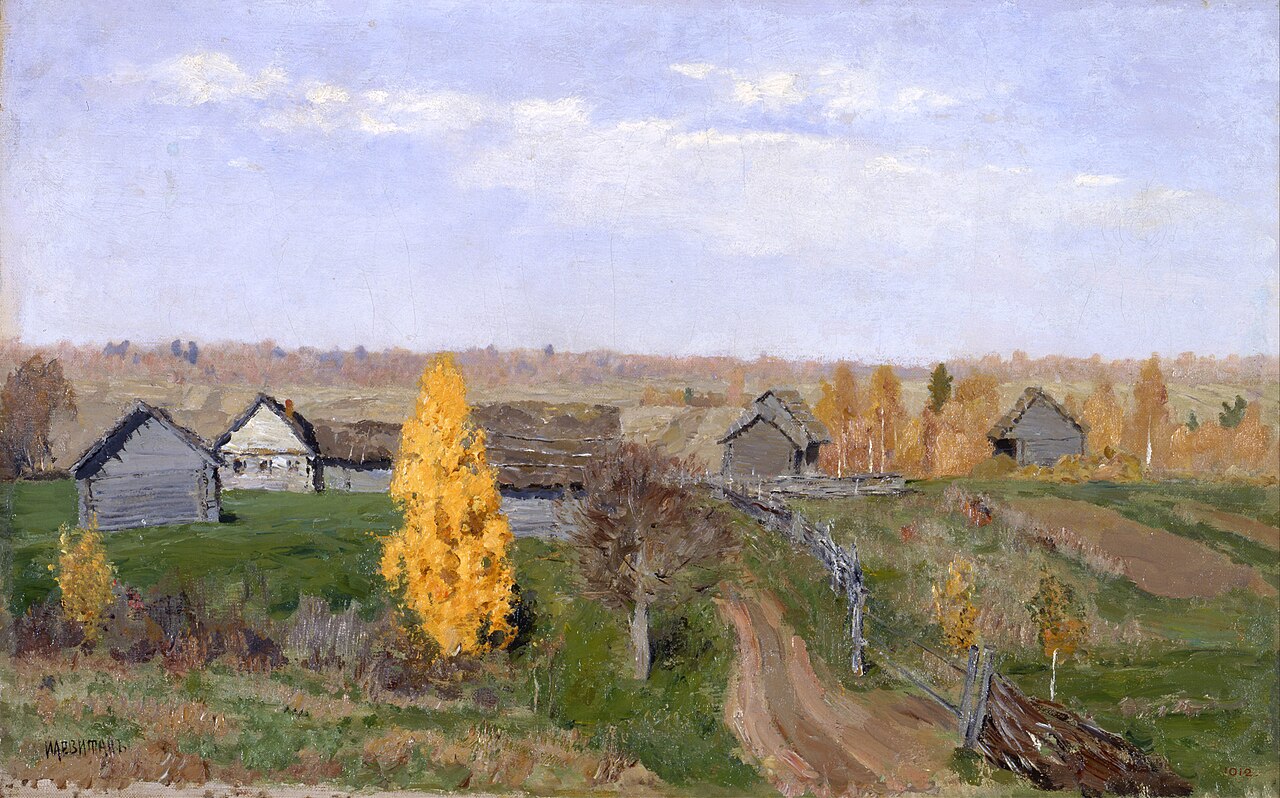}\hspace{4mm}
\includegraphics[height=40.2mm]{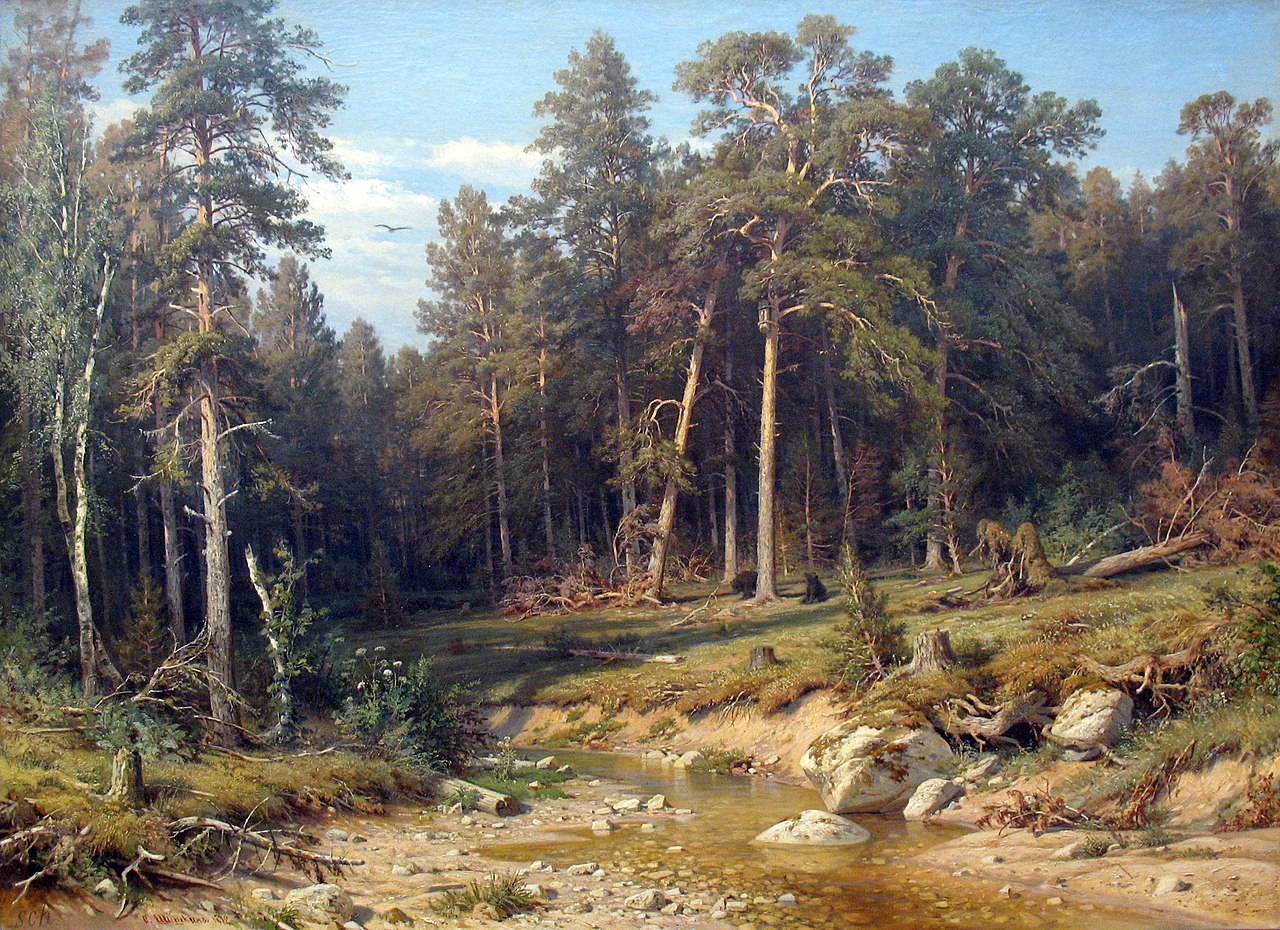}
\caption{Two landscapes by Isaac Levitan (left column) and two by Ivan
Shishkin (right column). In CSD the median cosine between the works of the
two painters is $0.319$, which is higher than the median of $0.289$ among
Levitan's own works. The works of Levitan are \emph{Birch Grove}
(1885--1889) and \emph{Golden Autumn. Slobodka} (1889), those of Shishkin
are \emph{Autumn} (1892) and \emph{A Pine Forest. Mast-Timber Forest in
Viatka Province} (1872). Images: public domain, via Wikimedia Commons.}
\label{fig:exemplars}
\end{figure}

Tab.~\ref{tab:worst-gap} lists the fifteen most negative cases on
Commons-89. In many
rows the worst competitor comes from the same tradition as the artist.
Levitan and Shishkin are both Russian landscape painters of the nineteenth
century (Fig.~\ref{fig:exemplars}). They share subjects such as birch and
pine forests, a low horizon and a deep view into the landscape. Levitan
uses softer colour and a looser handling, whereas Shishkin paints with
botanical precision. The median cosine does not reflect this difference. Leighton and Godward both belong to Victorian academic painting,
and Ky\=osai, T\=ohaku and Sharaku stand for Japanese traditions. Since such a reading is easily made after the fact, we compared the pairs
with PainterPalette~\cite{painterpalette}, a collection of painter metadata
that we did not write. It covers $16$ of the $20$ pairs, and for $10$ of
them it records a common movement or a documented influence. So an independent
source supports the reading for most of the covered pairs. In other rows,
for example Dor\'e and Bilibin, the competitor is a printmaker. This
points to the technique and not to a tradition, and we will come back to it
in \S\ref{sec:tradition}. The pattern is also not present everywhere.
Ten traditions are represented by at least three artists in Commons-89.
Seven of them contain an artist whose closest competitor belongs to the
same tradition, and in four of them the gap of such an artist is negative. So a common
tradition makes two artists close in most cases, but it leads to a negative
gap only in some of them. On the public corpora the worst competitors are neighbours
in art history as well, for example Monet and Loiseau, Braque and Gris,
Rubens and Jordaens, or Hokusai and Kuniyoshi.

The role of the tradition matters when an evaluation pool is assembled. An
artist who
is the only representative of a tradition cannot have a competitor from
this tradition. To measure this, we drew $128$ artists from ArtBench-10 and
varied only the number of style classes from which they come. The rate
falls from $25.9\%$ if all artists belong to one class to $13.6\%$ if they
are spread over ten classes, and the difference does not become smaller if
every artist is limited to $19$ works. So the rate follows the density with
which the traditions are sampled, and a corpus with one artist per
tradition understates the problem by about a factor of two.

One artist plays a particular role in Tab.~\ref{tab:worst-gap}. Francisco
Goya is the worst competitor of seven of the $89$ artists, a count that
Mantegna and Inness also reach, whereas about
one would be expected if the competitors were spread evenly. His works are
among the most compact pools of the corpus, with $\wmed = 0.745$, and they
are close to many pools whose own works are more scattered. Unlike the
hubs of \S\ref{sec:background}, he does not appear unusually often
among the nearest neighbours of single images, so an analysis of hubs at
the level of images would not have found him.

So the raw cosine has the wrong order for a considerable part of the
artists on all three corpora. The diagnostic says that the absolute reading
is unsafe for such an artist, but it does not say why. The examples of this section point to three different
causes, namely a compact pool as in the case of Goya, a difference in
technique as in the case of Dor\'e, and a common tradition as in the case
of Levitan and Shishkin. In the following two sections we separate them. The first one is a property of the readout and can be corrected
there. The other two are properties of the representation.

\section{Local Density and the Readout}
\label{sec:density}

The gap of \S\ref{sec:diagnostic} is defined on single pairs. An
evaluation, however, does not score pairs. It scores an image $x$ against
a whole pool $A$ by the mean cosine
\begin{equation}
  s(x, A) = \frac{1}{|A|} \sum_{a \in A} x^\top a .
\end{equation}
Therefore we repeat the diagnostic for this pooled score. The within value is now the median of
$s(x, A_k \setminus \{x\})$ over the works $x$ of artist $k$, so that a
work is never compared with itself. For each other artist $j$ we take the
median of $s(x, A_k)$ over the works $x$ of $j$, and the largest of these
medians is the cross value. The pooled gap is again the difference of the
two. On Commons-89 it is negative for $12$ of the $89$ artists. This is
less than the $20$ of the pairwise gap, because the mean over a pool
removes part of the variance of single pairs, but the effect itself
remains.

The embeddings of some artists lie very close together, and such a compact
pool has a high cosine to many works of other artists as well. This is the situation for which CSLS~\cite{conneau2018word} was
designed. In its original form the neighbours of a vector are taken from
the other one of two domains. Here there is a single corpus, so let $r(x)$
be the mean cosine of $x$ to its $m$ nearest neighbours in the corpus,
where $x$ itself is left out. The CSLS score of
an image against a pool is
\begin{equation}
  \mathrm{CSLS}(x, A) = 2\, s(x, A) - r(x) - \frac{1}{|A|} \sum_{a \in A} r(a) .
\end{equation}
So the score of a pair is lowered if one of the two embeddings has close
neighbours, that is, a large value of $r$. We use $m = 15$, which is below the median pool size of $19$. The
correction works on the embeddings that are already there, and hence it
needs no further pass through the encoder and no training.
App.~\ref{app:csdplus_grid} gives its cost, and App.~\ref{app:corpus} the
stability of its result. In
its original use the two neighbourhood terms come from two different
spaces of word embeddings. That the correction also works within a single
space of image embeddings fits the view that hubs are a property of
high-dimensional embeddings in general and not of language alone.

\begin{table}[t]
\centering
\caption{Number of artists with a negative gap for the pairwise cosine,
the pooled cosine and the pooled CSLS score.}
\label{tab:readouts}
\begin{tabular}{lrrrr}
\toprule
Corpus & Artists & Pairwise & Pooled cosine & Pooled CSLS \\
\midrule
Commons-89  & 89   & 20 (22.5\%)  & 12 (13.5\%)  & 4 (4.5\%) \\
WikiArt     & 128  & 19 (14.8\%)  & 10 (7.8\%)   & 3 (2.3\%) \\
ArtBench-10 & 1639 & 646 (39.4\%) & 479 (29.2\%) & 298 (18.2\%) \\
\bottomrule
\end{tabular}
\end{table}

With the CSLS score the number of negative pooled gaps on Commons-89 falls
from $12$ to $4$ (Tab.~\ref{tab:readouts}). The two corpora on which the readout was not
chosen behave in the same way, with a reduction from $10$ to $3$ on WikiArt
and from $479$ to $298$ on ArtBench-10.

\begin{figure}[t]
\centering
\includegraphics{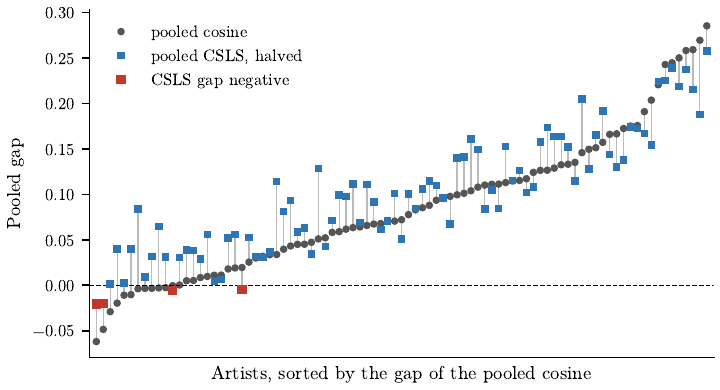}
\caption{Pooled gap of each of the $89$ artists for the cosine and for the
CSLS score. The gap of the CSLS score is divided by two, so that both are
on the scale of a cosine. The four artists whose gap stays negative under
CSLS are marked in red.}
\label{fig:before-after}
\end{figure}

Fig.~\ref{fig:before-after} shows the effect for each artist. The CSLS
score contains the cosine twice, so its gap is divided by two before it is
compared with the gap of the cosine. On this common scale the median gap
rises from $0.072$ to $0.099$. The gap grows for $58$ of the $89$ artists,
among them $11$ of the $12$ with a negative gap, and it becomes smaller
mainly for artists with a compact pool. For Goya it falls from $0.269$ to
$0.188$ and stays clearly positive. So the correction does not simply raise
all gaps. It moves margin from the compact pools, which have enough of it,
to the scattered ones. The result
depends little on the neighbourhood size. Between $m = 5$ and $m = 15$ four
negative gaps remain, and for $m = 20$ and $m = 50$ there are five and six.

A closer look shows that the density term does two things at once. The
neighbours of a work are searched in the whole corpus, and at $m = 15$
about $40\%$ of them are works of the same artist. So $r(x)$ also measures
how compact the artist's own pool is. If the works of the same artist are
excluded from the neighbourhood, $9$ instead of $4$ negative gaps remain.
Hence roughly half of the reduction from $12$ to $4$ comes from the
neighbours among other artists, which is the effect CSLS was built for, and
the other half is a normalisation of the artist's own pool. We keep the
variant with all neighbours nevertheless. The variant that excludes the
artist needs the label of the query, and in an evaluation this label is
what one wants to find out. The same holds for the balanced variant. In
the verification of \S\ref{sec:verification} the question does not arise,
because there the test artists are not part of the reference corpus.

CSLS is not the only readout that one could try. Tab.~\ref{tab:calibration}
compares it with two others. The margin scores an image against its pool
minus its score against the strongest other pool,
\begin{equation}
  m(x, A) = s(x, A) - \max_{B \neq A} s(x, B) .
\end{equation}
On both corpora it removes about as many negative gaps as
CSLS. The z-score $z(x, A) = (s(x, A) - \mu_A)/\sigma_A$ standardises the
score of an artist by the mean and the spread of the scores of the
artist's own works. It maps the within and the cross values of a pool by
the same increasing linear function, and therefore it cannot change the
sign of a gap. Its counts in Tab.~\ref{tab:calibration} equal those of the
cosine by construction, and we list it as a negative control. The z-score
is useful if scores of different artists are to be put on one scale, but
this is a different problem from the one the diagnostic detects. The
order of the readouts is the same on WikiArt as on Commons-89, so it does
not depend on our choice of artists.

\begin{table}[t]
\centering
\caption{Number of artists with a negative pooled gap under four readouts
of the same embeddings, at the standard input of $224$ pixels.}
\label{tab:calibration}
\begin{tabular}{lrr}
\toprule
Readout & Commons-89 ($89$) & WikiArt ($128$) \\
\midrule
Cosine $s(x, A)$ & 12 & 10 \\
CSLS & 4 & 3 \\
Margin $m(x, A)$ & 4 & 0 \\
z-score & 12 & 10 \\
\bottomrule
\end{tabular}
\end{table}

A second correction without new training changes the input of the encoder
and not its readout. The ViT-L/14 of CSD was trained on images of
$224 \times 224$ pixels, which it cuts into a grid of $16 \times 16$
patches, and each position in this grid has a learned positional
embedding. For a larger input of $336 \times 336$ pixels the grid has
$24 \times 24$ positions, and the learned positional embeddings are
interpolated to this grid~\cite{oquab2024dinov2}. The image is not
enlarged for this purpose. It is resized and cropped from the source file,
so the encoder sees more pixels of the work and not a larger copy of the
same ones. The cost is about $2.25$ times as many patches per pass through
the encoder. App.~\ref{app:csdplus_grid} varies the input between $224$
and $448$ pixels. The area under the ROC curve of the pair verification in
\S\ref{sec:verification} is highest between $280$ and $336$ pixels and
falls beyond, and most of the gain is already reached at $280$. We use
$336$ because the LoRA test of \S\ref{sec:generated} was run at this size.
Five crops of $224$ pixels instead of one larger input give a smaller gain,
because averaging over crops reduces the variance but adds no further
detail of the work. Neither variant recovers the parts of the image that
the square crop removes.

We call the protocol that combines these elements CSD+. It is not a new
encoder. It runs the diagnostic on the anchor corpus, keeps the raw cosine
if no gap is negative, and otherwise uses the CSLS readout, optionally
together with the larger input. Fig.~\ref{fig:csdplus-pipeline} shows where
the two choices sit in the unchanged pipeline of CSD, and the procedure of
Fig.~\ref{fig:procedure} in \S\ref{sec:discussion} shows when they are
applied. Tab.~\ref{tab:csdplus} gives the number of negative pooled gaps for
each combination of readout and input. With the CSLS readout it is $4$ for
each of the three inputs of the table, so for the diagnostic the larger input adds nothing to CSLS.
In the pair verification of \S\ref{sec:verification} it does. There the
larger input alone raises the area under the ROC curve from $0.891$ to
$0.904$, CSLS alone to $0.905$, and both together to $0.911$. The two gains
do not add up, since $0.013$ and $0.014$ would give more than the combined
$0.020$. So the two corrections do not act on independent errors. A
plausible explanation is that part of the apparent density is caused by
detail that the small input does not show, and CSLS then corrects what the
larger input leaves. The default of CSD+ is therefore the CSLS readout at
$224$ pixels, which needs no further pass through the encoder, and the
larger input is the stronger setting if the cost of embedding the anchors
again is acceptable.

\begin{figure}[t]
\centering
\resizebox{\linewidth}{!}{%
\begin{tikzpicture}[
  every node/.style={font=\small},
  trunk/.style={draw, rounded corners=2pt, line width=0.5pt, align=center, inner sep=4pt, fill=gray!8},
  opt/.style={draw, rounded corners=2pt, line width=0.5pt, align=center, inner sep=2.5pt, minimum width=33mm, font=\scriptsize},
  optbase/.style={opt, fill=gray!18, draw=gray!55},
  optrec/.style={opt, fill=red!16, draw=red!70, line width=0.8pt},
  optalt/.style={opt, fill=blue!8, draw=blue!45},
  chooser/.style={draw, rounded corners=3pt, line width=0.5pt, dashed, inner sep=5pt},
  arrow/.style={->, >=Latex, line width=0.6pt},
]
\node[trunk, minimum width=20mm] (img) at (0,0) {input\\image};
\node[trunk, minimum width=22mm, minimum height=14mm] (enc) at (5.6,0) {CSD ViT-L/14\\ \scriptsize (frozen)};
\node[trunk, minimum width=22mm, minimum height=14mm] (emb) at (8.6,0) {$x \in \mathbb{R}^{768}$\\ \scriptsize unit length};
\node[trunk, minimum width=18mm] (score) at (15.5,0) {score};
\draw[arrow] (img) -- node[above, font=\scriptsize] {input} (enc);
\draw[arrow] (enc) -- (emb);
\draw[arrow] (emb) -- node[above, font=\scriptsize] {readout} (score);
\node[optrec]  (i224) at (2.8,3.0) {$224$ centre crop (CSD+ default)};
\node[optrec]  (i336) at (2.8,2.35) {$336$ with interpolation (CSD+ option)};
\node[optalt]  (imc)  at (2.8,1.7) {five crops of $224$ (compared)};
\node[chooser, fit=(i224)(i336)(imc), label={[font=\scriptsize\itshape, anchor=south]above:one input}] (inputchooser) {};
\draw[arrow, gray!60] (inputchooser.south) -- ++(0,-0.7);
\node[optbase] (cos)  at (12.0,3.0) {cosine $s(x,A)$ (CSD)};
\node[optrec]  (csls) at (12.0,2.35) {CSLS$(x,A)$ (CSD+)};
\node[optalt]  (zsc)  at (12.0,1.7) {margin, z-score (compared)};
\node[chooser, fit=(cos)(csls)(zsc), label={[font=\scriptsize\itshape, anchor=south]above:one readout}] (readoutchooser) {};
\draw[arrow, gray!60] (readoutchooser.south) -- ++(0,-0.55);
\end{tikzpicture}%
}
\caption{The two choices of CSD+ in the unchanged pipeline of CSD. Grey is
the usual readout of CSD, red marks the choices of CSD+, and blue the
variants that we compared. When the choices are applied is shown in
Fig.~\ref{fig:procedure}.}
\label{fig:csdplus-pipeline}
\end{figure}

\begin{table}[t]
\centering
\caption{Number of artists of Commons-89 with a negative pooled gap for
each combination of readout and input. App.~\ref{app:csdplus_grid} gives
all five input sizes with the median gaps.}
\label{tab:csdplus}
\begin{tabular}{lrrr}
\toprule
Readout & $224$ & five crops of $224$ & $336$ with interpolation \\
\midrule
Cosine   & 12 & 10 &  9 \\
CSLS     &  4 &  4 &  4 \\
Margin   &  4 &  4 &  3 \\
z-score  & 12 & 10 &  9 \\
\bottomrule
\end{tabular}
\end{table}

Two properties of the corrected score have to be kept in mind. On the one
hand, the neighbourhoods are computed on the reference corpus, and so the
score depends on this corpus. It restores the order within a fixed pool,
but values from different corpora cannot be compared, and there is no CSLS
score for a single image without a reference pool. On the other hand, not
all negative gaps disappear. On Commons-89 Ky\=osai, Zurbar\'an, Leighton
and Klimt remain, and on ArtBench-10 nearly a fifth of the artists.
Moreover, a gap that changes its sign is not necessarily large afterwards.
For Levitan it rises from $-0.029$ to $0.001$ on the common scale, and
Shishkin remains his worst competitor. So the density is the cause that a
readout can correct, at the price that the score is tied to a fixed
reference pool, and a corrected gap close to zero remains a warning. The
larger input of CSD+ helps in the verification but does not reduce the
number of negative gaps below that of CSLS.

\section{Technique and Tradition}
\label{sec:tradition}

We start with the technique. Commons-89 contains mainly paintings, prints,
drawings and watercolours, and $62$
of the $89$ artists are represented by more than one of these techniques.
For such an artist the within value $\wmed$ mixes pairs of the same
technique with pairs such as a watercolour against an oil painting. If CSD
separates techniques strongly, the mixed pairs pull $\wmed$ down, and a
negative gap can arise although no competitor is particularly close. To
test this, we computed $\wmed$ over pairs of the same technique only. Any
restriction to subgroups raises a median a little, so the shift has to be
compared with a null model in which the technique labels of each artist are
permuted. Under this null model $\wmed$ moves by $-0.0007$ on average. With
the true labels it rises by $0.0325$ over all artists and by $0.0420$ for
the artists with a negative gap. For comparison, $17$ of the $20$ negative
gaps are smaller than $0.042$ in magnitude, so the effect of the technique
is of the same order as the gaps themselves. After a Benjamini-Hochberg correction of the permutation tests at a level of $0.05$,
the rise is significant for $22$ artists, and $8$ of the $20$ artists with
a negative gap are among them.

A global correction does not help here. We estimated a mean shift for each
combination of two techniques from pairs of different artists and
subtracted it from all cosines. The number of negative gaps stays at $20$,
and a permutation of the technique labels gives $19.9$.
The reason is that the effect of the technique is much larger within an
artist than between artists. For two works of the same artist the mean
cosine is $0.496$ if the technique is the same and $0.366$ if it differs,
whereas for two works of different artists the values are $0.185$ and
$0.154$. One offset per combination cannot fit both.
In other words, CSD does not encode the artist and the technique as two
separate things.

What does help is to keep the technique fixed. Sixty-seven artists have at
least ten paintings. If all their works are used, $15$ of them have a
negative gap, and with their paintings alone the number is $9$. The pools
become smaller by this restriction, and we know from
\S\ref{sec:diagnostic} that smaller pools change the rate. Therefore we
drew pools of the same reduced size from the mixed works. They give
$19.8 \pm 2.0\%$ against $13.4\%$ for paintings only, so most of the
improvement is due to the technique. Six artists lose their negative gap,
among them Leighton and Klimt, and no artist acquires one. Further controls
of this test are given in App.~\ref{app:controls}.

One might conclude that anchors should be matched in every available
category. This is not the case. WikiArt provides a genre label for each
work, and the same test shows a genre effect for $123$ of its $128$
artists, among them all $19$ with a negative gap
(App.~\ref{app:external}). If the corpus is restricted to its largest genre, the portraits,
the number of negative gaps among the $99$ artists with enough portraits
rises from $10$ to $17$. The explanation is that the subject is part of
what distinguishes one painter from another. If every artist is reduced to
portraits, the artists become more alike, and the cross value grows faster
than the within value. So the anchors of an artist should be matched in
technique and left mixed in subject. In practice this means that the unit
of an evaluation is an artist together with a technique. If an artist is to
be evaluated in oil and in watercolour, there are two pools and two scores.
Each of these pools still needs enough anchors, because small pools raise
the rate of negative gaps. This rule rests on these two
experiments, and it should be read as a well-supported hypothesis.

Not all negative gaps can be explained by density or technique. There
remains a group of artists whose closest competitor is a neighbour in art
history. For some of them the technique can be excluded directly, because
the pool of the flagged artist consists of one technique only. This is the case for T\=ohaku
with Sharaku, Church with Cole, Rubens with Murillo and Titian with
Mantegna. With the CSLS score the two deepest cases on
Commons-89 are Ky\=osai, whose competitor is It\=o Jakuch\=u, and
Zurbar\'an, whose competitor is Vel\'azquez. With the larger input of CSD+
the four remaining artists are Klimt, Zurbar\'an, Bierstadt and Ky\=osai,
with gaps between $-0.017$ and $-0.003$ on the common scale of
\S\ref{sec:density}. Their competitors under this readout are Mary
Cassatt, Vel\'azquez, Inness and Hokusai. So the two inputs leave partly
different artists at the margin. For three of the four the competitor
comes from the same tradition, while the competitor of Klimt, Cassatt, does
not. On ArtBench-10, where many
artists of each movement are present, $298$ artists remain. The question is
whether this is a weakness of CSD. To answer it, we repeated the analysis
on three other encoders of the same size, which were trained without style
labels (Tab.~\ref{tab:backbones}). CLIP and SigLIP were trained on pairs
of images and texts, and DINOv2 on images alone.

\begin{table}[t]
\centering
\caption{Four encoders on Commons-89. The left part gives the number of
artists with a negative pooled gap and the median gap of the cosine, the
right part the area under the ROC curve of the pair verification of
\S\ref{sec:verification}, as mean and standard deviation over $25$ splits.}
\label{tab:backbones}
\label{tab:backbone-comparison}
\setlength{\tabcolsep}{4pt}
\begin{tabular}{lrrrrr}
\toprule
& \multicolumn{3}{c}{Negative gaps} & \multicolumn{2}{c}{Verification} \\
\cmidrule(lr){2-4}\cmidrule(lr){5-6}
Encoder & Cosine & Median gap & CSLS & Cosine & CSLS \\
\midrule
CSD ViT-L     & 12 & 0.072 & 4  & $0.891 \pm 0.015$ & $0.905 \pm 0.015$ \\
CLIP-ViT-L/14 & 14 & 0.034 & 2  & $0.885 \pm 0.017$ & $0.914 \pm 0.012$ \\
SigLIP-large  & 12 & 0.045 & 2  & $0.902 \pm 0.016$ & $0.915 \pm 0.013$ \\
DINOv2-Large  & 40 & 0.006 & 18 & $0.786 \pm 0.022$ & $0.826 \pm 0.017$ \\
\bottomrule
\end{tabular}
\end{table}

CSD, CLIP and SigLIP behave alike, with $12$ to $14$ negative gaps under
the cosine, and DINOv2 is much worse with $40$. The four encoders differ in more than
their supervision, for example SigLIP works at $384$ pixels, so these
differences should not be attributed to the training objective alone
(Tab.~\ref{tab:backbone-config}). In the verification CSD, CLIP and SigLIP
are close together, and only DINOv2 falls behind. The CSLS
score raises the area under the ROC curve on every encoder. CSD has the largest median
gap of the four. Its typical margin is therefore the widest, but the number
of flagged artists is not lower than on CLIP and SigLIP. The CSLS score reduces
the count on every encoder, and on every encoder some artists remain. The
same holds on WikiArt (Tab.~\ref{tab:backbone-external}), where DINOv2
even reaches a negative median gap. Each of the other three encoders flags between $12$
and $17$ of the $19$ artists that CSD flags. For only $8$ to $12$ of the $19$ artists the worst competitor is the same
as on CSD, but where it changes, it
almost always stays within the tradition. For Monet, for example, it is
Loiseau on one encoder and Sisley on another, and for Picasso it is
Modigliani or Braque. Other pairs, such as Braque with Gris, Gainsborough
with Reynolds and Hokusai with Kuniyoshi, hold on all four encoders.

\begin{table}[t]
\centering\small
\setlength{\tabcolsep}{5pt}
\begin{tabular}{lrrrrrr}
\toprule
& \multicolumn{4}{c}{negative gaps} & \multicolumn{2}{c}{agreement with CSD} \\
\cmidrule(lr){2-5}\cmidrule(lr){6-7}
backbone & pairwise & robust & aggregated & CSLS & also neg. & same worst-other \\
\midrule
CSD ViT-L & $19$ & $11$ & $10$ & $3$ & -- & -- \\
CLIP-ViT-L/14 & $32$ & $17$ & $17$ & $12$ & $12/19$ & $11/19$ \\
SigLIP-large & $29$ & $18$ & $16$ & $8$ & $14/19$ & $12/19$ \\
DINOv2-Large & $90$ & $72$ & $73$ & $50$ & $17/19$ & $8/19$ \\
\bottomrule
\end{tabular}
\caption{Cross-backbone diagnostic on the WikiArt dump ($128$ artists). The last two columns compare each backbone against the CSD result on the $19$ artists CSD flags. \S\ref{sec:tradition} discusses.}
\label{tab:backbone-external}
\end{table}

\begin{table}[h]
\centering\small
\setlength{\tabcolsep}{4pt}
\resizebox{\linewidth}{!}{%
\begin{tabular}{llrllr}
\toprule
backbone & checkpoint & px & preprocessing & feature & dim \\
\midrule
CSD ViT-L & \texttt{yuxi-liu-wired/CSD} & $224$ & shortest side, centre crop & projection head & $768$ \\
CLIP-ViT-L/14 & \texttt{openai/clip-vit-large-patch14} & $224$ & model processor & \texttt{get\_image\_features} & $768$ \\
SigLIP-large & \texttt{google/siglip-large-patch16-384} & $384$ & model processor & \texttt{get\_image\_features} & $1024$ \\
DINOv2-Large & \texttt{facebook/dinov2-large} & $224$ & resize 256, centre crop, ImageNet norm & CLS token & $1024$ \\
\bottomrule
\end{tabular}%
}
\caption{Full configuration of the four backbones. All embeddings are L2-normalised after extraction. No dimensionality harmonisation is applied, since every readout compared here is scale-free and works within one backbone at a time. Apart from the training objective, the backbones differ in ways that are visible in the table and limit what the comparison can attribute. SigLIP runs at its native $384$ pixels, the other three at $224$. The preprocessing paths differ because we use no common pipeline and run each model with the pipeline it was released with. A ranking across rows therefore mixes objective, data, architecture and input resolution.}
\label{tab:backbone-config}
\end{table}

A last check uses no labels and no art history at all. The $20$ pairs of
\S\ref{sec:diagnostic}, each consisting of a flagged artist and the
worst competitor, were fixed before any clustering was done. We then
clustered the CSD embeddings of Commons-89 with UMAP~\cite{mcinnes2018umap}
and HDBSCAN~\cite{campello2013hdbscan} for ten random seeds, that is, with a
nonlinear projection followed by a density-based clustering that needs no
fixed number of clusters. Depending on
the seed, the works of such a pair fall into a common cluster at a rate of
$0.22$ to $0.33$, whereas the rate for pairs of artists in general is
$0.012$ to $0.049$. Twenty pairs of
artists drawn at random did not reach the rate of the flagged pairs in any
of $2000$ draws. So the pairs that the
diagnostic names are also neighbours in the geometry of the embedding.
App.~\ref{app:controls} gives further details of the clustering and of the
comparison of the encoders.

\begin{figure}[!tb]
\centering
\includegraphics[width=\linewidth]{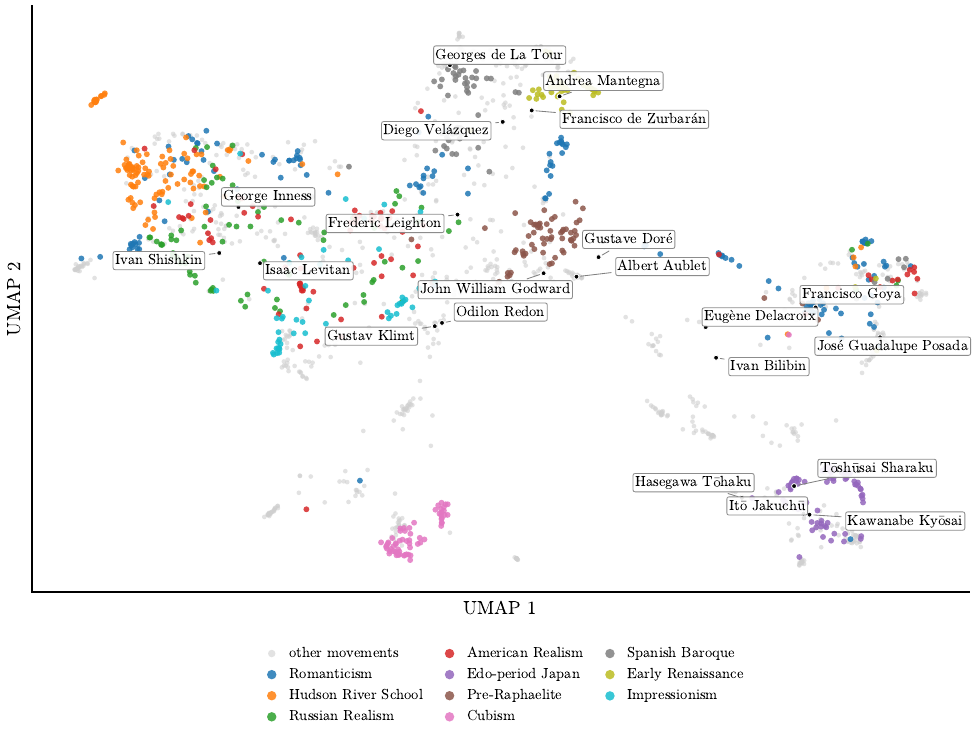}
\caption{UMAP projection of the $1645$ CSD embeddings from $768$ to two
dimensions. On average $43\%$ of the $15$ nearest neighbours of a work in the
original space are also among its $15$ nearest neighbours in the figure, so
groups that overlap here can still be separated in the original space. The
ten largest movements are coloured and the others are grey. The labelled
markers are the centroids of the eight most negative artists of
Tab.~\ref{tab:worst-gap}, of their worst competitors and of further artists that are discussed in the
text.}
\label{fig:umap}
\end{figure}

Fig.~\ref{fig:umap} gives an impression of this geometry, and the pairs of
Tab.~\ref{tab:worst-gap} can be found in it. The centroids of Levitan and
Shishkin are the nearest neighbours of each other, Ky\=osai, Sharaku and
Jakuch\=u form one group, and Posada and Delacroix lie close to Goya. The clusters
themselves correspond to families of styles and not to single artists. The largest one has $310$ works and joins landscape painters from four
countries. The technique is visible as well, in the form of two clusters of
works on paper around Goya and around the popular print.

Taken together, the two causes of this section call for different answers.
The effect of the technique can be avoided with one pool per artist and
technique. The overlap within a tradition is common to all four encoders,
although they were trained in different ways, and one cannot expect that
another choice of encoder removes it.

For CSD there is a plausible reason why artists of one tradition and of one
technique come so close, although we can offer it only as a hypothesis.
LAION-Styles carries several tags per image, for artists as well as for
media and movements, and the contrastive loss of CSD treats two images as
a positive pair as soon as they share any tag. Two artists of one
tradition often share the tags of the movement and of the medium, so
training pulls their works together instead of pushing them apart. The
binding to the technique would have the same origin, because a common
medium tag is a positive signal by construction. Somepalli et
al.~\cite{somepalli2024csd} already observe in their discussion that
impressionist landscape painters, and Picasso and Braque, are hard to
separate. Our results add that this is not limited to a few movements or
to Western art.

\section{A Test on Unseen Artists}
\label{sec:verification}

So far the CSLS score was judged by the same diagnostic that motivated it,
and on the corpus on which its neighbourhoods were computed. An independent
test should use artists that the correction has never seen. If the
separation of causes in the last two sections is correct, such a test
should also show a particular pattern. The correction should help most
where the competitor is a close stylistic neighbour, and it should help
little where two artists are far apart anyway.

We use pair verification for this test, a task known from face
recognition~\cite{schroff2015facenet}. Given two works, one has to decide
whether they are by the same artist. This is also a question of its own,
for example when a work is to be authenticated or attributed, and the
verifier must then decide for artists that it has never seen. The $89$ artists are split into $72$
reference artists and $17$ test artists. From the test artists we draw
$1500$ pairs of the same artist and $1500$ pairs of different artists and
compute the area under the ROC curve (AUC), for which $0.5$ is the level of chance, and this is
repeated for $25$ random splits. The density
term $r(x)$ of the CSLS score is computed against the works of the
reference artists only. So no information about the test artists enters
the correction.

Besides the cosine and the CSLS score we include two methods that are
trained on the reference artists. The first one is a logistic regression
on the pair feature
\begin{equation}
  \phi(a, b) = \bigl[\,|a - b|,\; a \odot b,\; (a + b)/2\,\bigr] \in \mathbb{R}^{2304} ,
\end{equation}
which gives a linear model the difference, the elementwise product and the
mean of the two embeddings, as is common for pairs of
sentences~\cite{conneau2017infersent}. The second one learns a Mahalanobis
distance. It reaches $0.887 \pm 0.022$ and is thus level with the cosine,
so learning a metric directly gains nothing on unseen artists.
Tab.~\ref{tab:robustness} gives the results for both inputs of CSD+.

\begin{table}[t]
\centering
\caption{Area under the ROC curve of the pair verification with random
negative pairs on Commons-89, as mean and standard deviation over $25$
splits, and its range over the splits.}
\label{tab:robustness}
\begin{tabular}{llrr}
\toprule
Method & Input & AUC & Range \\
\midrule
Cosine & $224$ & $0.891 \pm 0.015$ & $[0.861, 0.932]$ \\
CSLS & $224$ & $0.905 \pm 0.015$ & $[0.871, 0.937]$ \\
Cosine & $336$ & $0.904 \pm 0.013$ & $[0.872, 0.933]$ \\
CSLS & $336$ & $0.911 \pm 0.013$ & $[0.876, 0.936]$ \\
Logistic regression on pairs & $224$ & $0.923 \pm 0.013$ & $[0.887, 0.950]$ \\
Logistic regression on pairs & $336$ & $0.923 \pm 0.012$ & $[0.894, 0.944]$ \\
\bottomrule
\end{tabular}
\end{table}

On pairs that are drawn at random, the CSLS score raises the AUC from
$0.891 \pm 0.015$ to $0.905 \pm 0.015$. The gain is small, but it appears
in all $25$ splits. The splits overlap in their artists, so we determined
the uncertainty with a bootstrap over artists, which gives a $95\%$
interval from $0.011$ to $0.018$ for the gain. The correction of Nadeau and
Bengio~\cite{nadeau2003inference} for overlapping splits gives a wider
interval from $0.008$ to $0.020$, which also excludes zero.
Tab.~\ref{tab:uncertainty-designs} in App.~\ref{app:csdplus_grid} compares
further ways of computing the uncertainty. The gain is not uniform,
however. If the AUC is computed for each artist separately, $66$ of the
$89$ artists improve, and for the others the corrected score is slightly
worse. So the correction is a gain for a corpus as a whole and not a
guarantee for each single artist.

Most random pairs of different artists are far apart in art history, and
for them the raw cosine works well. To test the prediction, we need a
second way of drawing the negative pairs, in which each work of a test
artist is paired with works of the artist's worst competitor among the test
artists. We call the two ways the random regime and the worst-neighbour
regime. In the worst-neighbour regime the AUC of the cosine on Commons-89
falls to $0.709$, so pairs within a tradition cost about $0.18$ in AUC. As
a counterpart we restricted the test to artists with a gap above $0.05$.
There the cosine reaches $0.932$. The gap of an artist therefore tells in
advance on which pairs the raw cosine can be trusted. The
competitor is selected with the raw cosine, which could favour the
corrected score. If it is selected with the CSLS score instead, the gain
falls from $0.030$ to $0.023$, so by about a quarter, and remains clearly
positive. App.~\ref{app:controls} gives further checks of these
choices.

Negative pairs drawn from the same movement give a similar but noisier
test (App.~\ref{app:controls}). The logistic regression on pairs keeps the same lead of $0.018$
over the CSLS score at $224$ pixels in both regimes, and of $0.012$ over
CSLS with the larger input. So with supervision the AUC is higher
by a constant offset, but the difficulty within a tradition remains. If
labelled pairs exist, the logistic regression is the best method, and it
does not profit from the larger input.

\begin{table}[t]
\centering
\caption{Verification AUC for the cosine and the CSLS score in both
regimes, mean over $25$ splits, with the gain $\Delta$. The last column is
the skewness of the occurrence counts for $15$ neighbours. The grid is a
separate run, so its first row differs from the values in the text in the
third decimal.}
\label{tab:grid}
\setlength{\tabcolsep}{3.5pt}
\begin{tabular}{llrrrrrrr}
\toprule
& & \multicolumn{3}{c}{Random} & \multicolumn{3}{c}{Worst neighbour} & \\
\cmidrule(lr){3-5}\cmidrule(lr){6-8}
Corpus & Encoder & cos & CSLS & $\Delta$ & cos & CSLS & $\Delta$ & Skew \\
\midrule
Commons-89 & CSD & 0.890 & 0.905 & 0.015 & 0.709 & 0.737 & 0.028 & 0.93 \\
 & CLIP & 0.883 & 0.912 & 0.028 & 0.707 & 0.755 & 0.049 & 1.57 \\
 & SigLIP & 0.901 & 0.913 & 0.013 & 0.719 & 0.751 & 0.032 & 1.09 \\
 & DINOv2 & 0.786 & 0.824 & 0.038 & 0.583 & 0.659 & 0.076 & 1.66 \\
\midrule
WikiArt & CSD & 0.866 & 0.868 & 0.002 & 0.681 & 0.700 & 0.019 & 1.44 \\
 & CLIP & 0.842 & 0.861 & 0.020 & 0.659 & 0.700 & 0.041 & 1.92 \\
 & SigLIP & 0.860 & 0.870 & 0.010 & 0.673 & 0.709 & 0.036 & 1.91 \\
 & DINOv2 & 0.739 & 0.761 & 0.022 & 0.546 & 0.600 & 0.054 & 2.12 \\
\midrule
ArtBench-10 & CSD & 0.900 & 0.908 & 0.008 & 0.590 & 0.621 & 0.031 & 1.60 \\
 & CLIP & 0.877 & 0.899 & 0.021 & 0.564 & 0.614 & 0.050 & 2.48 \\
 & SigLIP & 0.884 & 0.899 & 0.015 & 0.564 & 0.607 & 0.043 & 2.59 \\
 & DINOv2 & 0.786 & 0.810 & 0.024 & 0.484 & 0.545 & 0.061 & 2.59 \\
\bottomrule
\end{tabular}
\end{table}

Tab.~\ref{tab:grid} shows the result for the three corpora and the four
encoders. In all twelve cells the gain in the worst-neighbour regime is
larger than the gain in the random regime, and it lies between $0.019$ and
$0.076$. This is the pattern that was predicted. The table also explains a
result that looks like a failure at first. For CSD on WikiArt the gain in
the random regime is only $0.002$, with a corrected interval from $-0.002$ to
$0.006$. So on easy pairs the correction does nothing there. In the
worst-neighbour regime of the same cell the gain is $0.019$. The aggregate
over random pairs therefore hides the effect, because it is dominated by
pairs for which no correction is needed.

The size of the gain differs between the encoders, and this can be related
to a quantity that is measured without any AUC. The usual index of hubness
is the skewness of the occurrence counts~\cite{radovanovic2010hubs}. The
occurrence count of a work is the number of other works that have it among
their $15$ nearest neighbours, and a high skewness means that a few works
appear in very many of these lists. On all three corpora CSD has the lowest skewness of the four
encoders, and on all three it receives the smallest correction in the
worst-neighbour regime. DINOv2 receives the largest correction on all three. This agrees with the explanation by
local density. It is an association and no proof, because the skewness and
the AUC of the raw cosine are correlated with each other.

One column of Tab.~\ref{tab:grid} must not be read from top to bottom. The
AUC of the cosine in the worst-neighbour regime falls from $0.709$ on
Commons-89 to $0.590$ on ArtBench-10, and this looks as if the score became
weaker with a growing number of artists. In fact the worst competitor is
searched among $16$ test artists on Commons-89 and among $326$ on
ArtBench-10, so the larger corpus is searched harder for a difficult
negative. If the search is limited to $16$ competitors everywhere, CSD
reaches $0.709$, $0.699$ and $0.730$ on the three corpora and the decline
disappears. The same holds for the other three encoders
(App.~\ref{app:controls}). This is the same effect of a maximum as in
\S\ref{sec:diagnostic}. The gains change only slightly, because both
scores of a cell see the same pairs.

The test thus confirms the separation of causes on artists that the
correction has not seen. The CSLS score helps where the competitor is
close, and the gap of an artist tells in advance whether this is the case.

\section{Tests with Generated Images}
\label{sec:generated}

\begin{figure}[t]
\centering
\setlength{\tabcolsep}{2pt}
\begin{tabular}{@{}ccccc@{}}
& {\footnotesize Original} & {\footnotesize Seedream~5.5} & {\footnotesize FLUX~3} & {\footnotesize Nano Banana~2} \\[2pt]
\rotatebox[origin=c]{90}{\footnotesize Grimshaw} & \begin{minipage}[c]{0.225\linewidth}\centering\vspace*{3pt}\includegraphics[width=\linewidth,height=2.35cm,keepaspectratio]{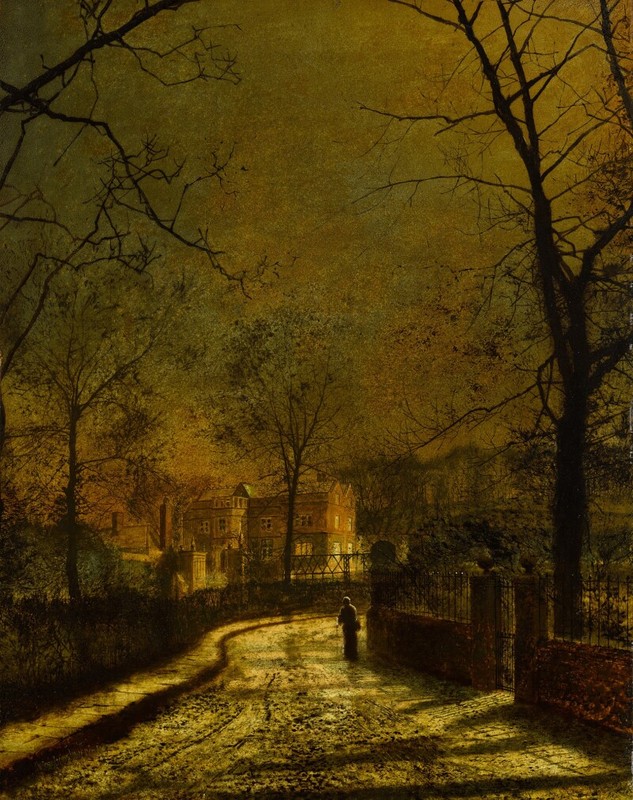}\par\vspace*{3pt}\end{minipage} & \begin{minipage}[c]{0.225\linewidth}\centering\vspace*{3pt}\includegraphics[width=\linewidth,height=2.35cm,keepaspectratio]{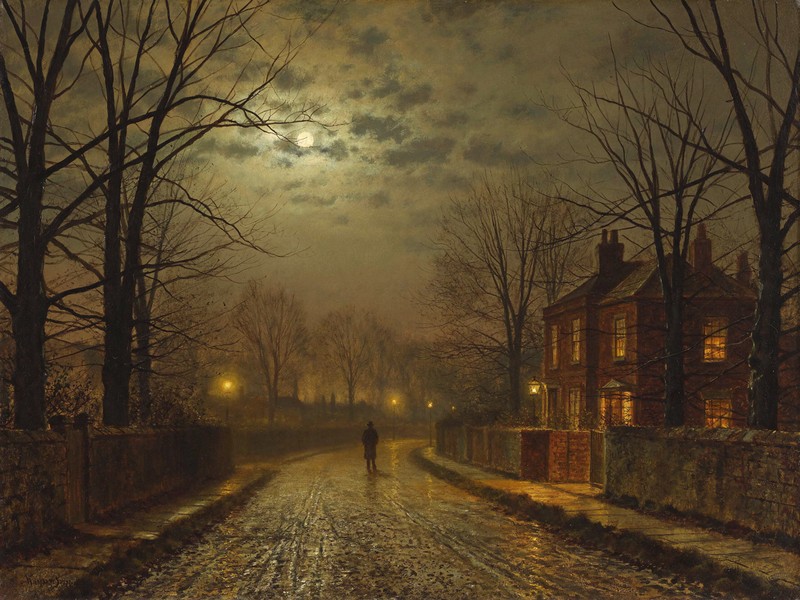}\par\vspace*{3pt}\end{minipage} & \begin{minipage}[c]{0.225\linewidth}\centering\vspace*{3pt}\includegraphics[width=\linewidth,height=2.35cm,keepaspectratio]{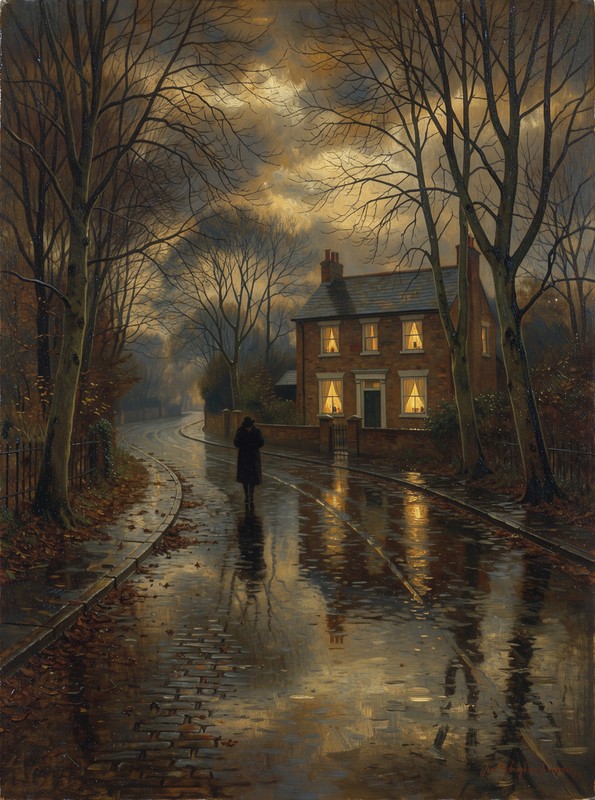}\par\vspace*{3pt}\end{minipage} & \begin{minipage}[c]{0.225\linewidth}\centering\vspace*{3pt}\includegraphics[width=\linewidth,height=2.35cm,keepaspectratio]{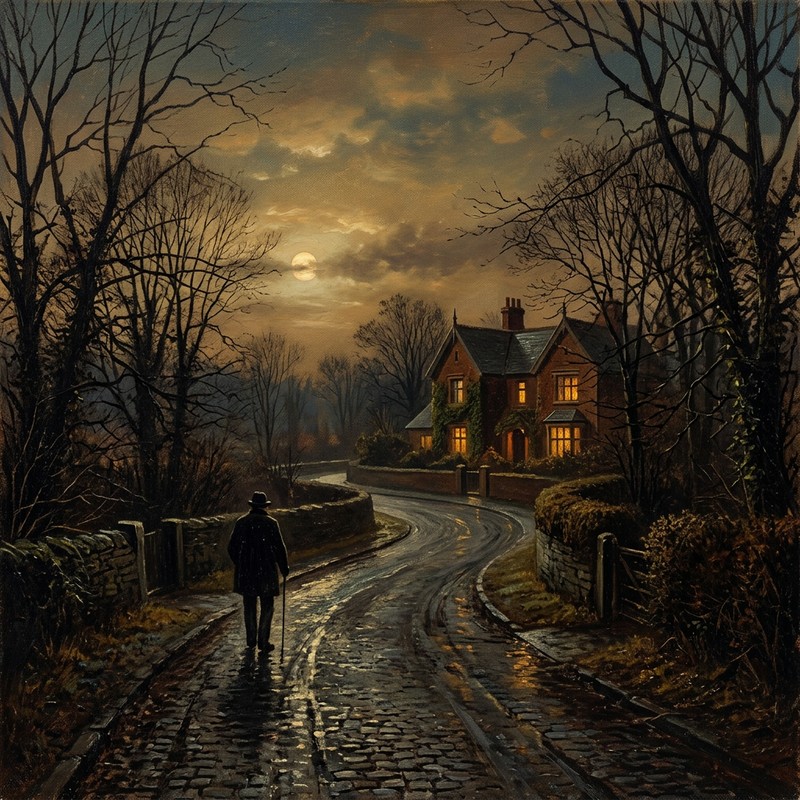}\par\vspace*{3pt}\end{minipage} \\
\rotatebox[origin=c]{90}{\footnotesize Hals} & \begin{minipage}[c]{0.225\linewidth}\centering\vspace*{3pt}\includegraphics[width=\linewidth,height=2.35cm,keepaspectratio]{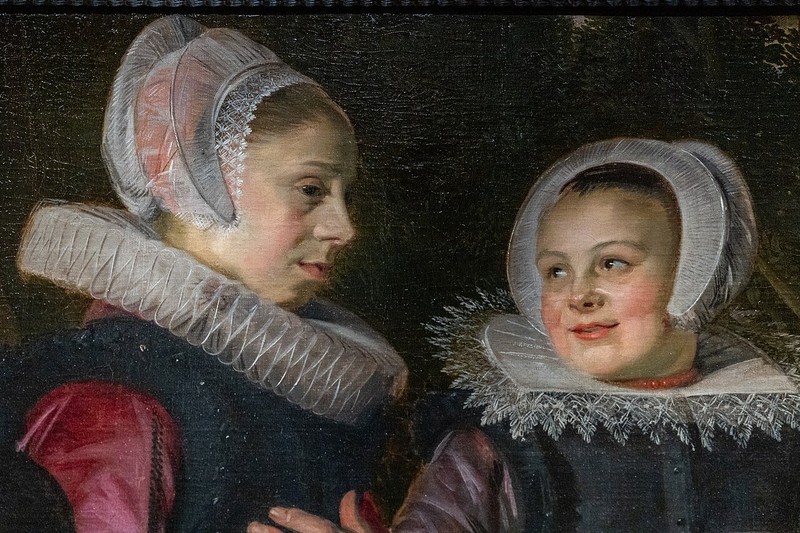}\par\vspace*{3pt}\end{minipage} & \begin{minipage}[c]{0.225\linewidth}\centering\vspace*{3pt}\includegraphics[width=\linewidth,height=2.35cm,keepaspectratio]{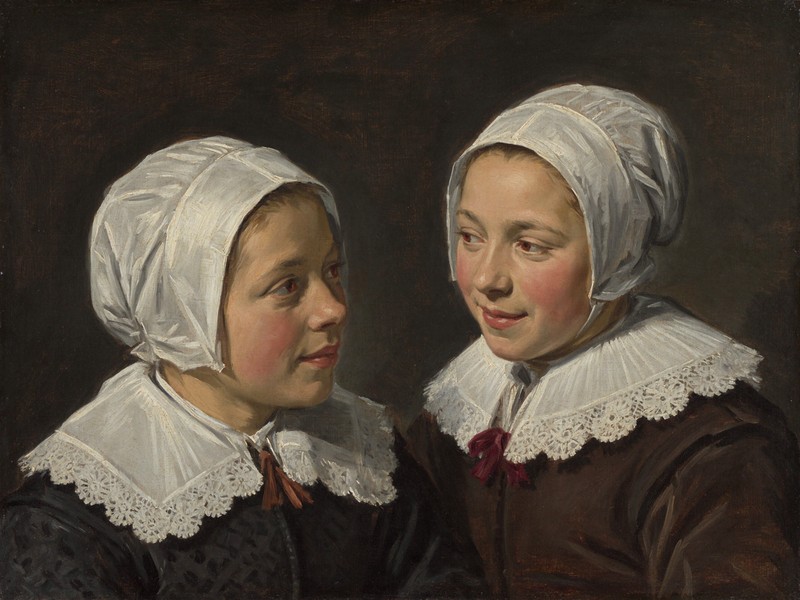}\par\vspace*{3pt}\end{minipage} & \begin{minipage}[c]{0.225\linewidth}\centering\vspace*{3pt}\includegraphics[width=\linewidth,height=2.35cm,keepaspectratio]{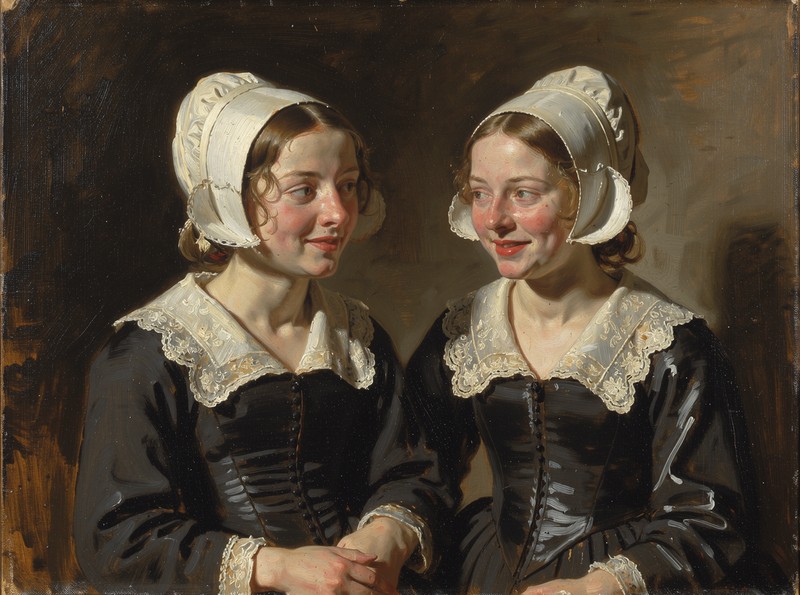}\par\vspace*{3pt}\end{minipage} & \begin{minipage}[c]{0.225\linewidth}\centering\vspace*{3pt}\includegraphics[width=\linewidth,height=2.35cm,keepaspectratio]{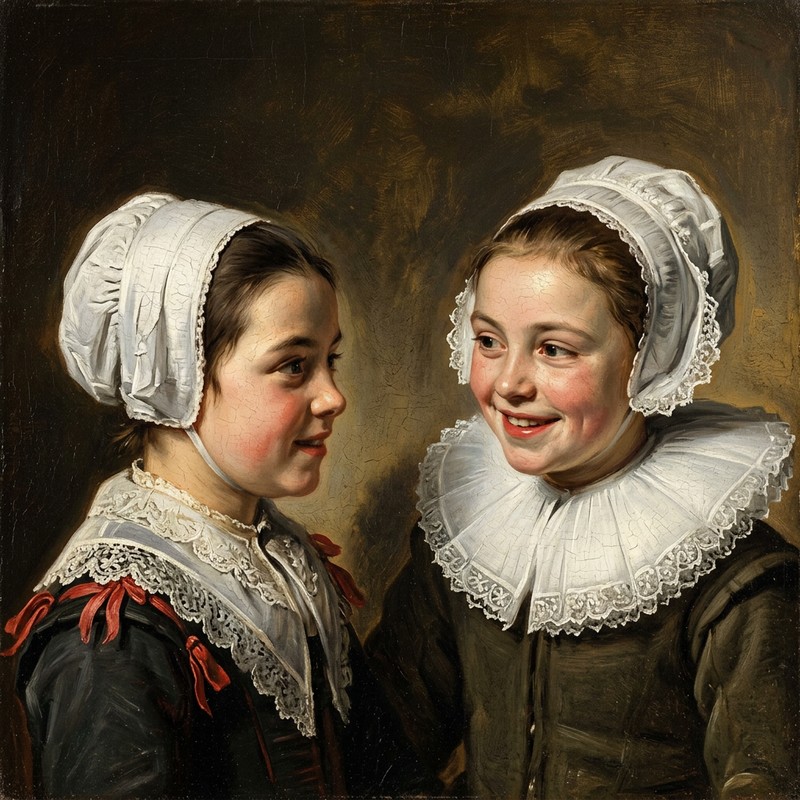}\par\vspace*{3pt}\end{minipage} \\
\rotatebox[origin=c]{90}{\footnotesize Zurbar\'an} & \begin{minipage}[c]{0.225\linewidth}\centering\vspace*{3pt}\includegraphics[width=\linewidth,height=2.35cm,keepaspectratio]{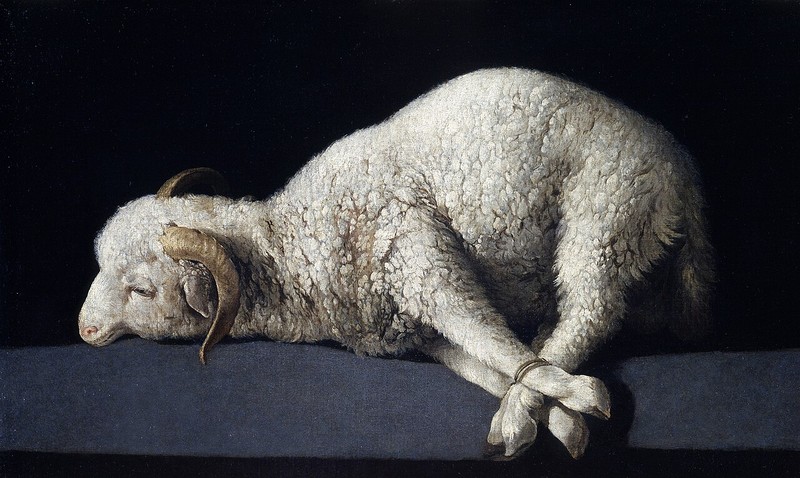}\par\vspace*{3pt}\end{minipage} & \begin{minipage}[c]{0.225\linewidth}\centering\vspace*{3pt}\includegraphics[width=\linewidth,height=2.35cm,keepaspectratio]{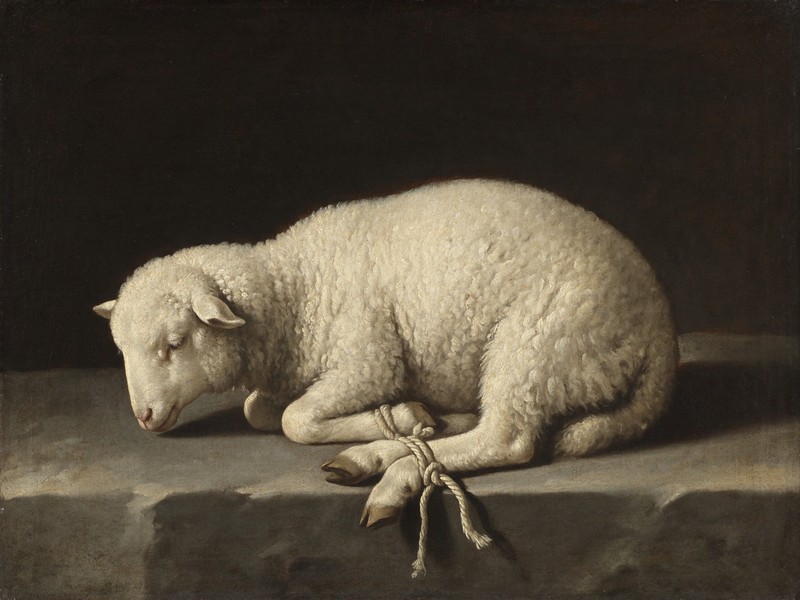}\par\vspace*{3pt}\end{minipage} & \begin{minipage}[c]{0.225\linewidth}\centering\vspace*{3pt}\includegraphics[width=\linewidth,height=2.35cm,keepaspectratio]{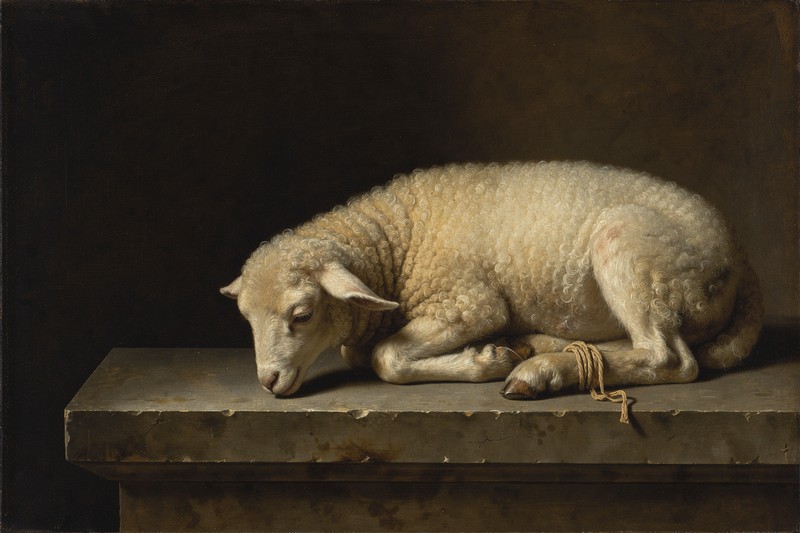}\par\vspace*{3pt}\end{minipage} & \begin{minipage}[c]{0.225\linewidth}\centering\vspace*{3pt}\includegraphics[width=\linewidth,height=2.35cm,keepaspectratio]{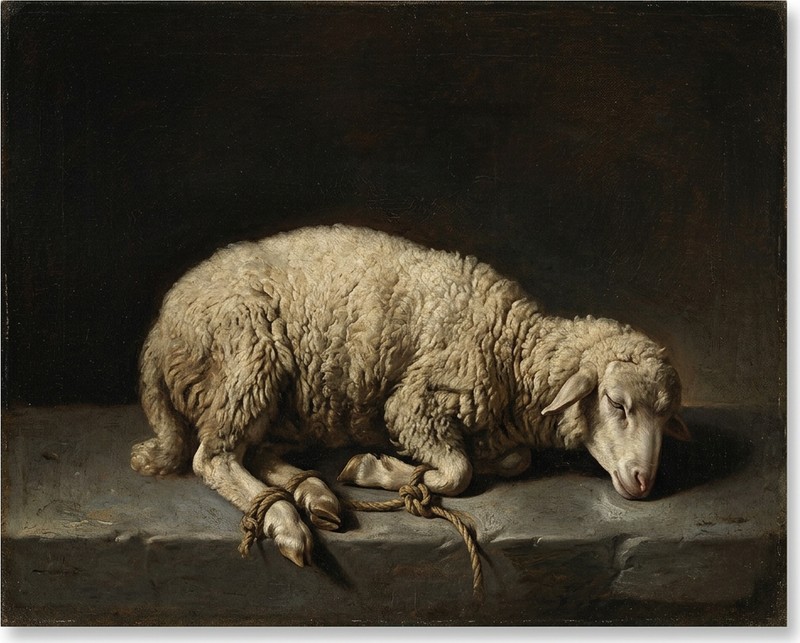}\par\vspace*{3pt}\end{minipage} \\
\end{tabular}
\caption{One work of each of three artists and the images generated from
it. For Grimshaw and Hals the pool of the artist has the highest score for all
four images of the row. For Zurbar\'an it has the highest score for none
of them. Margins of paper are cut off here.}
\label{fig:replicas}
\end{figure}

All experiments so far use authentic works. We close with two tests on
generated images. The first one uses unmodified models and many artists,
the second one adapts a model to a few pairs of artists. For the first
test we generate images for four artists with a large positive gap and for four artists with
a negative gap. The images come from four unmodified text-to-image models,
Seedream~5.5, FLUX~3, Kling~3 Omni and Nano Banana~2. Each prompt names the
artist and contains the technique and a description of the motif of one of
six works that were drawn at random from the pool of this artist, so that
the images stay with the subjects of the artist. In total $187$ generated
images enter the analysis, and five further ones were refused by a model or
show a picture frame. Fig.~\ref{fig:replicas} shows examples.

Each original and each generated image is scored against the $89$ pools
with the mean cosine, and the original is removed from its own pool for
this. Tab.~\ref{tab:replicas} counts how often the pool of the intended
artist has the highest score. For the artists with a positive gap this is
the case for $23$ of the $24$ originals and for $88$ of the $95$ generated
images. For the artists with a negative gap it is the case for $2$ of the
$24$ originals and for $1$ of the $92$ generated images, and this holds for
each of the four models. So the score treats a generated image as it treats
the work it is derived from. For an artist with a negative gap the first rank
therefore says nothing about the generator, because the authentic works do
not reach it either.
The agreement goes beyond the first rank. We compare the margin of an
image, that is, the score of the intended pool minus the highest score of
another pool. Over the $48$ works the margin of the original and the mean
margin of its generated images have a correlation of $0.92$. If the mean of
each artist is subtracted first, a correlation of $0.59$ remains. Hence a
work that the score separates poorly from other artists also leads to
generated images that are separated poorly.

The CSLS score does not change this result. With it the pool of the
intended artist has the highest score for $82$ of the $95$ generated images
in the first group and for $3$ of the $92$ in the second. For Ky\=osai and
Zurbar\'an this was to be expected, because their gap stays negative under
CSLS. The pooled gaps of Levitan and Dor\'e, however, become positive under
CSLS, with $0.001$ and $0.040$ on the scale of Fig.~\ref{fig:before-after},
and their images are still not assigned to them. So a gap that has turned
positive is not enough if it stays far below the median of $0.099$.

The test also shows where the images go instead, and here the two causes
of \S\ref{sec:tradition} appear again. The first one is the technique.
The pool of Dor\'e contains paintings, drawings and prints in about equal
parts, and the six drawn works are three oil paintings, two wood engravings
and one ink drawing. For the three works on paper the pool of Dor\'e has
rank $4$, $6$ and $9$ among the $89$ pools, and for the images generated
from them the median rank is $4.5$. For the three paintings the ranks are
$18$, $42$ and $43$, and the median rank of the generated images is $27.5$.
The name in the prompt is the same in both cases. The winning pools follow
the technique as well. They belong to the printmakers Piranesi, Goya and
Posada in the first case and to painters such as Cole and G\'ericault in
the second. Goya is the opposite case. His pool consists of etchings only,
and all images generated for him are etchings and have rank $1$. We
repeated the generation for Goya with three of the models and left out his
name, so that the prompt contains only the technique and the motif. The
pool of Goya still has the highest score for all $18$ images. For Hals,
whose technique is shared by most painters of the corpus, the same control
gives $10$ of $18$ instead of $16$ of $18$. Thus for Goya the first rank
shows that an image looks like an etching, and it does not show that the
image imitates Goya. This also puts the compact pool of Goya from
\S\ref{sec:diagnostic} in another light. Apart from the Japanese printmakers, only two other artists of the
corpus are represented mainly by prints, namely Piranesi and Posada, so for
Goya density and technique can hardly be told apart. A positive gap is a necessary condition, as stated in
\S\ref{sec:diagnostic}, and this example shows that it is not a
sufficient one.

The second cause is the tradition. The images generated for Levitan go to
Shishkin in $19$ of $24$ cases, those for Zurbar\'an go to La~Tour in $12$
of $23$ cases, and those for Ky\=osai go to Sharaku in $15$ of $21$ cases.
For each of the three artists this is the worst competitor that the
pairwise diagnostic of \S\ref{sec:diagnostic} finds on the authentic
works. The score places these images in
the right tradition, namely Russian landscape painting, tenebrism and
Japanese painting and printmaking, and it does not find the artist within
it. So the neighbour that the diagnostic names before any image is
generated is also the artist with whom the generated images are confused.

One might object that an unmodified model simply does not know these
artists well enough. An earlier experiment with FLUX.1-dev~\cite{labs2024flux}
shows that this is a real limit of prompting. We generated $66$ images for
each of $15$ artists of Commons-89 from the name, the movement, the
technique and one of $22$ neutral subjects, and asked whether the pool of
the artist has the highest score. Three groups appear
(App.~\ref{app:t2i_extras}). For artists such as La~Tour and Sharaku the
name works as a trigger, and the pool of the artist is first for $48\%$
and $26\%$ of the images. For a second group the images reach the right
tradition but not the artist. Hokusai, for example, is never first but
among the first five pools for $24\%$ of the images. For a third group,
among them Van Gogh, Monet, Klimt, Leighton and Levitan, the pool of the
artist is first for less than $3\%$. This is an audit of the generator and
not a test of the diagnostic, because a low value mixes how well the model
knows the name, how it reacts to the prompt and how CSD scores the result.

The second test therefore adapts the generator to the artist. We trained
LoRA adapters~\cite{hu2022lora} for FLUX.1-dev on both artists of three
pairs, Sharaku and Ky\=osai, Shishkin and Levitan, and Godward and
Leighton, and in each pair the second artist has the negative gap. Two
further adapters for Hokusai and Monet serve as positive controls. The
training images are not among the anchors of the evaluation, and the
prompts at inference are the captions of the anchors, so the images are
generated for the same subjects against which they are scored. This test
was run with the larger input of CSD+ for anchors and images. For each
image we ask whether its mean cosine to the pool of its own artist is
higher than to the pool of the partner. The share of such images is called
bidirectional pair discrimination, or bidir, in App.~\ref{app:t2i_extras}.

\begin{table}[t]
\centering
\caption{LoRA test: number of images for the artist with the negative gap
that are closer to the pool of this artist than to the pool of the partner,
without adaptation, with adaptation and the cosine, and with adaptation and
the CSLS score.}
\label{tab:lora-stress}
\begin{tabular}{llrrr}
\toprule
Artist & Partner & Unmodified & LoRA, cosine & LoRA, CSLS \\
\midrule
Ky\=osai & Sharaku & 0 of 66 & 2 of 60 & 44 of 60 \\
Leighton & Godward & 3 of 66 & 0 of 57 & 2 of 57 \\
Levitan & Shishkin & 0 of 66 & 9 of 60 & 18 of 60 \\
\bottomrule
\end{tabular}
\end{table}

With the cosine all three pairs show the same asymmetry
(Tab.~\ref{tab:lora-stress}). Every image of Sharaku, Godward and Shishkin
is closer to its own artist, whereas at least $85\%$ of the images of
Ky\=osai, Leighton and Levitan are closer to the partner. With the adapter, the images move into the region of the embedding that
the training works already occupy. According to the diagnostic, this region
overlaps with that of the partner. Therefore, within this recipe, the adaptation does not create a distinction that CSD does
not make for the authentic works. The CSLS score changes this for two of
the three pairs. It raises the share for Ky\=osai from $3\%$ to $73\%$ and
for Levitan from $15\%$ to $30\%$, and it leaves Leighton near zero. The
size of the gap does not order this outcome, because the pair with the most
negative pairwise gap at $224$ pixels is the one that gains most. The
outcome follows instead the gap of the anchors under the readout of the
test. With the CSLS score and the larger input, the pairwise gap of
Ky\=osai and Levitan becomes positive, with $0.011$ and $0.007$ on the
common scale of \S\ref{sec:density}, while that of Leighton stays negative
at $-0.009$. With three pairs this is a case
series and no more, and it holds only for the images of this recipe. The
practical consequence is that the diagnostic should be run on the anchors
with the readout that will be used for the generated images.

The positive controls confirm the converse only weakly. Without adaptation
the pools of Hokusai and Monet are never first among the $89$ pools. With
adaptation they are first for $8\%$ and $27\%$ of the images, which is a
clear rise but far from the values of Sharaku, Godward and Shishkin with
their adapters, $97\%$, $96\%$ and $68\%$. An adapter thus lifts an artist
that the name alone does not reach, but within limits. Both tests agree
in their main point, within the limits of their size. For the artists with
a negative gap that we tested, neither the prompt nor the adaptation of
our recipe made the raw cosine a valid score of the artist.

\begin{table}[t]
\centering
\caption{Number of images for which the pool of the intended artist has
the highest score among the $89$ pools, and median rank of this pool for
the generated images.}
\label{tab:replicas}
\begin{tabular}{lrccr}
\toprule
Artist & Gap & Originals & Generated & Median rank \\
\midrule
Goya       & $0.255$  & 6 of 6 & 23 of 23 & 1 \\
Sharaku    & $0.193$  & 6 of 6 & 24 of 24 & 1 \\
Hals       & $0.200$  & 5 of 6 & 21 of 24 & 1 \\
Grimshaw   & $0.268$  & 6 of 6 & 20 of 24 & 1 \\
\midrule
Levitan    & $-0.030$ & 1 of 6 & 0 of 24 & 3.5 \\
Zurbar\'an & $-0.008$ & 1 of 6 & 1 of 23 & 4 \\
Ky\=osai   & $-0.085$ & 0 of 6 & 0 of 21 & 5 \\
Dor\'e     & $-0.066$ & 0 of 6 & 0 of 24 & 18 \\
\bottomrule
\end{tabular}
\end{table}

\section{Recommendations and Limits}
\label{sec:discussion}

For the practice of evaluation we draw five recommendations from the
results. First, the reference corpus should contain several artists of each
technique that is evaluated. If an artist shares the technique with very
few others, as Goya does with his etchings on Commons-89, the gap is
positive and the score nevertheless recognises little more than the
technique, so such an artist is reported separately. Second, the diagnostic should be run on the anchor corpus before
cosine values are averaged over artists. The rate that it returns should be
reported together with the number of artists and the number of anchors per
artist, because without these two numbers it cannot be interpreted. Third,
if negative gaps occur, the CSLS score should be tried in place of the raw
cosine, provided that the reference pool is fixed and documented. It costs
no additional pass through the encoder, and it removes two thirds of the
negative gaps for pooled scores on Commons-89. If the anchors can be
embedded again, the larger input of CSD+ adds a further gain in
verification. A gap that is only just positive afterwards should be
treated like a negative one. Fourth, the anchors of an artist should be of one
technique, while the subjects should stay mixed. An artist who is to be evaluated in
several techniques gets one pool and one score per technique. Fifth, the artists that
are still flagged after these measures should be reported separately or
left out of the average. For these artists a high score may only show that
an image belongs to the right tradition, and the worst competitor that the
diagnostic returns tells with whom their images will be confused. For such
artists other means are needed, and we would first try a comparison by
people with reference works and an attribution at the level of the
tradition. Trained calibrations, prototypes per artist and other encoders
remain untested here. If labelled pairs of works are available, a logistic
regression on pair features is better than every readout of the cosine in
our verification, with the absolute difference $|a - b|$ as its strongest
part. Fig.~\ref{fig:procedure}
summarises the procedure.

\begin{figure}[!tb]
\centering
\begin{tikzpicture}[
  stp/.style={draw, align=center, font=\footnotesize, inner sep=3pt,
               text width=2.45cm, minimum height=11mm},
  dec/.style={draw, diamond, aspect=1.9, align=center, font=\footnotesize,
              inner sep=0.3pt, text width=1.35cm},
  fin/.style={draw, rounded corners=5pt, align=center, font=\footnotesize,
              inner sep=3pt, text width=2.0cm, minimum height=11mm},
  arr/.style={->, >=Latex, line width=0.5pt},
  lab/.style={font=\scriptsize}]
\node[stp] (s1) at (0,0)   {compute the gap with the raw cosine};
\node[stp] (s2) at (3.2,0) {use the CSLS score of CSD+ in a fixed pool};
\node[stp] (s3) at (6.4,0) {use one pool per artist and technique};
\node[stp, text width=8.7cm, minimum height=7mm] (s0) at (3.2,1.85)
  {make sure that each technique is shared by several artists of the corpus};
\draw[arr] (s0.south -| s1.north) -- (s1.north);
\node[dec]  (d1) at (0,-1.9)   {negative gaps?};
\node[dec]  (d2) at (3.2,-1.9) {negative gaps?};
\node[dec]  (d3) at (6.4,-1.9) {negative gaps?};
\node[fin, text width=8.7cm, minimum height=7mm] (ok) at (3.2,-3.7)
  {average the score over the artists with a positive gap};
\node[fin] (rep) at (9.45,-1.9) {report the flagged artists separately};
\draw[arr] (s1) -- (d1);
\draw[arr] (s2) -- (d2);
\draw[arr] (s3) -- (d3);
\draw[arr] (d1.east) -- ++(0.2,0) |- node[lab, pos=0.25, right] {yes} (s2.west);
\draw[arr] (d2.east) -- ++(0.2,0) |- node[lab, pos=0.25, right] {yes} (s3.west);
\draw[arr] (d3.east) -- node[lab, above] {yes} (rep.west);
\draw[arr] (d1.south) -- node[lab, right] {no} (d1.south |- ok.north);
\draw[arr] (d2.south) -- node[lab, right] {no} (d2.south |- ok.north);
\draw[arr] (d3.south) -- node[lab, right] {no} (d3.south |- ok.north);
\end{tikzpicture}
\caption{Recommended procedure before scores are averaged over artists.
Rectangles are steps, diamonds are decisions and rounded boxes are
outcomes. The gap is computed again after each step, and a gap close to
zero is treated like a negative one. Where the CSLS score and the optional
larger input of CSD+ sit in the pipeline is shown in
Fig.~\ref{fig:csdplus-pipeline}.}
\label{fig:procedure}
\end{figure}

The gap needs at least two anchors per
artist, so it is not defined for an evaluation with a single reference
image. It tests the order of two medians.
It is a necessary condition, and an artist who passes it can still be
poorly separated from the nearest competitor. The CSLS score depends on the
reference pool, and therefore values from different corpora or from
different versions of one corpus must not be compared. The controls for
attribution and technique were only possible on Commons-89, because
the large dumps provide neither attribution labels nor technique labels,
and the LoRA test needs anchors that are not among the training images of
the adapter. This is the reason why we built a small audited corpus in
addition to the large ones.
The effect of the technique was measured on CSD only. The neighbourhood
size $m = 15$ and the input of $336$ pixels were chosen on Commons-89. A
sweep of $m$ on WikiArt gives an equally flat picture, so we do not expect
that this choice matters much. The sweep of the input size was not
repeated on the other corpora, but $280$ and $336$ pixels differ by only
about $0.001$ in the area under the ROC curve.

Finally, the tests with generated images are small. The first one covers
eight artists, four generators and at most one image per work and
generator. The LoRA test covers one generator, three pairs of artists and
two positive controls, and it is a case series and not a study of
generators in general. The first test shows that
a generated image is ranked like the work it is derived from. Neither test shows
how well a generator imitates an artist, and for a flagged artist the raw
cosine cannot show this.

\section{Conclusion}
\label{sec:conclusion}

The introduction asked three questions. The answer to the first one is
that the rate grows with the number of artists and falls with the number
of anchors per artist. If both are made equal, the three corpora lie close together. With
ten anchors per artist a high value of the raw CSD cosine is no evidence for
the particular artist for about a fifth to a quarter of $89$ artists. For the second question the
diagnostic alone is not enough, because a negative gap has three causes.
The CSLS readout of CSD+, which is not a new encoder, corrects the density, and its optional larger input
adds a gain in verification. The technique is corrected by the choice of the
anchors. The overlap within a tradition, however, remains on all four
encoders. The test on unseen artists supports this separation, because the
correction helps most against the closest neighbour for every corpus and
encoder. The procedure of Fig.~\ref{fig:procedure} follows from it. Each
cause has its own measure, a different readout for the density, a different
choice of anchors for the technique and a separate report for the
tradition, and the measures are applied in this order because each one is
more costly or more restrictive than the one before. The answer to the third question is yes. Images from four
text-to-image models are scored like the works they are derived from, and
they are confused with the competitor that the diagnostic names
beforehand. In our case series, a generator adapted with LoRA did not change this under the
raw cosine. Among the three causes the technique reaches furthest, because
it also acts where the gap is positive. The images for Goya land in his
pool even if the prompt does not name him.

For the comparisons from which this paper started, the consequence is the
following. If two methods are compared on the same artists and anchors, the
level of the score of each artist cancels in the difference, and the
diagnostic does not question the sign of such a difference. What it
questions is the meaning. For about a fifth of the artists a method can
gain score by moving towards the tradition or the technique and not towards
the artist. A difference of two to four percentage points in the average
should therefore be reported separately for the artists that pass the
diagnostic and for those that do not, and this split is known from the
anchor corpus before any image is generated.

\bibliographystyle{plainnat}
\bibliography{refs}

\clearpage
\appendix
\section{Further controls for the main text}
\label{app:controls}

This appendix collects controls that support statements of the main text
but are not needed to follow it. They are ordered by the sections they
belong to.

\paragraph{Attribution (\S\ref{sec:diagnostic}).}
If the $55$ works on which the two stages of the audit disagree are left
out, $17$ of $88$ artists are flagged. Raffaello is an instructive case.
His pool holds four copies after him and one work of his school, and
without them the pair with Mantegna is no longer flagged. His shallow gap
of $-0.009$ is therefore at least partly due to the attribution.

\paragraph{Technique (\S\ref{sec:tradition}).}
The direction of the effect is almost always the same. For $56$ of the
$62$ artists with several techniques, the pairs across two techniques have
lower cosines than the pairs within one technique. By comparison, tests
for several modes in the cosines of the $27$ artists with a single
technique flag only one to five of them, depending on the
test~\cite{pfister2013bimodality}. So the technique explains most of the
apparent spread within an artist. Averaged over all pairs, the effect of
the technique within an artist is $4.2$ times the effect between artists.
If the baseline between artists is subtracted in each combination of
techniques, the contrast within an artist shrinks only from $0.311$ to
$0.212$, by $32\%$. Neither the composition of the artists nor the subject
is controlled in this comparison, so it shows an association and not the
cost of a change of technique.

In the test on paintings alone, $15$ of the $67$ artists are flagged with
all works and $9$ with their paintings, which is $22.4\%$ against
$13.4\%$. Besides the pools of reduced size from the mixed works, random
subsets of $70$ artists of the mixed corpus give $19.5 \pm 2.2\%$
(App.~\ref{app:corpus}). So about two and a half of the nine percentage
points are due to the size of the pools, and the rest is due to the
technique. The six artists that lose their negative gap are Leighton,
Bussiere, Klimt, Tanner, Ky\=osai and Raffaello. Only painting has enough
artists for this test, so for the printmakers and for the most strongly
mixed pools the question remains open. If the test is repeated with the
CSLS score on the larger input of CSD+ and with the pairwise gap, the count
among the $67$ artists moves from $4$ to $5$. Leighton loses his negative
gap, Sargent and T\=ohaku acquire one, and Klimt, Bierstadt and
Vel\'azquez keep theirs. So part of the residual is due to the technique
and part is not.

\paragraph{Encoders and clusters (\S\ref{sec:tradition}).}
In the verification the distance of DINOv2 to CSD is $0.105$, with a
$95\%$ interval of $[0.082, 0.127]$ over the paired splits, and this is
the only difference among the four encoders that the splits resolve. In
the clustering, the flagged pairs share a cluster at a rate that is
$10.6 \pm 5.0$ times the rate of pairs of artists in general, and the
random draws stay below it on all ten seeds. Between $15$ and $19$ of the
$20$ pairs share a cluster at all, and works that HDBSCAN leaves unassigned
count against a common cluster. The clusters agree only weakly with the
labels of the movements, with an adjusted Rand index of $0.033 \pm 0.002$,
against $0.145 \pm 0.009$ for the labels of the artists. App.~\ref{app:geometry} gives the composition of all clusters.

\paragraph{Selection of the competitor (\S\ref{sec:verification}).}
Selecting the competitor on anchors that are not scored changes almost
nothing. In the same way the artists for the restricted test can be
selected on held-out anchors, which gives $0.913$ against $0.917$. Both
values lie below the $0.932$ of the main text, because there the gap is
computed among the test artists only and therefore admits more artists.

\paragraph{Movement labels as negatives (\S\ref{sec:verification}).}
A movement label would be an obvious alternative to the worst competitor.
If the negative pairs are drawn from the same movement, the cosine reaches
$0.746$, which is close to the worst-neighbour regime, but this test is
noisier. Of our $45$ movement labels, $22$ have a single artist. A split
contains only $2.1 \pm 1.2$ movements with several test artists, and only
$28\%$ of the artists have their worst competitor in their own movement.
Therefore the worst competitor is the more stable test for pairs within a
tradition.

\paragraph{External corpora in the verification (\S\ref{sec:verification}).}
For CSD on WikiArt the corrected interval of the gain in the random regime
runs from $-0.002$ to $0.006$. We regard a gain below $0.01$ as negligible,
because the smallest gain that we recommend is $0.014$, and the interval
lies below this margin. On ArtBench-10 the gain in the random regime is
$0.008$, with an interval from $0.005$ to $0.011$, which is small but
established. The restricted test with artists whose gap exceeds $0.05$
shows the other side. There CSLS adds $0.013$ on Commons-89 and $0.002$ on
the two other corpora (Tab.~\ref{tab:verification-crosscorpus}).

\paragraph{Number of competitors (\S\ref{sec:verification}).}
Tab.~\ref{tab:argmax-pool} gives the AUC of the cosine in the
worst-neighbour regime for all four encoders, once with the competitor
searched among all test artists and once among $16$ of them.

\begin{table}[h]
\centering
\caption{AUC of the cosine in the worst-neighbour regime on the corpora
with $89$, $128$ and $1639$ artists, once with the competitor searched
among all test artists and once among $16$ of them. Commons-89 has only
$16$ competitors, so its column must not move.}
\label{tab:argmax-pool}
\setlength{\tabcolsep}{4pt}
\begin{tabular}{lrrrrrrr}
\toprule
& \multicolumn{3}{c}{All competitors} & \multicolumn{3}{c}{$16$ competitors} & \\
\cmidrule(lr){2-4}\cmidrule(lr){5-7}
Encoder & $89$ & $128$ & $1639$ & $89$ & $128$ & $1639$ & Spread \\
\midrule
CSD ViT-L & 0.709 & 0.681 & 0.590 & 0.709 & 0.699 & 0.730 & 0.030 \\
CLIP-ViT-L/14 & 0.707 & 0.659 & 0.564 & 0.707 & 0.682 & 0.702 & 0.024 \\
SigLIP-large & 0.719 & 0.673 & 0.564 & 0.719 & 0.691 & 0.706 & 0.028 \\
DINOv2-Large & 0.583 & 0.546 & 0.484 & 0.583 & 0.560 & 0.587 & 0.027 \\
\bottomrule
\end{tabular}
\end{table}

\section{Corpus details}
\label{app:corpus}

The 400-artist list released with the CSD code repository (file
\texttt{artists\_400.txt}) contains 400 raw entries with substantial
duplication. The same artist appears repeatedly under different name
spellings, e.g.\ ``Mucha'', ``Alphonse Mucha'' and ``Alfonse Mucha''.
After canonical-name deduplication 367 distinct artists remain. We
augment the list with fourteen non-Western artists from Edo-period Japan
(Hokusai, Hiroshige, Utamaro, Sharaku, Itō Jakuchū, Hasegawa Tōhaku,
Kawanabe Kyōsai), Ming-dynasty China (Shen Zhou) and modern Chinese ink
(Wu Changshuo), Joseon-dynasty Korea (Kim Hong-do), Persian miniature
(Beh\-z\=ad), Indian Academic realism (Raja Ravi Varma), and Mexican
popular print and Surrealism (Posada, Kahlo). To fill specific
under-represented traditions, we further augment with Spanish Baroque
(Vel\'azquez, Murillo, Zurbar\'an), French Romanticism (Delacroix,
G\'ericault), American Tonalism (Whistler, Inness), Cubism (Juan Gris,
Robert Delaunay, Albert Gleizes, Jean Metzinger), Mughal miniature
(Ustad Mansur), Naive art (Henri Rousseau), Early-Renaissance Italian
painting (Botticelli, Mantegna), and additional Symbolist and French
Academic entries (Odilon Redon, Gustave Moreau,
William-Adolphe Bouguereau). The augmented list is the input to the
fetcher.

For each canonical name we try to fetch public-domain artworks from
Wikimedia Commons via the Wikidata-resolved Commons category. The
fetcher tries structured sub-categories (\emph{Paintings by X},
\emph{Sculptures by X}, \emph{Woodblock prints by X}, etc.) and
falls back to the Wikidata-canonical category for the specific artist.
It never falls back to the bare canonical name, because on Commons this
often resolves to a disambiguation node such as ``Vermeer (surname)''
that pulls in homonyms. We fix a target of
twenty-five artworks per artist with a hard minimum of fifteen.
All retained anchor images were obtained from Wikimedia Commons, which
hosts only public-domain or freely licensed files.

The fetched candidates then go through a two-stage attribution audit.
Stage 1 applies regular-expression filters on the filename together
with the Wikimedia ``Artist'' metadata field. Stage 2 sends the textual
metadata of every borderline file (image description, object name,
categories, permission and credit fields) to a
language-model judge, which classifies the entry into one of seven
attribution labels: \emph{master}, \emph{workshop},
\emph{school}, \emph{after}, \emph{attributed}, \emph{imitator}, or
\emph{off\_topic}/\emph{wrong\_person}. Of the 2736 candidates across 113
fetched artists, 2273 pass the audit (2117 master, 14 workshop, 7 school,
117 after, 18 attributed). The counts reflect a logged re-run
of the judge on $91$ files whose first call had returned malformed
responses. The re-run reclassified nine ambiguous records as autograph
and removed four works as not by the named artist. The remaining 463
files are imitator-class works, modern photographs of unrelated
subjects, or files that share an artist's name but depict a different
person (a horse named ``Wilm Vermeir'' is the most striking example).

A second per-image content audit runs on every retained candidate via a
local vision--language model (Qwen3-VL-32B-Instruct~\citep{qwen3vl}) with a structured
JSON-schema prompt. Each image is classified along four axes:
\emph{is\_artwork\_by\_artist}, \emph{artwork\_kind} $\in$
\{\texttt{painting}, \texttt{drawing}, \texttt{watercolor},
\texttt{print}, \texttt{mixed\_media}, \texttt{sculpture},
\texttt{other}, \texttt{photograph}\}, \emph{has\_frame}, and
\emph{has\_museum\_context}. The first axis is false for historical
photographs of the artist or their travels, museum-reproduction pages,
book covers, memorial plaques and similar non-work imagery. The
\texttt{photograph} category marks camera-made works and non-art
imagery, not the photographic reproduction every anchor consists of.
Items flagged in this way, or as not the artist's own work, are
dropped, and so are items with the frame or museum-context flag, so
that frame pixels do not leak into the embedding under
multi-crop or pos-interp readouts. On the post-attribution pool the
content audit removes about $20\%$ of the images overall and leaves
$1795$ admissible strict-clean anchors. The removals are concentrated in
artists whose Commons pages are dominated by archive photographs of the
painter, framed museum-display shots and book-cover scans rather than
autograph artworks. On top of the two automated audits, one author then
reviewed every remaining anchor image in a complete pass with logged
decisions. Near-duplicate reproductions of one physical work were
reduced to a single keeper, reproduction prints by another hand were
relabelled as \emph{after}, and misattributions were removed. $90$
reproduction photographs were also cropped to the artwork region where
mounts, colour-calibration bars, scroll brocade or book folds intruded,
and the cropped rows re-embedded in all backbones.
The review reduces the pool to $1645$ anchors across $89$ artists. The
ten-anchor floor is applied after every stage that removes images, not
only once at the end. Both figures quoted here, $1795$ after the content
audit and $1645$ after the review, are therefore already net of the
artists that the respective stage pushed below it.
Together with the fifteen-anchor minimum at the attribution stage these
pipeline steps remove twenty-four artists in total, among them Caravaggio,
Frida Kahlo and Leonardo da Vinci at the attribution stage, Rembrandt ($8$
strict-clean anchors after frame/museum removal) and William Morris at
the content audit, and Mikhail Vrubel and Ustad Mansur at the review.
The full list is part of the audit records. The remaining candidates are
entries that we attempted but could not fetch because they have no
usable Wikimedia presence. Most of them are contemporary digital
illustrators whose works are under copyright. Fig.~\ref{fig:pipeline}
summarises the cascade.

\begin{figure}[h]
\centering
\resizebox{\linewidth}{!}{%
\begin{tikzpicture}[
  every node/.style={font=\small},
  pipestep/.style={draw, rounded corners=2pt, line width=0.5pt, align=center, inner sep=5pt, fill=gray!8, minimum height=12mm, minimum width=28mm},
  arrow/.style={->, >=Latex, line width=0.6pt},
]
\node[pipestep] (fetch)  at (0,0)    {fetched\\113 artists\\2736 images};
\node[pipestep] (attr)   at (4.0,0)  {attribution audit\\113 artists\\2273 images};
\node[pipestep] (att2)   at (8.0,0)  {content audit\\+ ten-anchor floor\\91 artists\\1795 images};
\node[pipestep] (cont)   at (12.0,0) {human review\\+ floor reapplied\\89 artists\\1645 images};
\node[pipestep] (final)  at (16.0,0) {working corpus\\89 artists\\1645 anchors};
\draw[arrow] (fetch) -- (attr);
\draw[arrow] (attr)  -- (att2);
\draw[arrow] (att2)  -- (cont);
\draw[arrow] (cont)  -- (final);
\end{tikzpicture}%
}
\caption{Corpus pipeline, with artists and images after each stage. The
ten-anchor floor is re-applied whenever a stage removes images, so every
count shown is already net of artists who fell below it. The final box
repeats the counts of the fourth.}
\label{fig:pipeline}
\end{figure}

For each of the 89 retained artists we manually attach a primary
art-movement or tradition label drawn from a 47-label vocabulary covering
the major Western traditions (Hudson River School, Pre-Raphaelite, Vienna
Secession, Dadaism, etc.) and the non-Western categories the corpus
addition introduces (Edo-period Japan, Momoyama-period Japan,
Muromachi-period Japan, Meiji-period Japan, Ming-dynasty China, Late-Qing
China, modern Chinese ink, Joseon-dynasty Korea, Persian miniature, Indian
Academic Realism, Mexican popular print). Movement labels are author-curated from standard art-historical references. Borderline
cases, such as Modigliani as ``Modernism'' rather than ``School of
Paris'' or Hubert Robert as ``Romanticism'' rather than
``Pre-Romantic'', are defensible but not unique. Each artist enters the
aggregate statistics only once, so the headline numbers are robust to
single-artist re-classification. Under reasonable label revisions, the
absolute purity values for individual clusters in
App.~\ref{app:geometry} would shift by at most a few percentage points,
and the qualitative ranking among methods would not change.

The full roster of the $89$ retained artists with their primary movement
labels is part of the per-artist audit records. The corpus spans $45$ of
the $47$ vocabulary labels, from single-artist traditions (Persian
miniature, Joseon-dynasty Korea) up to Romanticism with eight members.
The ten traditions represented by at least three artists that
\S\ref{sec:diagnostic} analyses are identifiable from those records.

Wikimedia Commons coverage is uneven across the 89 retained artists: Rubens,
Goya and Klimt each have several hundred public-domain works indexed on
Commons, while less canonical artists in the same public-domain era are
represented by fewer than thirty. The working pool is formed after the
automated audits and the complete human review, i.e.\ the two-stage
attribution audit above, the per-image content audit and the review of
\S\ref{sec:corpus}. It holds the post-audit strict-clean set per artist,
capped at $30$ admissible images. Artists whose audited pool exceeds
$30$ are downsampled quality-first, preferring frame-free and
museum-context-free images. Artists whose post-audit pool falls short are
kept as-is, provided they clear the ten-anchor strict-clean
floor. The resulting per-artist
counts have $\mathrm{median}=19$ and lie in $[10, 27]$. We do not weight
the downstream analysis by sample reliability: an artist with thirty
broadly-sampled works available is described by a more representative
sample of their oeuvre than one with thirty narrowly-themed works.

\paragraph{Per-artist bootstrap CIs for the discrimination gap.}
Tab.~\ref{tab:worst-gap-full23} extends the aggregate bootstrap of
\S\ref{sec:diagnostic} to per-artist $95\%$-confidence intervals on
$\gap$ (seed $0$). These intervals use the target-only design, in which
the artist's own anchors are redrawn while the competing pools keep
their original composition. This matches the question the interval
asks. The aggregate count redraws every pool instead, so the two
figures are not two readings of one scheme. The boundary cases
are visible in the bounds. The third
robust\_negative slot alternates because the CI upper bounds of
Bilibin and Tōhaku at seed $0$ are $-0.004$ and $+0.001$. The
multiplicity bands, the coverage simulation behind reading a pointwise
interval as roughly a $90\%$ statement, and the reason an
artist-level bootstrap cannot widen the scope of the intervals are in
App.~\ref{app:robustness}.

\paragraph{Cross-fitting the competitor selection.}
For the inferential reading of a flag, that a negative gap says
something about the artist rather than about this pool, the argmax in
$\cmed$ is a selection effect. We bound it by cross-fitting. Each artist's anchors are split, the competitor is
selected on one part, and both gap terms are estimated on the other.
Since halving anchors inflates the rate on its own
(\S\ref{sec:diagnostic}), the in-sample column is computed at the same
evaluation depth. Splitting each artist in half gives $27.2\%$ in sample
against $16.7\%$ cross-fitted, with the mean gap moving by $+0.033$
(Tab.~\ref{tab:gap-crossfit}). The two columns form a bracket and do not
converge, because cross-fitting errs the other way. A selection part
that is small enough to leave a deep evaluation part picks the full
corpus's worst competitor for only $45\%$ of artists, and a competitor
that is not the worst inflates the gap. Of
the $20$ artists flagged at full depth, eleven stay negative in the
majority of cross-fits: Bierstadt, Bilibin, Church, Delacroix, Dor\'e,
Hiroshige, Kyōsai, Leighton, Levitan, Rubens and Tōhaku. That list
contains the flagged member of the deep shared-tradition pairs this
paper names, among them Levitan/Shishkin, Leighton/Godward,
Hiroshige/Utamaro and Sharaku/Kyōsai. Of those pairs only the shallow
Vel\'azquez/La~Tour drops out; it is negative in exactly half of the
cross-fits. The correction therefore removes tail cases and leaves the
examples the argument rests on.

\begin{table}[h]
\centering\small
\begin{tabular}{rrrrr}
\toprule
eval.\ share & in sample & cross-fitted & bias in $\gap$ & same competitor \\
\midrule
$0.75$ & $22.3\%$ & $11.0\%$ & $+0.042$ & $45\%$ \\
$0.50$ & $27.2\%$ & $16.7\%$ & $+0.033$ & $65\%$ \\
$0.25$ & $38.6\%$ & $24.4\%$ & $+0.046$ & $79\%$ \\
\midrule
$1.00$ & $22.5\%$ & -- & -- & -- \\
\bottomrule
\end{tabular}
\caption{Negative-gap rate when the worst competitor is
selected on the evaluation anchors (in sample) or on a disjoint part of
them (cross-fitted), at equal evaluation depth. The last row is the
full-depth number, without a cross-fitted counterpart. The
last column shows how often the cross-fitted selection picks the same
competitor as the full corpus.}
\label{tab:gap-crossfit}
\end{table}

\begin{table}[h]
\centering
\small
\setlength{\tabcolsep}{4pt}
\begin{tabular}{lrrrlrll}
\toprule
Artist & $n$ & $w_k$ & $c_k$ & Worst other & Gap $g_k$ & 95\% CI & Cls \\
\midrule
Kawanabe Kyōsai & 20 & +0.449 & +0.534 & Tōshūsai Sharaku & -0.085 & [-0.146, -0.036] & $\bullet$ \\
Hasegawa Tōhaku & 12 & +0.420 & +0.499 & Tōshūsai Sharaku & -0.078 & [-0.171, +0.001] & $\circ$ \\
Gustave Doré & 23 & +0.263 & +0.330 & Francisco Goya & -0.066 & [-0.139, -0.015] & $\bullet$ \\
Ivan Bilibin & 21 & +0.250 & +0.284 & José Guadalupe Posada & -0.035 & [-0.138, -0.004] & $\bullet$ \\
Isaac Levitan & 20 & +0.289 & +0.319 & Ivan Shishkin & -0.030 & [-0.079, +0.012] & $\circ$ \\
Albert Aublet & 16 & +0.225 & +0.247 & Francisco Goya & -0.022 & [-0.092, +0.079] & $\circ$ \\
Frederic Leighton & 17 & +0.293 & +0.313 & John William Godward & -0.020 & [-0.103, +0.044] & $\circ$ \\
Gustav Klimt & 18 & +0.283 & +0.302 & Odilon Redon & -0.020 & [-0.112, +0.072] & $\circ$ \\
Diego Velázquez & 15 & +0.422 & +0.440 & Georges de La Tour & -0.018 & [-0.095, +0.048] & $\circ$ \\
Eugène Delacroix & 15 & +0.433 & +0.449 & Francisco Goya & -0.017 & [-0.081, +0.040] & $\circ$ \\
Albert Bierstadt & 24 & +0.401 & +0.417 & George Inness & -0.016 & [-0.059, +0.033] & $\circ$ \\
John Singer Sargent & 18 & +0.270 & +0.284 & Edward Hopper & -0.014 & [-0.103, +0.032] & $\circ$ \\
Raffaello Sanzio & 15 & +0.455 & +0.465 & Andrea Mantegna & -0.009 & [-0.146, +0.089] & $\circ$ \\
Gaston Bussiere & 21 & +0.266 & +0.274 & John William Waterhouse & -0.009 & [-0.133, +0.040] & $\circ$ \\
Francisco de Zurbarán & 13 & +0.419 & +0.427 & Georges de La Tour & -0.008 & [-0.080, +0.032] & $\circ$ \\
Frederic Edwin Church & 19 & +0.473 & +0.480 & Thomas Cole & -0.007 & [-0.075, +0.048] & $\circ$ \\
Henry Ossawa Tanner & 23 & +0.260 & +0.265 & James McNeill Whistler & -0.005 & [-0.068, +0.021] & $\circ$ \\
Utagawa Hiroshige & 17 & +0.584 & +0.587 & Kitagawa Utamaro & -0.003 & [-0.069, +0.060] & $\circ$ \\
Titian & 19 & +0.370 & +0.372 & Andrea Mantegna & -0.002 & [-0.042, +0.037] & $\circ$ \\
Peter Paul Rubens & 17 & +0.427 & +0.428 & Bartolomé Esteban Murillo & -0.001 & [-0.061, +0.039] & $\circ$ \\
\bottomrule
\end{tabular}
\caption{All 20 negative-gap artists on the strict-clean anchor set with per-artist bootstrap CIs (1000 resamples, seed 0); duplicate resampled anchors are excluded from the within-median so no artificial unit cosines enter. Cls: $\bullet$~=~robust\_negative (CI upper $<0$), $\circ$~=~ambiguous (CI contains 0).}
\label{tab:worst-gap-full23}
\end{table}

\paragraph{Sensitivity to the artist-pool composition.}
Because $c_k$ is a maximum over other artists, we also test how
sensitive the gap is to the artist set itself
(Tab.~\ref{tab:subsample-sensitivity}). Under leave-one-artist-out (S1),
removing any single artist shifts the aggregate neg-gap fraction by less
than one percentage point on average. Of the $15$ artists in main-text
Tab.~\ref{tab:worst-gap}, $6$ flip positive when their worst-other
is removed, all of them in the shallow tail (Vel\'azquez and Zurbar\'an
on La~Tour, Delacroix on Goya, Bierstadt on Inness, Raffaello on
Mantegna, Bussiere on Waterhouse), and the remaining $9$ stay negative.
Under random $70$-artist subsamples (S2, $100$ resamples, seed $0$) the
neg-gap fraction is $0.195 \pm 0.022$, range $[8, 17]$ artists. The
identity-preservation check under the same subsamples (S3) holds at
$100\%$. For raw cosine, however, this follows from the definition,
since removing other artists changes neither gap term. The real
subsample test is the CSLS variant below, whose scores depend on the
pool and could have flipped.

\begin{table}[h]
\centering
\small
\setlength{\tabcolsep}{5pt}
\begin{tabular}{@{}lll@{}}
\toprule
Setup & raw cosine & CSLS ($k{=}15$) \\
\midrule
S1 leave-one-out (89 runs): neg-gap fraction & $0.22 \pm 0.006$ & $0.04 \pm 0.004$ \\
\phantom{S1} flips when worst-other removed & $6/15$ & $3/4$ \\
S2 random 70-artist subsamples (100 runs) & $0.20 \pm 0.022$, counts $[8,17]$ & $0.03 \pm 0.013$, counts $[0,4]$ \\
S3 worst-other identity preserved & $100\%$, gap negative in $100\%$ & $100\%$, gap negative in $93\%$ \\
\bottomrule
\end{tabular}
\caption{Sensitivity of the raw-cosine and CSLS-aggregated gap to
artist-pool variation: leave-one-out (S1), random $70$-artist subsamples
(S2), and worst-other stability conditional on both pair members being in
the pool (S3).}
\label{tab:subsample-sensitivity}
\label{tab:csls-subsample-sensitivity}
\end{table}

\paragraph{CSLS pool sensitivity.}
Because CSLS depends on the reference pool, we repeat the subsample
analysis for the CSLS-aggregated gap. On the full pool of $89$ artists,
four artists have a negative CSLS gap: Kyōsai, Zurbar\'an, Leighton and
Klimt, with gaps from $-0.041$ to $-0.010$ on the unhalved scale of the CSLS
score, which is $-0.021$ to $-0.005$ on the common scale of
\S\ref{sec:density}.
Tab.~\ref{tab:csls-subsample-sensitivity} reports the same three setups.
The fractions are essentially unchanged under leave-one-out and random
subsampling. For all four artists the worst-other identity is preserved
in $100\%$ of subsamples, and the gap remains negative in $93\%$ of
conditional runs ($100\%$ for Kyōsai and Zurbar\'an, $93\%$ for
Leighton, $77\%$ for Klimt). CSLS is therefore at least as pool-robust
as raw cosine, and the residual after CSD$+$ is not an artefact of the
candidate-pool composition.

Tab.~\ref{tab:medium-confound} lists the per-artist technique test of
\S\ref{sec:tradition}. For every artist with a mixed technique record it
gives the shift in $\wmed$ under same-technique restriction, its
permutation-null $p$-value and the Benjamini--Hochberg verdict.

\begin{table}[t]
\centering\small
\setlength{\tabcolsep}{4pt}
\begin{tabular}{lrlrrrl}
\toprule
artist & $n$ & techniques & $\gap$ & $\Delta \wmed$ & $p$ & worst-other \\
\midrule
Kawanabe Kyōsai$^{\ast}$ & $20$ & pt~15, wc~4, pr~1 & $-0.0852$ & $+0.0615$ & $0.005$ & Tōshūsai Sharaku \\
Hasegawa Tōhaku & $12$ & pt~12 & $-0.0783$ & $+0.0000$ & $1.000$ & Tōshūsai Sharaku \\
Gustave Doré$^{\ast}$ & $23$ & pt~9, dr~7, pr~7 & $-0.0662$ & $+0.1212$ & $<0.001$ & Francisco Goya \\
Ivan Bilibin$^{\ast}$ & $21$ & pr~7, dr~7, pt~5, wc~2 & $-0.0347$ & $+0.0995$ & $<0.001$ & José Guadalupe Posada \\
Isaac Levitan & $20$ & pt~19, dr~1 & $-0.0304$ & $+0.0239$ & $0.054$ & Ivan Shishkin \\
Albert Aublet & $16$ & pt~8, wc~4, dr~3, pr~1 & $-0.0217$ & $-0.0150$ & $0.606$ & Francisco Goya \\
Frederic Leighton$^{\ast}$ & $17$ & pt~15, pr~1, dr~1 & $-0.0200$ & $+0.0378$ & $0.002$ & John William Godward \\
Gustav Klimt & $18$ & pt~16, dr~2 & $-0.0197$ & $+0.0381$ & $0.069$ & Odilon Redon \\
Diego Velázquez & $15$ & pt~13, pr~1, dr~1 & $-0.0176$ & $+0.0407$ & $0.037$ & Georges de La Tour \\
Eugène Delacroix & $15$ & wc~9, dr~5, pr~1 & $-0.0168$ & $+0.0282$ & $0.074$ & Francisco Goya \\
Albert Bierstadt & $24$ & pt~22, dr~1, wc~1 & $-0.0162$ & $+0.0222$ & $0.149$ & George Inness \\
John Singer Sargent & $18$ & pt~16, wc~2 & $-0.0139$ & $+0.0016$ & $0.361$ & Edward Hopper \\
Raffaello Sanzio & $15$ & pt~13, pr~2 & $-0.0093$ & $+0.0943$ & $0.022$ & Andrea Mantegna \\
Gaston Bussiere$^{\ast}$ & $21$ & pt~13, pr~6, dr~2 & $-0.0086$ & $+0.1248$ & $<0.001$ & John William Waterhouse \\
Francisco de Zurbarán$^{\ast}$ & $13$ & pt~11, pr~2 & $-0.0082$ & $+0.0367$ & $0.011$ & Georges de La Tour \\
Frederic Edwin Church & $19$ & pt~19 & $-0.0071$ & $+0.0000$ & $1.000$ & Thomas Cole \\
Henry Ossawa Tanner$^{\ast}$ & $23$ & pt~16, dr~5, wc~2 & $-0.0048$ & $+0.0360$ & $0.012$ & James McNeill Whistler \\
Utagawa Hiroshige$^{\ast}$ & $17$ & pr~12, dr~2, pt~2, wc~1 & $-0.0031$ & $+0.0878$ & $<0.001$ & Kitagawa Utamaro \\
Titian & $19$ & pt~19 & $-0.0022$ & $+0.0000$ & $1.000$ & Andrea Mantegna \\
Peter Paul Rubens & $17$ & pt~17 & $-0.0014$ & $+0.0000$ & $1.000$ & Bartolomé Esteban Murillo \\
\bottomrule
\end{tabular}
\caption{Technique composition of the negative-gap artists on Commons-89 under raw cosine (pt painting, pr print, dr drawing, wc watercolour, mm mixed media). $\Delta \wmed$ is the shift when $\wmed$ is restricted to same-technique pairs; $p$ is one-sided against a permutation null with shuffled technique labels at fixed group sizes. Starred rows are significant under Benjamini--Hochberg at $q < 0.05$ over the $89$ artists. \S\ref{sec:tradition} discusses.}
\label{tab:medium-confound}
\end{table}

\section{Robustness of the diagnostic}
\label{app:robustness}

This appendix collects the statistical controls behind the headline
numbers of \S\ref{sec:diagnostic}.

The aggregate bootstrap and the per-artist intervals estimate different
quantities, and so they use different resampling designs. The
per-artist interval asks how much a single artist's own anchor sample
could have moved that artist's gap. It therefore resamples the target
artist only and holds the competing pools at their original
composition, since $\cmed$ is a property of the corpus the artist is
evaluated against and not of the artist's sample. The aggregate count
instead asks how many artists would be flagged on a differently sampled
corpus. Here every pool is redrawn with replacement at its original
size, and pairs of identically drawn anchors are excluded from the
within-median so that no artificial unit cosines enter. Across $1000$
such resamples the count is $23.7 \pm 2.7$, range $[16, 34]$. This
distribution is not centred on the observed $20$. The reason is the
same arithmetic as in the anchor-depth sweep of
Tab.~\ref{tab:inventory-controls}, and not an instability. A bootstrap
draw of $n$ items holds about $0.632n$ distinct ones. Each replicate is
therefore effectively a thinner anchor pool, here with $11.9$ distinct
anchors against $18.5$, and a thinner pool raises the rate. Redrawing
at the matched effective depth without replacement accounts for about a
third of the upward shift. The remainder comes from the duplicates
themselves, which make both gap terms noisier while the cross term is a
maximum over $88$ of them. The aggregate count is therefore stable in
spread, but by construction it lies above the point estimate. The
corpus's rate is the point estimate and not the bootstrap mean.

The per-artist intervals are pointwise, and multiplicity matters as
soon as they are scanned. Asking about one artist chosen independently
of the data is a single question. The three artists reported below
zero, however, are the winners of a scan over $89$, and the error rate
for a scan is family-wise. Under a band that controls this error rate,
no individual artist survives. This holds both for Bonferroni and for
one critical value calibrated on the bootstrap distribution of the
largest studentised deviation across artists, which gives a quantile of
$4.38$ against the pointwise $1.96$ (Tab.~\ref{tab:gap-bands}). Both
instruments are conservative at $1000$ replicates. We therefore report
the three as the artists with the most clearly negative pointwise
intervals, not as a set established at $95\%$ across the corpus.
Nothing at corpus level depends on this distinction. The rate needs no
per-artist significance, and the named pairs have independent support
from the leave-one-out flip analysis, from surviving cross-fitting, and
from having an art-historically identifiable partner.

\begin{table}[t]
\centering\small
\begin{tabular}{lrrrr}
\toprule
band & half-width & neg. & pos. & ambiguous \\
\midrule
pointwise $95\%$ & $0.075$ & $3$ & $41$ & $45$ \\
Bonferroni over $89$ & $0.126$ & $0$ & $21$ & $68$ \\
bootstrap sup-$t$ & $0.171$ & $0$ & $14$ & $75$ \\
\bottomrule
\end{tabular}
\caption{Per-artist gap bands over the $89$ artists, on the same $1000$ replicates. The pointwise band is the right tool for a question about one artist picked independently of the data. The other two would be needed to report the three artists a scan returned. At $1000$ replicates the Bonferroni bounds sit at the extreme order statistics, so they bound conservatism without estimating it. In simulation calibrated to this corpus the pointwise band covers 91.5 per cent against a nominal $95$, and an artist whose population gap is non-negative is declared robustly negative 0.07 per cent of the time.}
\label{tab:gap-bands}
\end{table}

When a single artist is examined, what matters is whether the pointwise
interval covers the population gap, and a maximum statistic gives no
guarantee of that at this sample size. On synthetic corpora fitted to
reproduce this corpus's mean within median, mean cross median, mean
worst-cross median and negative-gap fraction, so that the population
gap is known, the nominal $95\%$ interval covers in $91.5\%$ of cases, and in $96.7\%$ among artists
whose true gap lies within $0.01$ of zero. Reading a single artist's
interval as roughly a $90\%$ statement is the right adjustment. The
shortfall is in width and not in direction. An
artist whose population gap is non-negative is declared robustly
negative $0.07\%$ of the time. The classification therefore errs
towards calling artists ambiguous, which is the harmless direction for
a diagnostic whose purpose is to withhold trust.

The resampling redraws works inside the same $89$ people, so the
intervals cover their anchor samples and not a population of artists.
An artist-level bootstrap does not repair that. Resampling the $89$
artists with replacement leaves about $56$ distinct ones. Since every
artist's cross term is a maximum over the others, fewer distinct
competitors mean a systematically larger gap. The artist-level resample
lands at $17.3\%$, to four significant figures the value
that random subsets of $56$ artists of Commons-89 give. So it measures a smaller corpus and not the uncertainty of
this one. With the artist set fixed and no superpopulation model, what
is estimable is how the rate varies over artist subsets of a stated
size, and this is what that curve reports.

The gap compares one location statistic of the within-artist
distribution with the same statistic for the worst competitor, and this
choice is not neutral. Replacing the median by the
mean yields $14$ negative gaps, by a $10\%$ trimmed mean $17$, by the
upper quartile $13$ and by the lower quartile $31$, against $20$ for
the median. The order of magnitude is stable, between $15\%$ and $35\%$
of artists, but the membership of the flagged set is not. Measured
against the median set, the Jaccard overlap ranges from $0.55$ to
$0.68$, so between a third and nearly half of the individual entries
change with the statistic. This is why \S\ref{sec:diagnostic} reports
the rate as the result and the named pairs as illustration, and why a
single artist whose gap sits near zero should be read as provisional
rather than categorically failed.

Comparing two medians also says nothing in itself about whether the two
distributions overlap. For each artist we therefore compute the
probability that a random within-artist pair scores higher than a
random pair against that artist's own worst competitor. This probability
is the Mann-Whitney statistic. On this corpus the gap and that
probability correlate at $\rho = 0.96$ over the $89$ artists, but this
figure is inflated because both are computed on the same anchors.
Recomputing the gap on half of each artist's anchors and the
probability on the other half leaves $\rho = 0.57 \pm 0.17$ across the
$25$ splits of \S\ref{sec:verification}. This is the honest number for
predicting an artist's separability from a gap measured elsewhere.
Among the $20$ flagged artists the probability averages $0.469$ with a
median of $0.485$. None reaches $0.7$, and only four exceed
chance at all. Among the $69$ unflagged artists it averages $0.692$.
Two artists have a positive gap and below-chance separation, Itō Jakuchū
at $+0.006$ and Botticelli at $+0.002$, and both sit inside the boundary
region their CIs already report as undetermined. None of this makes the
gap a verdict on the embedding. The same artists are separated well
above chance from a randomly drawn different artist
(\S\ref{sec:verification}), and what the diagnostic picks out is the
closest competitor, not the average one.

\section{Full CSD+ readout grid and ablations}
\label{app:csdplus_grid}

\paragraph{Full $6 \times 4$ readout grid.}
The compact $4 \times 3$ form in the main text (Tab.~\ref{tab:csdplus})
shows the negative-gap count for the three operational input pipelines
(vanilla 224, multi-crop $5\times 224$, pos-interp 336) crossed with the
four score-level variants. Tab.~\ref{tab:csdplus-full} below adds the
two intermediate pos-interp resolutions (280, 392) and the high
resolution (448) that the scale ablation explores, plus the per-cell
median gap.

\begin{table}[h]
\centering
\small
\setlength{\tabcolsep}{4.5pt}
\begin{tabular}{l rr rr rr rr}
\toprule
& \multicolumn{2}{c}{cosine} & \multicolumn{2}{c}{CSLS} & \multicolumn{2}{c}{cohort-$z$} & \multicolumn{2}{c}{margin} \\
\cmidrule(lr){2-3}\cmidrule(lr){4-5}\cmidrule(lr){6-7}\cmidrule(lr){8-9}
input variant & \#neg & med & \#neg & med & \#neg & med & \#neg & med \\
\midrule
vanilla 224     & 12 & $+0.072$ & 4 & $+0.199$ & 12 & $+0.516$ & 4 & $+0.123$ \\
5-crop avg 224  & 10 & $+0.079$ & 4 & $+0.202$ & 10 & $+0.470$ & 4 & $+0.131$ \\
pos-interp 280  &  8 & $+0.082$ & 3 & $+0.198$ &  8 & $+0.517$ & 2 & $+0.136$ \\
pos-interp 336  &  9 & $+0.081$ & 4 & $+0.201$ &  9 & $+0.559$ & 3 & $+0.134$ \\
pos-interp 392  &  7 & $+0.079$ & 4 & $+0.194$ &  7 & $+0.541$ & 3 & $+0.120$ \\
pos-interp 448  & 11 & $+0.076$ & 4 & $+0.187$ & 11 & $+0.531$ & 4 & $+0.114$ \\
\bottomrule
\end{tabular}
\caption{Gaps of the CSLS score are on its unhalved scale, which is twice the common scale of \S\ref{sec:density}. Full CSD+ grid: every input-pipeline variant crossed with every
score-level variant on the discrimination diagnostic. \#neg is the count,
out of $89$ artists, with negative aggregated gap; med is the median gap
over artists. The compact main-text form (Tab.~\ref{tab:csdplus}) keeps
three input pipelines; pos-interp $280$/$392$/$448$ add only marginal
variation under CSLS.}
\label{tab:csdplus-full}
\end{table}

\paragraph{Scale ablation.}
Bilinear interpolation of the trained positional embeddings to a $g \times g$
grid permits inference at $14g \times 14g$ on the same backbone weights.
Tab.~\ref{tab:scale-ablation} reports the diagnostic and verification AUC
at five grid sizes, and the two quantities do not behave alike. Verification
AUC is single-peaked: it rises from $0.905$ at $224$ to $0.912$ at $280$,
stays there at $336$ and falls away to $0.901$ at $448$. The negative-gap
count does not follow a curve at all. Under CSLS it is $4$, $3$, $4$, $4$,
$4$, and under raw cosine $12$, $8$, $9$, $7$, $11$. So the middle three
resolutions are indistinguishable on that axis, and only the two ends are
clearly worse. Most of the available gain is therefore reached at
$280$. Between $280$ and $392$ the differences are within run-to-run
fluctuation, and $448$ is worse on every measure.

Between $280$ and $336$ the marginal standard deviations are equal
($0.013$ each), and the paired comparison across the same $25$ splits
does not separate them either. Under CSLS, $280$ leads in $20$ of $25$
splits by $+0.0012$; under raw cosine, $336$ leads in $16$ of $25$ by
$+0.0011$. Neither margin reaches a tenth of the split-to-split spread. We
report pos-interp at $336$ because the LoRA test and the audit of the
unmodified generator (\S\ref{sec:generated}, App.~\ref{app:t2i_extras})
were run at that setting. The size $280$ is equally defensible, and its embedding cost is
$1.6\times$ that of vanilla, compared with $2.25\times$ for $336$. Beyond $336$ the positional
grid moves far enough from its trained $16 \times 16$ size that every
measure degrades.

\begin{table}[h]
\centering\small
\begin{tabular}{rrrrr}
\toprule
input size & \# neg gap (cos) & cos AUC & \# neg gap (CSLS) & CSLS AUC \\
\midrule
$224$ (vanilla) & $12$ & $0.891 \pm 0.015$ & $4$ & $0.905 \pm 0.015$ \\
$280$           & $8$  & $0.903 \pm 0.013$ & $3$ & $\mathbf{0.912 \pm 0.013}$ \\
$336$ (recipe)  & $9$  & $0.904 \pm 0.013$ & $4$ & $0.911 \pm 0.013$ \\
$392$           & $7$  & $0.900 \pm 0.014$ & $4$ & $0.907 \pm 0.014$ \\
$448$           & $11$ & $0.894 \pm 0.015$ & $4$ & $0.901 \pm 0.015$ \\
\bottomrule
\end{tabular}
\caption{Scale ablation for the pos-interp input variant, on the
anchor set of Commons-89 in \S\ref{sec:corpus}. AUC as mean $\pm$ std over
the same 25 artist-disjoint splits as main-text
Tab.~\ref{tab:robustness}; the vanilla 224 row matches that table by
construction.}
\label{tab:scale-ablation}
\end{table}

\paragraph{Multi-crop aggregation versus pos-interp.}
Multi-crop aggregation replaces the single centre crop with five
$224 \times 224$ crops sampled from a $256 \times 256$ base, namely the
four corners plus the centre. Each crop is embedded independently, and the
embeddings are averaged. This recovers some of the spatial coverage that
the single centre crop loses. If one reads the $K$-crop average as a
deterministic five-crop estimator of the crop-distribution mean
$z = \mathbb{E}_c[\phi(x_c)]$, it has variance
$\mathrm{Var}(\hat z) = \tfrac{\sigma^2}{K}\bigl[1 + (K{-}1)\,\bar\rho\bigr]$,
where $\sigma^2$ is the per-crop variance under crop position and $\bar\rho$
is the average between-crop correlation. The five-crop construction draws
all crops from a $256 \times 256$ base. So the crops overlap substantially,
and $\bar\rho$ is large. Therefore the bracketed factor stays well
above the $1/K$ ideal that independent crops could reach, and the variance
reduction is real but much weaker. More importantly, multi-crop is a
\emph{variance} reduction at fixed \emph{bias}. Every crop still discards
the image content outside the $256 \times 256$ base, so any structural
error from truncating peripheral pixels is shared across all $K$ samples
and survives the average.
Pos-interp $336$ does not remove that bias either, and it is worth being
exact about what it does. The preprocessing resizes the shortest side to
the input size and then takes a square centre crop. Hence the visible
fraction of the artwork is identical at $224$, $336$ and $448$; for a
$2{:}1$ work it is $40\%$ of its width at every setting. Only the sampling
density inside that fixed field changes. Pos-interp therefore addresses a
different error than multi-crop does. It addresses resolution, while
multi-crop addresses crop position, and neither addresses the
content that the square crop discards. Recovering that
would require a non-square or multi-view input protocol, which we do not
test.

\paragraph{Full verification-robustness statistics.}
Tab.~\ref{tab:robustness-full} expands main-text Tab.~\ref{tab:robustness} with the
AUC range, EER and full paired-comparison statistics that the prose of
\S\ref{sec:verification} relies on but that the slim inline table
omits.

\begin{table}[h]
\centering\small
\begin{tabular}{lrrr}
\toprule
Method & AUC mean $\pm$ std & AUC range & EER mean $\pm$ std \\
\midrule
Cosine on vanilla CSD (224)         & $0.891 \pm 0.015$ & $[0.861, 0.932]$ & $0.190 \pm 0.016$ \\
CSLS on vanilla CSD (224)           & $0.905 \pm 0.015$ & $[0.871, 0.937]$ & $0.176 \pm 0.016$ \\
Cosine on pos-interp 336   & $0.904 \pm 0.013$ & $[0.872, 0.933]$ & $0.175 \pm 0.015$ \\
CSLS on pos-interp 336 (CSD+)  & $0.911 \pm 0.013$ & $[0.876, 0.936]$ & $0.171 \pm 0.014$ \\
Pair-logreg on vanilla CSD (224)    & $0.923 \pm 0.013$ & $[0.887, 0.950]$ & $0.159 \pm 0.017$ \\
Pair-logreg on pos-interp 336       & $0.923 \pm 0.012$ & $[0.894, 0.944]$ & $0.159 \pm 0.015$ \\
\bottomrule
\end{tabular}

\smallskip
\centering\small
\begin{tabular}{llrr}
\toprule
Left method & Right method & Left wins / 25 & mean $\Delta$AUC \\
\midrule
Pair-logreg vanilla & Cosine vanilla & 25 & $+0.032$ \\
CSLS vanilla & Cosine vanilla & 25 & $+0.014$ \\
Cosine pos-interp 336 & Cosine vanilla & 25 & $+0.013$ \\
CSLS pos-interp 336 & Cosine vanilla & 25 & $+0.020$ \\
Pair-logreg pos-interp 336 & Cosine vanilla & 25 & $+0.032$ \\
Pair-logreg pos-interp 336 & Pair-logreg vanilla & 14 & $+0.000$ \\
CSLS pos-interp 336 & CSLS vanilla & 23 & $+0.006$ \\
CSLS pos-interp 336 & Pair-logreg vanilla & 0 & $-0.012$ \\
\bottomrule
\end{tabular}
\caption{Per-method statistics for the verification robustness study
(Section~\ref{sec:verification}, cleaned corpus, $25$ artist-disjoint splits).
Top: mean$\pm$std AUC, AUC range, and EER. Bottom: paired-split
comparisons (left wins / 25, mean $\Delta$AUC).}
\label{tab:robustness-full}
\end{table}

Tab.~\ref{tab:uncertainty-designs} reports the headline improvements of
\S\ref{sec:verification} under four uncertainty designs: the naive
independent-splits interval, the overlap-corrected interval of
\citet{nadeau2003inference} (factor $2.77$ in width), the artist-cluster
bootstrap that resamples the $89$ artists, and the stricter per-artist
improvement. No design changes a sign or brings an interval near zero.

\begin{table}[t]
\centering\small
\setlength{\tabcolsep}{4pt}
\begin{tabular}{lrrrrrrr}
\toprule
& & \multicolumn{3}{c}{$95\%$ half-width} & \multicolumn{3}{c}{per artist} \\
\cmidrule(lr){3-5}\cmidrule(lr){6-8}
improvement over raw cosine & $\Delta$ & naive & N--B & boot & $\Delta$ & $\pm$ & helped \\
\midrule
CSLS on vanilla $224$ & $+0.0142$ & $0.0021$ & $0.0058$ & $0.0037$ & $+0.0104$ & $0.0031$ & $66/89$ \\
pos-interp $336$ & $+0.0130$ & $0.0024$ & $0.0066$ & $0.0040$ & $+0.0103$ & $0.0038$ & $64/89$ \\
both combined & $+0.0200$ & $0.0027$ & $0.0073$ & $0.0046$ & $+0.0149$ & $0.0047$ & $70/89$ \\
pair-logreg on vanilla $224$ & $+0.0319$ & $0.0029$ & $0.0080$ & $0.0060$ & $+0.0236$ & $0.0060$ & $72/89$ \\
\bottomrule
\end{tabular}
\caption{The same paired improvements under four uncertainty designs, given as half-widths so that the cost of each assumption is directly comparable. The naive half-width assumes $25$ independent replications. The splits are not independent, because they are drawn from the same $89$ artists and two splits share about $3.2$ test artists in expectation. Nadeau--Bengio corrects for exactly that overlap and is the most conservative. The artist-cluster bootstrap moves the estimand from the split sample to the artist sample and lands between the two. The per-artist block asks a different and stricter question, namely what an individual artist gains. It is smaller because the gain is not uniform; the last column counts artists with a positive difference. No design changes a sign or brings an interval near zero.}
\label{tab:uncertainty-designs}
\end{table}

\paragraph{Hard-negative regimes.}
Tab.~\ref{tab:hard-negatives} expands the hard-negative discussion of
\S\ref{sec:verification} with the full four regimes
(random, same-movement, worst-neighbour, positive-gap control) crossed
with the three readouts (cosine vanilla 224, CSLS on pos-interp 336,
pair-feature logistic regression on pos-interp 336). The
worst-neighbour regime is the within-tradition stress test that
\S\ref{sec:diagnostic} predicts to be hard. It lowers AUC by about
$0.18$ absolute on cosine vanilla, and CSLS narrows the gap by the same
fraction as in the random regime. Same-movement negatives lower AUC
almost to the worst-neighbour level, with about twice its spread across
splits. The positive-gap control is consistent with the expectation that
artists with a clean diagnostic, that is $g_k > 0.05$, reach AUC above
$0.92$ across all readouts.

\begin{table}[h]
\centering\small
\setlength{\tabcolsep}{5pt}
\begin{tabular}{lrrrr}
\toprule
& random & same-movement & worst-neighbour & pos-gap control \\
\midrule
Cosine (224)        & $0.890 \pm 0.013$ & $0.746 \pm 0.058$ & $0.709 \pm 0.026$ & $0.932 \pm 0.015$ \\
CSLS, 336           & $0.910 \pm 0.013$ & $0.763 \pm 0.051$ & $0.743 \pm 0.028$ & $0.949 \pm 0.013$ \\
Pair-logreg, 336     & $0.922 \pm 0.013$ & $0.756 \pm 0.055$ & $0.755 \pm 0.033$ & $0.957 \pm 0.012$ \\
\bottomrule
\end{tabular}
\caption{Verification AUC under four negative-sampling regimes
(worst-neighbour: $B=\arg\max_j c_{k,j}$; positive-gap control:
$g_k > 0.05$), same 25 artist-disjoint splits and corpus as
Tab.~\ref{tab:robustness}. Same-movement is high-variance and defined on $22$ of the $25$ splits,
because few test artists share a movement label.}
\label{tab:hard-negatives}
\end{table}

Tab.~\ref{tab:verification-crosscorpus} gives the same regimes for the
cosine and the CSLS score on all three corpora.

\begin{table}[t]
\centering\small
\setlength{\tabcolsep}{4pt}
\resizebox{\linewidth}{!}{%
\begin{tabular}{lcccc}
\toprule
corpus & random & same-movement & worst-neighbour & positive-gap ctrl \\
\midrule
Commons-89 & $0.890$ / $0.905$ ($+0.015$) & $0.746$ / $0.754$ ($+0.008$) & $0.709$ / $0.737$ ($+0.028$) & $0.932$ / $0.945$ ($+0.013$) \\
WikiArt dump & $0.866$ / $0.868$ ($+0.002$) & $0.780$ / $0.788$ ($+0.008$) & $0.681$ / $0.700$ ($+0.019$) & $0.916$ / $0.917$ ($+0.002$) \\
ArtBench-10 & $0.900$ / $0.908$ ($+0.008$) & $0.776$ / $0.799$ ($+0.023$) & $0.590$ / $0.621$ ($+0.031$) & $0.958$ / $0.960$ ($+0.002$) \\
\bottomrule
\end{tabular}%
}
\caption{Cosine / CSLS verification AUC and their difference, by negative-sampling regime, on the same 25-split protocol across all three corpora. On every corpus CSLS is applied to the same embeddings as the cosine column, so the column isolates the readout. The input-resolution lever is reported separately in Tab.~\ref{tab:robustness}. The CSLS gain is largest under worst-neighbour negatives on every corpus. This is the regime the diagnostic of \S\ref{sec:diagnostic} flags; a shared movement label is a coarser proxy for that competitor. The gain does not disappear outside that regime. Under paired intervals corrected for the overlap between splits, the random-regime gain is an established gain in eleven of the twelve corpus/backbone cells. Only CSD on WikiArt is negligible against a $0.01$ margin, at $+0.002\,[-0.002, +0.006]$ (\S\ref{sec:verification}). That single cell, the least hubness backbone on the broadest corpus, is the joint minimum of the two effects and not the general case. Pair-logreg columns and per-split spreads are in the result files.}
\label{tab:verification-crosscorpus}
\end{table}

\paragraph{Pair-feature ablation across 25 splits.}
Pair-feature logistic regression on
$\phi(a, b) = [\,|a-b|, a \odot b, (a+b)/2\,]$ reaches AUC $0.923 \pm 0.013$
on raw 768-d CSD across 25 splits. Tab.~\ref{tab:feature-ablation} takes the
three blocks apart. The elementwise difference reaches $0.915$ on its own,
within $0.008$ of the full feature set, while the elementwise product reaches
$0.892$, essentially the unsupervised cosine baseline, and adds nothing when
combined with the difference. The pair location $(a+b)/2$ scores $0.498$ on its
own, which is chance level here. Where a pair sits says little by itself
about whether its two images match. Added to the other two blocks, it is
worth the remaining $0.008$, because it lets the classifier condition the
difference coordinates on where in the space a pair sits. The gain of the
supervised head over cosine therefore comes from a reweighting of the
$768$ difference coordinates and not from a new kind of comparison.

\begin{table}[h]
\centering\small
\begin{tabular}{lrrr}
\toprule
feature blocks & dim & AUC & range \\
\midrule
$\cos(a,b)$ alone (unsupervised baseline) & $1$ & $0.891 \pm 0.015$ & $[0.861, 0.932]$ \\
$(a+b)/2$ only & $768$ & $0.498 \pm 0.010$ & $[0.484, 0.518]$ \\
$a \odot b$ only & $768$ & $0.892 \pm 0.015$ & $[0.863, 0.934]$ \\
$|a-b|$ only & $768$ & $0.915 \pm 0.014$ & $[0.883, 0.950]$ \\
$|a-b| \,\Vert\, a \odot b$ & $1536$ & $0.915 \pm 0.014$ & $[0.883, 0.950]$ \\
all three blocks & $2304$ & $0.923 \pm 0.013$ & $[0.887, 0.950]$ \\
all three blocks $+\ \cos(a,b)$ & $2305$ & $0.924 \pm 0.013$ & $[0.889, 0.950]$ \\
\bottomrule
\end{tabular}
\caption{Pair-feature block ablation for the supervised readout, across the same 25 artist-disjoint splits as Tab.~\ref{tab:robustness}. The elementwise difference carries the discrimination on its own, and the elementwise product adds nothing beyond it. For this reason we call $|a-b|$ the dominant block. The pair location $(a+b)/2$ is at chance on its own in this experiment and helps only in combination with the other blocks. Adding the raw cosine to the full feature set changes the mean by $0.001$, so the classifier is not merely relearning it.}
\label{tab:feature-ablation}
\end{table}

\paragraph{Embedding-postproc baselines.}
Embedding-level recipes, fitted on the $72$ training artists per split,
do not substitute for the score-level readout. Centering raises cosine
verification AUC by about $+0.02$ but leaves the aggregated
negative-gap count unchanged. All-but-the-top~\citep{mu2018allbutthetop}
and PCA whitening lose AUC on both readouts. Transductive variants
do not generalise to held-out artists, which is the same pattern that
App.~\ref{app:geometry} documents for UMAP-projected features.

\paragraph{What the density term corrects for.}
Tab.~\ref{tab:csls-density-decomposition} decomposes the CSLS reduction of
\S\ref{sec:density} by reference-pool construction, and reports the
same-artist share of the neighbourhood across $k$.

\begin{table}[t]
\centering\small
\begin{tabular}{lrr}
\toprule
reference pool for $r(x)$ & mean $r(x)$ & \# neg gap \\
\midrule
raw cosine, no density term & --- & $12/89$ \\
\midrule
same pool (operational readout) & $0.611$ & $4/89$ \\
own class excluded from $r(x)$ & $0.560$ & $9/89$ \\
own class excluded, pool balanced & $0.532$ & $8/89$ \\
\bottomrule
\end{tabular}
\caption{What the CSLS density term corrects for, in the aggregated-pool regime at $k = 15$. The diagnostic's reference pool is the evaluation corpus, so the neighbours entering $r(x)$ include other works by the query's own artist. The first row therefore mixes within-artist cohesion with cross-artist local density. Excluding the query's class isolates the cross-artist component, and balancing the pool also removes per-artist pool-size effects. Sweeping $k$ shows the same mechanism in the same-artist share of the neighbourhood (share / \# neg gap): $k{=}5$: $56.4\%$ / $4$; $k{=}15$: $39.7\%$ / $4$; $k{=}30$: $27.1\%$ / $7$; $k{=}50$: $19.3\%$ / $6$. The first row is the operational readout, because a query's artist is what the score has to determine and cannot be excluded in advance. The other two serve the decomposition. In the inductive verification protocol of \S\ref{sec:verification} the share is zero by construction.}
\label{tab:csls-density-decomposition}
\end{table}

\paragraph{CSLS k-sensitivity.}
The CSLS readout has one hyperparameter, the neighbourhood size $k$ at
which local densities are estimated; we use $k = 15$ throughout.
Tab.~\ref{tab:csls_k_sensitivity} sweeps $k \in \{5, 10, 15, 20, 50\}$
on both input variants, and the counts move by at most two flags across
the sweep. The variation is small, but it is not noise. As $k$ grows, the
same-artist share of the neighbourhood declines
(Tab.~\ref{tab:csls-density-decomposition}), and this moves the count
towards the value obtained when the query's own class is excluded
outright. So the choice of $k$ trades one component of the density term
against the other and does not add or remove signal. The main-text
headline ``$12/89 \to 4/89$ under CSLS'' is the $k=15$ row of the
vanilla-$224$ block.

\begin{table}[h]
\centering
\small
\setlength{\tabcolsep}{5pt}
\begin{tabular}{r rrr rrr}
\toprule
& \multicolumn{3}{c}{vanilla 224 (default)} & \multicolumn{3}{c}{pos-interp 336} \\
\cmidrule(lr){2-4}\cmidrule(lr){5-7}
$k$ & agg. & pairw. & med.\ gap & agg. & pairw. & med.\ gap \\
\midrule
5  & 4 & 8  & $+0.171$ & 3 & 6 & $+0.179$ \\
10 & 4 & 9  & $+0.174$ & 4 & 5 & $+0.179$ \\
\textbf{15} & \textbf{4} & \textbf{8} & $\mathbf{+0.168}$ & \textbf{4} & \textbf{5} & $\mathbf{+0.173}$ \\
20 & 5 & 9  & $+0.161$ & 3 & 5 & $+0.170$ \\
50 & 6 & 10 & $+0.156$ & 3 & 7 & $+0.173$ \\
\bottomrule
\end{tabular}
\caption{Gaps of the CSLS score are on its unhalved scale, which is twice the common scale of \S\ref{sec:density}. Sensitivity of the CSLS readout to the neighbourhood size $k$ on
both input variants: artists with negative gap in the aggregated-pool
regime (agg.) and at pairwise level (pairw.), and the median pairwise
gap. The default $k=15$ row is set in bold.}
\label{tab:csls_k_sensitivity}
\end{table}

\paragraph{CSLS runtime cost.}
The readout needs no additional encoder forward pass, but it is not
free. Each query needs one $k$ nearest-neighbour search against the
reference gallery, and the gallery has to be held in memory. At $768$
dimensions in single precision, this takes $5$~MB for our $1645$ anchors,
$121$~MB for the WikiArt dump and $178$~MB for ArtBench. Exact blocked
search on one CPU core costs $17$, $362$ and $543$~microseconds per query
against those three galleries. So the cost is linear in gallery size and,
at the scale at which a style-fidelity evaluation works, negligible
compared with a single diffusion sampling step. A tree index does not help
at these dimensions. \texttt{sklearn}'s
 neighbour search on the same
data is slower than the blocked product while agreeing with it to
$3 \times 10^{-7}$, as expected for $768$ dimensions.

\section{Manifold geometry, clustering and cluster compositions}
\label{app:geometry}

\paragraph{Intrinsic dimensionality.}
We make no claim about intrinsic dimension, and the reason is worth one
paragraph. The two-nearest-neighbour estimator of
\citet{facco2017intrinsic} returns $\mathrm{ID} = 3.14 \pm 0.11$ on
subsamples of Commons-89. However, the value halves or doubles with the
distance convention, with cosine $3.08$ against Euclidean $6.17$. A
different estimator returns $13.94$ \citep{levina2004maximum}, and the
value falls with sample size. Restricting to one work per artist raises it
from $4.90 \pm 0.82$ to $8.69 \pm 1.06$, so much of the small value is the
tightness of artist clusters rather than the dimension of anything. What survives needs no estimator. The top three principal
components explain $7.5\%$, $5.9\%$ and $5.2\%$ of variance, $18.6\%$
together. So the variance is spread thinly across many directions, and a
projection onto a handful of linear axes will not capture the geometry.

\paragraph{Full clustering sweep.}
Tab.~\ref{tab:clustering-sweep} reports adjusted Rand index and normalised
mutual information against artist labels and against art-movement labels for
$17$ configurations spanning nine projection/clusterer families. We report ARI and NMI rather than purity because
they correct for chance agreement at the relevant number of clusters, which
varies across methods. The methods cover three approaches to the same
underlying question: cluster artists in CSD's output space.

\begin{table}[t]
\centering
\small
\begin{tabular}{lrrrrrr}
\toprule
Method & \#cl & \%noise & ARI(art) & NMI(art) & ARI(mov) & NMI(mov) \\
\midrule
RandProj-100 + avg-link, K=120 & 120 & 0.0\% & +0.234 & +0.668 & +0.032 & +0.192 \\
Spectral-20D + Leiden, $r$=1.5 & 27 & 0.0\% & +0.203 & +0.605 & +0.033 & +0.138 \\
Spectral-20D + Leiden, $r$=1.0 & 24 & 0.0\% & +0.187 & +0.595 & +0.035 & +0.133 \\
Correlation-kNN + Leiden, $r$=1.5 & 18 & 0.0\% & +0.157 & +0.578 & +0.043 & +0.134 \\
UMAP-20D + HDBSCAN & 43 & 12.2\% & +0.150 & +0.636 & +0.033 & +0.203 \\
RandProj-100 + avg-link, K=80 & 80 & 0.0\% & +0.149 & +0.601 & +0.027 & +0.143 \\
Spectral-20D + Leiden, $r$=0.5 & 17 & 0.0\% & +0.134 & +0.557 & +0.031 & +0.101 \\
Cosine-kNN + Leiden, $r$=1.5 & 16 & 0.0\% & +0.130 & +0.555 & +0.043 & +0.131 \\
UMAP-50D + HDBSCAN & 36 & 18.0\% & +0.118 & +0.603 & +0.037 & +0.192 \\
RandProj-100 + avg-link, K=40 & 40 & 0.0\% & +0.108 & +0.515 & +0.027 & +0.095 \\
Correlation-kNN + Leiden, $r$=1.0 & 13 & 0.0\% & +0.108 & +0.528 & +0.040 & +0.111 \\
Cosine-kNN + Leiden, $r$=1.0 & 13 & 0.0\% & +0.107 & +0.522 & +0.039 & +0.106 \\
Cosine-kNN + Leiden, $r$=0.5 & 7 & 0.0\% & +0.081 & +0.463 & +0.034 & +0.059 \\
Correlation-kNN + Leiden, $r$=0.5 & 7 & 0.0\% & +0.080 & +0.466 & +0.029 & +0.053 \\
Spectral-20D + HDBSCAN & 33 & 34.1\% & +0.059 & +0.538 & +0.020 & +0.176 \\
RandProj-100 + HDBSCAN & 14 & 74.2\% & +0.008 & +0.289 & +0.009 & +0.124 \\
NMF-20D on $|X|$ + HDBSCAN & 2 & 27.8\% & +0.004 & +0.072 & -0.000 & +0.016 \\
\bottomrule
\end{tabular}
\caption{Clustering pipelines compared on Commons-89. ARI and NMI are computed on the full corpus against artist labels (1645 points, 89 classes) and against art-movement labels (45 movements drawn from the corpus-author vocabulary), treating HDBSCAN's $-1$ noise label as a single class so high-noise methods do not get scored on the easy interior alone. $r$ denotes the Leiden resolution parameter. RandProj-100~+~avg-link at $K{=}120$ is strongest by global ARI(art), at a granularity close to one cluster per artist. Among methods that choose their own cluster count, Spectral-20D~+~Leiden leads. UMAP-20D~+~HDBSCAN is the more interpretable pipeline because the noise label flags marginal works explicitly.}
\label{tab:clustering-sweep}
\end{table}

\paragraph{ARI computation note.} HDBSCAN-style methods produce a $-1$ ``noise''
label for points that do not belong to any high-density region. We compute
ARI on the \emph{full} corpus, treating the $-1$ label as a single class of
its own. This is the fair cross-method comparison. A method that labels
half the corpus as noise pays for that decision in the ARI formula, because
the noise points form one giant cluster against which the artist labels
score poorly. Scoring the non-noise subset alone would instead inflate
high-noise methods. The most visible case is RandProj-100 + HDBSCAN, which
labels $74.2\%$ of the corpus as noise and would otherwise be judged only
on the easy interior it retains.

The pattern in the table is more nuanced than ``linear methods fail,
non-linear methods work''. Among pipelines that choose their own cluster count, the strongest is
graph-based on a non-linear projection (Spectral-20D + Leiden, $r=1.5$,
ARI $0.203$ at the canonical seed of Tab.~\ref{tab:clustering-sweep}).
Density clustering on a linear projection fails completely
(RandProj-100 + HDBSCAN at ARI $0.008$, NMF + HDBSCAN at $0.004$). However, \emph{hierarchical} clustering on a linear
projection with a generous fixed cluster budget is the strongest overall.
RandProj-100 + average-linkage at $K{=}120$ reaches ARI $0.234$ at a
granularity close to one cluster per artist, while UMAP-20D + HDBSCAN
reaches $0.150$. The structural distinction is therefore not linear versus
non-linear projection alone. It is whether the chosen clusterer can
tolerate the geometry it is handed. Random projection to $100$ dimensions
preserves pairwise distances to within $18\%$ at the first and
ninety-ninth percentiles, which is enough for hierarchical merging.
HDBSCAN's mutual-reachability construction, however, depends on the fine
structure of local density, which the projection does damage. We report that failure as observed; its mechanism is not
established.

Across ten random seeds, Spectral-20D + Leiden res $=1.5$ sits at
full-corpus ARI $0.204 \pm 0.001$, with low across-seed variability.
The $0.203$ in Tab.~\ref{tab:clustering-sweep} is the canonical-seed value
and lies within one standard deviation of the ten-seed mean.
UMAP-20D + HDBSCAN follows at $0.145 \pm 0.009$, with a mean cluster count
of $47.5$ and a $16\%$ noise rate, and UMAP-50D + HDBSCAN is comparable
at $0.141 \pm 0.007$. We treat both UMAP-20D + HDBSCAN and Spectral-20D +
Leiden as defensible recipes for art-historical discovery. The former is
easier to interpret cluster by cluster, because HDBSCAN's noise label flags
marginal works honestly. The latter is stronger on the global ARI metric
and assigns every point.

\paragraph{Cluster compositions.}
Tab.~\ref{tab:cluster-comp} reports the $12$ largest UMAP-20D +
HDBSCAN clusters at the canonical seed (\texttt{random\_state=42}).
``Purity'' denotes the fraction of points in a cluster whose movement
label matches the cluster's majority movement.

\begin{table}[t]
\centering
\footnotesize
\setlength{\tabcolsep}{4pt}
\begin{tabularx}{\linewidth}{@{}rrrlX@{}}
\toprule
Cluster & $n$ & Purity & Majority movement & Top-5 artists \\
\midrule
18 & 310 & 0.28 & Hudson River School & Spitzweg (23), Cole (23), Bierstadt (22), Aivazovsky (19), Church (18) \\
33 & 112 & 0.39 & Early Renaissance & Mantegna (23), Bosch (17), Fra Angelico (13), van Eyck (13), Arcimboldo (12) \\
31 & 70 & 0.44 & Post-Impressionism & Seurat (18), Monet (15), Van Gogh (13), Baluschek (6), Klimt (5) \\
9 & 66 & 0.67 & Cubism & Gris (26), Picabia (20), Gleizes (17), Rivera (2), Delaunay (1) \\
32 & 60 & 0.40 & Spanish Baroque & La Tour (17), Vel\'azquez (9), Zurbar\'an (9), Vermeer (8), Murillo (6) \\
2 & 51 & 0.55 & Romanticism & Goya (24), Delacroix (6), Piranesi (5), Dor\'e (2), Turner (2) \\
37 & 44 & 0.45 & American Realism & Hopper (11), Whistler (8), Eakins (7), Repin (3), Sargent (3) \\
38 & 40 & 0.95 & Pre-Raphaelite & Rossetti (18), De Morgan (10), Waterhouse (10), Leighton (1), Grimshaw (1) \\
15 & 38 & 0.50 & Mexican popular print & Posada (19), Bilibin (8), Dor\'e (6), Dulac (2), Aublet (1) \\
5 & 34 & 0.94 & Edo-period Japan & Utamaro (20), Hiroshige (12), Delacroix (2) \\
35 & 34 & 0.53 & Spanish Baroque & Murillo (18), Rubens (6), Leighton (3), Dor\'e (2), Titian (2) \\
29 & 30 & 0.43 & Russian Realism & Levitan (6), Sargent (5), Aublet (4), Kuindzhi (3), Leighton (2) \\
\bottomrule
\end{tabularx}
\caption{Top-12 non-noise HDBSCAN clusters on UMAP-20D, sorted by size.
Purity is the fraction of points whose movement label matches the
cluster's majority movement; the surrounding text reads the clusters
against the negative-gap pairs of \S\ref{sec:diagnostic}.}
\label{tab:cluster-comp}
\end{table}

A striking cluster is number $9$, a $66$-point grouping with $67\%$
Cubism purity composed of Juan Gris ($26$), Francis Picabia ($20$) and
Albert Gleizes ($17$), the early-twentieth-century cohort where
pictorial conventions transfer across signatures. The pipeline finds
this stylistic family without ever seeing the movement labels. Cluster
$33$ (Early Renaissance, $112$ points, $39\%$ purity) groups Andrea
Mantegna ($23$), Hieronymus Bosch ($17$), Fra Angelico ($13$), Jan van
Eyck ($13$), Botticelli and Raffaello, including the
Mantegna/Raffaello pair that the diagnostic still flags at gap $-0.009$
after the attribution cleanup. Cluster $5$
(Edo Japan, $34$ points, $94\%$ purity) merges Kitagawa Utamaro ($20$)
with Utagawa Hiroshige ($12$). The East-Asian brush painters Shen Zhou,
Kawanabe Kyōsai, Hasegawa Tōhaku and Itō Jakuchū scatter
over two small ink clusters and the noise label. These are the same
artists that \S\ref{sec:diagnostic} flags as mutually confused. Cluster $32$
(Spanish Baroque / Dutch Golden Age, $60$ points, $40\%$ purity) groups
Georges de La Tour ($17$), Diego Vel\'azquez ($9$), Francisco de
Zurbar\'an ($9$), Johannes Vermeer ($8$) and Bartolom\'e Esteban
Murillo ($6$) into a single tenebrism cluster, recovering the
Vel\'azquez/La Tour pair that \S\ref{sec:diagnostic} also flags.
Cluster $38$
(Pre-Raphaelite, $40$ points, $95\%$ purity) groups Dante Gabriel
Rossetti ($18$), Evelyn De Morgan ($10$) and John William Waterhouse
($10$). Godward's academic nudes, produced roughly a generation later, now
hold a cluster of their own. The
largest cluster, number $18$ with $310$ points and $28\%$ movement
purity, is the landscape family. Spitzweg, Cole, Bierstadt,
Aivazovsky, Church, Inness, Shishkin and Gifford share one cluster
across four countries, and the pipeline makes no
attempt to separate the individual artists within it. This mirrors the
Bierstadt/Inness and Levitan/Shishkin flags of
\S\ref{sec:diagnostic}. Two further clusters group the graphic arts
across traditions. One is a Goya-centred works-on-paper cluster of $51$
points that absorbs Delacroix and Piranesi. The other is a popular-print
cluster of $38$ points with Posada ($19$), Bilibin ($8$) and Dor\'e ($6$).
In them, the technique binding of \S\ref{sec:tradition} shows up as
geometry.
Because the corpus construction crops mounts, margins and book folds
out of the anchor photographs (\S\ref{sec:corpus}), these clusters
reflect the printed image itself rather than shared photographic
context. The limit on
artist-level discrimination is in CSD's output itself, not in the
post-processing applied on top of it.

Fig.~\ref{fig:umap} in \S\ref{sec:tradition} shows a two-dimensional
UMAP projection of Commons-89. It underestimates the separation of clusters
that are clean in the $20$-dimensional space in which the clustering
operates, so quantitative statements should be read off
Tab.~\ref{tab:cluster-comp}.

\paragraph{Hub measures around Goya.}
Four measures have to be kept apart when \S\ref{sec:diagnostic} calls Goya
the worst competitor of seven artists. The mean median cross similarity to the rest of the
corpus is the quantity that $\cmed$ takes a maximum of. On it, he ranks
first of $89$ at $0.231$ against $0.221$ for the runner-up. He is the
argmax competitor for seven artists, tied for first with Mantegna and Inness,
against about one expected if the worst-other assignment were uniform.
 On the image-level occurrence count for $15$ neighbours with same-artist
neighbours excluded he ranks tenth, and on reverse nearest-neighbour count
$46$th, which is mid-table. The works that are nearest neighbours of many
other works are by Inness, Vel\'azquez and Whistler and not by Goya. An
artist becomes dangerous for a pooled absolute score by sitting close to
many pools in median terms, not by owning many individual
nearest-neighbour slots.
 On the breadth measure the CSLS readout acts as
intended. Goya's hub-inclusion falls from $0.056$ under raw cosine, rank
$45$, to $0.022$ under CSLS, rank $87$, which is a $61\%$ reduction.

\paragraph{Verification on UMAP-projected features.}
Tab.~\ref{tab:verification-umap} reports the verification AUCs under
manifold-projected features. The numbers come from a single fixed
reference split and not from the 25-split mean of main-text
Tab.~\ref{tab:robustness}, which explains why the cosine-on-raw baseline
reads $0.875$ here. UMAP-20D lowers cosine to $0.793$ and pair-logreg to
$0.810$. UMAP-50D is no better, and concatenating raw and projected
features does not help over the raw vector alone. The drop remains when
the projection is repeated at three seeds, with cosine on UMAP-20D at
$0.792$ to $0.795$ against the raw $0.875$, so it is not one unlucky
embedding. The claim is the narrow one that this configuration
($n_{\text{neighbours}} = 15$, $\mathrm{min\_dist} = 0$, fitted on training
artists only) did not help, not a general failure of manifold-aware
projection.

\begin{table}[h]
\centering
\begin{tabular}{lrrr}
\toprule
Method & AUC & EER & recall @ $p$=0.95 \\
\midrule
Pair-logreg on raw CSD-768D & 0.916 & 0.172 & 0.605 \\
Cosine on raw CSD-768D & 0.875 & 0.201 & 0.534 \\
Pair-logreg on concat[raw 768D, UMAP-20D] & 0.897 & 0.196 & 0.486 \\
Pair-logreg on UMAP-50D only & 0.822 & 0.266 & 0.371 \\
Pair-logreg on UMAP-20D only & 0.810 & 0.263 & 0.141 \\
Cosine on UMAP-20D & 0.793 & 0.289 & 0.258 \\
Cosine on UMAP-50D & 0.794 & 0.283 & 0.343 \\
\bottomrule
\end{tabular}
\caption{Verification with manifold-projected features. UMAP is fit on the 72 training artists only; the 17 test artists are projected via \texttt{umap.transform()}. Numbers are on a single fixed reference split, not the 25-split mean of main-text Tab.~\ref{tab:robustness}. Manifold projection drops verification AUC by $0.08$ (cosine) and $0.11$ (pair-logreg), confirming that UMAP collapses individual-artist discrimination in favour of stylistic-family compactness.}
\label{tab:verification-umap}
\end{table}

\section{External-Corpus Validation}
\label{app:external}

\S\ref{sec:corpus} introduces the two corpora that we did not audit and
\S\ref{sec:diagnostic} reports the counts and the subset curves. This
appendix records the construction details, the full worst-other reading
and the third-party metadata check. Before it is applied externally, the
implementation used here reproduces the numbers of Commons-89 ($20/89$,
$12/89$, $4/89$) exactly, with the same code path, embedding protocol
(CSD ViT-L at $224$, fp32) and readout definitions. The embedder agrees
with the main-text scorer to a cosine similarity of $\geq 0.9999$.

The worst-other identities on the external corpora map onto shared
traditions throughout: Monet against Loiseau and Pissarro against Loiseau
(Impressionist landscape), Braque against Gris (Cubism), Rubens against
Jordaens (Flemish Baroque), Gainsborough against Reynolds (English
portrait school), Hokusai against Kuniyoshi (ukiyo-e), Cranach and
Holbein against Memling (Northern Renaissance). This reading is confirmed
by metadata we did not author. PainterPalette~\citep{painterpalette}
aggregates painter metadata over WikiArt and Wikidata. In it,
$16$ of the $19$ WikiArt pairs share a movement or style label or are
connected by a documented influence, teacher or friendship link (Jordaens
was Rubens's pupil; Caillebotte was influenced by Sisley; Modigliani and
Dal\'i by Picasso). The remaining three lack the relevant metadata fields
and so do not contradict the reading. The same check on Commons-89
covers $16$ of the $20$ negative-gap pairs and confirms $10$. The
four missing ones are the two Edo-period pairs with Sharaku, Raffaello with
Mantegna and Bussiere with Waterhouse, which are absent from PainterPalette.
\paragraph{How much the two external corpora share.}
ArtBench recovers its artist labels from filenames written in WikiArt's own
URL convention. The two corpora therefore overlap, and the question is by
how much. Measuring it on embeddings does not work. ArtBench is distributed as $256 \times
256$ centre-cropped squares, while the WikiArt dump keeps native resolution and
aspect ratio. The same painting is therefore presented differently to the
encoder, and a cosine threshold reports the preprocessing gap instead of the
overlap. A perceptual hash does not have that problem if both sides are first
forced into the same geometry. The hash uses greyscale at $32 \times 32$ and a
2D DCT, and it thresholds the top-left $8\times8$ without the DC term at its
median. Which geometry to force is itself unknown, since ArtBench does not
document whether its squares were squashed or cropped. We therefore hash the
WikiArt side both ways and let the data decide. Cropping yields $8330$ exact
hash collisions against $321$ for squashing, which identifies the convention.
With cropping and a Hamming distance of at most $4$, $12639$ of ArtBench's
$60000$ images, or $21.1\%$, are near-duplicates of a WikiArt image. This is
a lower bound on shared provenance, because a corpus can reproduce the same
painting from a different photograph and so land outside the threshold.

A fifth of one corpus therefore also sits in the other. The two are
externally sourced corpora with overlapping provenance and not independent
replications. The diagnostic itself, however, does not depend on this
overlap. Removing all $12639$ near-duplicates and reapplying the ten-work
floor leaves $47361$ works over $1457$ artists, after $182$ have dropped below
the floor. The negative-gap rate moves from $39.4\%$ to
$39.1\%$, and the worst-other identity is unchanged for $1081$ of the $1457$
artists present in both. The rate rather than the count is the comparable
quantity here because the artist count itself changed, for the reasons
\S\ref{sec:diagnostic} gives. Our reimplementation reproduces the published
$646$ of $1639$ before the removal, so the difference afterwards is
attributable to the removal alone.

The word confirmed needs two qualifications. PainterPalette
aggregates over WikiArt, so for the WikiArt rows the check is external
to us but not to the corpus. It is unambiguously independent only for
the rows of Commons-89, where the images are ours and the labels are
not. In either case the two channels differ. The pairs are flagged from
embeddings with no label input, and the confirmation uses movement and
influence fields that play no part in the flagging. The agreement is
therefore not circular, even where the provenance is shared.
The unconfirmed remainder are pairs with adjacent but distinct movement
labels (Bierstadt/Inness as Romanticism versus Tonalism,
Leighton/Godward as Academic versus Neoclassicism, Aublet/Goya as
Academic versus Romanticism), the graphic-arts pair Bilibin/Posada
(Art Nouveau versus Modernismo), and pairs whose
movement fields are empty there (Hiroshige/Utamaro, both ukiyo-e;
Sargent/Hopper).

The filename-derived artist identities on ArtBench carry a residual
alias risk: $194$ of the $1639$ slugs are single first names
(``Andrea'', ``Boris''), which may merge or truncate identities. The
risk works against the finding and does not produce it. These
slugs show a $23.7\%$ negative rate against $41.5\%$ for full-name
slugs, and excluding them raises the headline fraction slightly
($600/1445$, $41.5\%$, versus $646/1639$, $39.4\%$).

Tab.~\ref{tab:external-validation} gives the number of negative gaps of
the three corpora for the pairwise cosine, the pooled cosine and the pooled
CSLS score side by side.

\begin{table}[t]
\centering
\small
\setlength{\tabcolsep}{5pt}
\begin{tabular}{lrrr}
\toprule
Corpus & pairwise neg.\ (robust) & aggregated neg. & CSLS neg. \\
\midrule
Commons-89           & 20 / 22.5\% (4)    & 12 / 13.5\%  & 4 / 4.5\%    \\
WikiArt dump             & 19 / 14.8\% (11)   & 10 / 7.8\%   & 3 / 2.3\%    \\
ArtBench-10              & 646 / 39.4\% (289) & 479 / 29.2\% & 298 / 18.2\% \\
\bottomrule
\end{tabular}
\caption{Negative-gap counts under the three readouts, same code path
on all corpora; sizes in Tab.~\ref{tab:corpora}. Robust: gap negative in at least $95\%$ of $100$ whole-corpus bootstrap
resamples (duplicate draws excluded), a coarser screen than the
per-artist intervals of \S\ref{sec:diagnostic}; on Commons-89
Kyōsai, Doré, Tōhaku and Bilibin.}
\label{tab:external-validation}
\end{table}

The CSLS readout of \S\ref{sec:density} transfers as well. The aggregated
negative-gap count drops from $10$ to $3$ on WikiArt and from $479$ to
$298$ on ArtBench. The relative reduction is weaker on ArtBench. This is
consistent with \S\ref{sec:tradition}, since with $1639$ artists spread over
ten style classes a larger share of the negatives is genuine tradition
overlap and not a readout artefact.

Because CSLS neighbourhoods are computed over individual works, the
very unequal WikiArt class sizes ($42$--$1889$ works) could in
principle let large classes dominate the density terms. Capping every
artist at $50$ works (seeded) leaves the picture unchanged, with $18$
pairwise, $11$ aggregated and $2$ CSLS negatives against $19$, $10$
and $3$ uncapped. The readouts are therefore not driven by pool-size
imbalance.

Taken together with the subset curves of \S\ref{sec:diagnostic}, the
negative fraction is governed by how densely stylistic neighbourhoods are
sampled and not by corpus size as such. This is the operational reason why
the diagnostic must be run on each candidate evaluation corpus instead of
being transferred from a reference corpus. It is also the reason why the
rate, and not the list of flagged artists, is the quantity that carries over.

Tab.~\ref{tab:category-replication} collects the category-axis
replications of \S\ref{sec:tradition} across corpora and axes: technique
on Commons-89, genre and style on WikiArt and ArtBench.

\begin{table}[t]
\centering\small
\setlength{\tabcolsep}{5pt}
\begin{tabular}{llrrr}
\toprule
corpus & axis & artists affected & of which neg.\ gap & mean $\Delta \wmed$ \\
\midrule
Commons-89 & technique & $32/89$ & $10/20$ & $+0.0420$ \\
WikiArt & genre & $123/128$ & $19/19$ & $+0.0806$ \\
WikiArt & style & $71/128$ & $12/19$ & $+0.0483$ \\
ArtBench-10 & style & $243/1639$ & $113/646$ & $+0.0152$ \\
\bottomrule
\end{tabular}
\caption{Category-confound test across corpora and category axes, each against a permutation null with shuffled labels at fixed group sizes (\S\ref{sec:tradition}). $\Delta \wmed$ is averaged over the negative-gap artists. Counts are per-artist flags at unadjusted $p < 0.05$, a screen rather than a corrected count; under Benjamini--Hochberg the curated row is $22/89$ and $8/20$ (\S\ref{sec:tradition}).}
\label{tab:category-replication}
\end{table}

\section{T2I evaluation: bare-prompt tier classification and pretest notes}
\label{app:t2i_extras}

\paragraph{Bare-prompt tier-classification (full table).}
\S\ref{sec:generated} reports three regimes for bare FLUX.1-dev artist
recognition; Tab.~\ref{tab:bare-trigger} gives the full per-artist counts
out of the $66$ generations per artist (22 neutral subjects $\times$ 3 seeds,
prompted as \texttt{"<movement> <medium> by <artist>, <subject>"}).

\begin{table}[h]
\centering\small
\begin{tabular}{llrr}
\toprule
Tier & Artist & Top-1 / 66 & Top-5 / 66 \\
\midrule
1 (trigger works) & Georges de La Tour      & 32 (48\%) & 43 (65\%) \\
                   & Tōshūsai Sharaku    & 17 (26\%) & 66 (100\%) \\
\midrule
2 (tradition,      & Frederic Edwin Church   & 17 (26\%) & 57 (86\%) \\
not artist)        & Kawanabe Kyōsai        &  0 ( 0\%) & 42 (64\%) \\
                   & Itō Jakuchū         &  8 (12\%) & 47 (71\%) \\
                   & Sanford R.\ Gifford     &  6 ( 9\%) & 44 (67\%) \\
                   & Ivan Shishkin           &  5 ( 8\%) & 27 (41\%) \\
                   & John W.\ Godward        &  2 ( 3\%) & 15 (23\%) \\
                   & Hokusai                 &  0 ( 0\%) & 16 (24\%) \\
\midrule
3 (misses)         & Vincent van Gogh        &  1 ( 2\%) & 15 (23\%) \\
                   & Diego Vel\'azquez       &  0 ( 0\%) &  3 ( 5\%) \\
                   & Claude Monet            &  0 ( 0\%) &  1 ( 2\%) \\
                   & Gustav Klimt            &  0 ( 0\%) &  0 ( 0\%) \\
                   & Frederic Leighton       &  0 ( 0\%) &  0 ( 0\%) \\
                   & Isaac Levitan           &  0 ( 0\%) &  0 ( 0\%) \\
\bottomrule
\end{tabular}
\caption{Bare FLUX.1-dev recognition test. For each prompted artist, the
number of generations whose top-1 (and top-5) CSD match in the 89-class
strict-clean corpus is the prompted artist. Embedding pipeline:
pos-interp $336$ (the input-side variant of \S\ref{sec:density}; the
default readout there is CSLS on vanilla $224$).
Tier-1 (trigger works, top-1 far above the rest of the panel and the top-5 mass on the artist),
Tier-2 (tradition reached: top-1 low but the top-5 mass sits in the
artist's neighbourhood), Tier-3 (the name token misses). The tiers are
qualitative descriptions and not an operational threshold. Hokusai,
Godward and van~Gogh sit near the Tier-2/Tier-3 boundary.}
\label{tab:bare-trigger}
\end{table}

\paragraph{Full LoRA stress-test table.}
Tab.~\ref{tab:lora-stress-full} expands the compact
Tab.~\ref{tab:lora-stress} of \S\ref{sec:generated} with all
$(\text{pair}, \text{condition}, \text{artist})$ cells across the three
negative-gap pairs and two positive-control artists, and reports
Bidir(cos), Bidir(CSLS) and Top-1 across all $89$ classes for every
cell. The positive members of the negative-gap pairs (Sharaku, Godward,
Shishkin) are included here for completeness. In the main-text table
they are dropped, because their raw-cosine bidir sits at or near $100\%$ in every
condition.

\begin{table}[h]
\centering\footnotesize
\setlength{\tabcolsep}{4.5pt}
\begin{tabular}{ll rrrr rrrr}
\toprule
& & \multicolumn{4}{c}{bare FLUX.1-dev} & \multicolumn{4}{c}{LoRA} \\
\cmidrule(lr){3-6}\cmidrule(lr){7-10}
Pair (gap) & Artist & $N$ & cos & CSLS & top-1 & $N$ & cos & CSLS & top-1 \\
\midrule
Edo ($-0.085$)       & Tōshūsai Sharaku  & 66 & 66 & 41 & 17 & 69 & 69 & 69 & 67 \\
                     & Kawanabe Kyōsai   & 66 &  0 & 28 &  0 & 60 &  2 & 44 &  0 \\
Victorian ($-0.020$) & Frederic Leighton & 66 &  3 &  9 &  0 & 57 &  0 &  2 &  0 \\
                     & John W.\ Godward  & 66 & 63 & 58 &  2 & 75 & 75 & 75 & 72 \\
Russian ($-0.030$)   & Isaac Levitan     & 66 &  0 &  0 &  0 & 60 &  9 & 18 &  0 \\
                     & Ivan Shishkin     & 66 & 66 & 66 &  5 & 63 & 63 & 63 & 43 \\
\midrule
Tier-3 control       & Claude Monet      & 66 & -- & -- &  0 & 48 & -- & -- & 13 \\
Tier-2 control       & Hokusai           & 66 & -- & -- &  0 & 39 & -- & -- &  3 \\
\bottomrule
\end{tabular}
\caption{Full LoRA stress test. Entries are counts of generations, out of
the $N$ of that condition, passing each readout: bidirectional pair
discrimination under raw cosine and CSLS (\S\ref{sec:generated}) and top-1
match in the 89-class corpus, all on pos-interp $336$ embeddings of
generations and anchors. Bare-condition CSLS values that depart from
cosine (Kyōsai $28/66$ up; Sharaku $41/66$, Godward $58/66$ down) are an
artefact of CSLS on the bare-prompt distribution, where generations land in an
artist's neighbourhood for unrelated reasons such as subject overlap.
They are not a recovery of the negative gap on T2I generations. The
LoRA-CSLS column is the operational reading.}
\label{tab:lora-stress-full}
\end{table}

\begin{figure}[t]
\centering
\begin{minipage}[t]{0.27\linewidth}
\centering
\includegraphics[width=\linewidth]{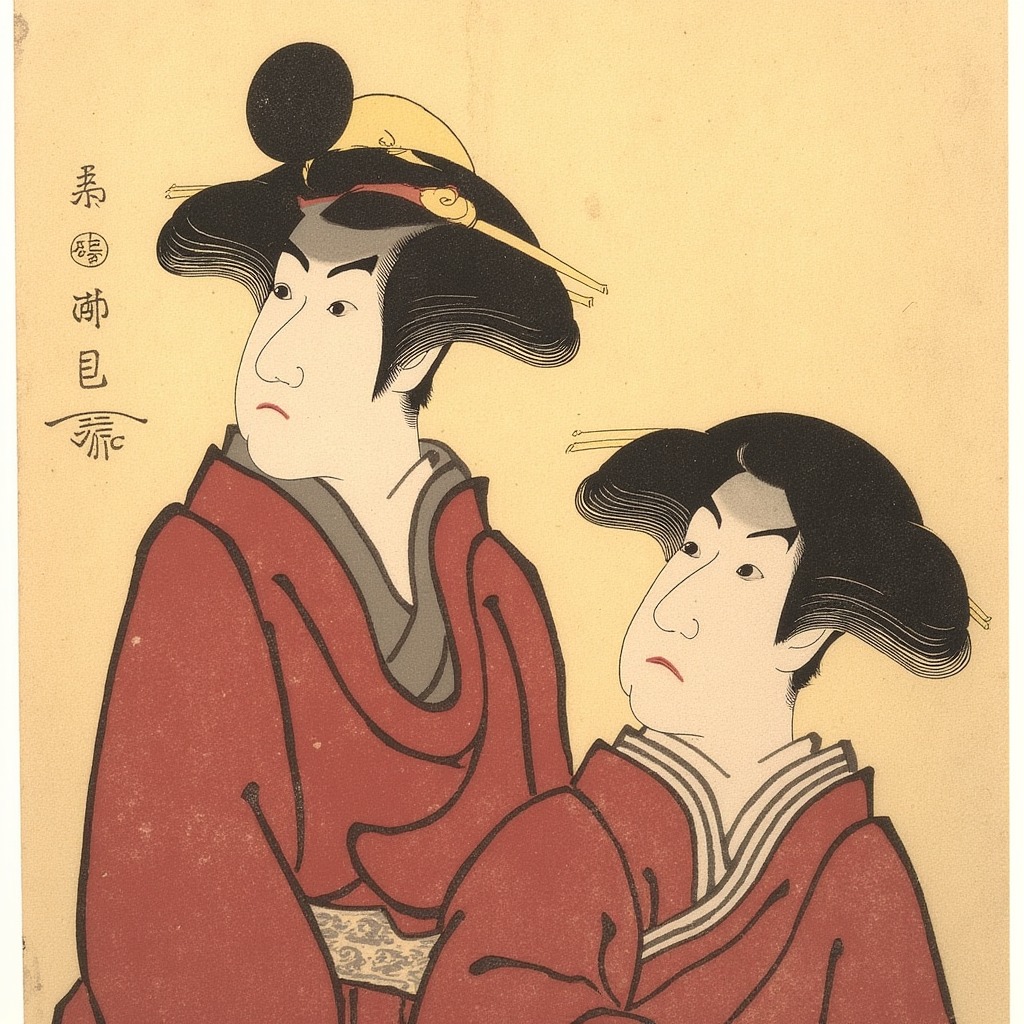}
\\\scriptsize (a) Flux-LoRA, Tōshūsai Sharaku style
\end{minipage}\hspace{0.02\linewidth}
\begin{minipage}[t]{0.27\linewidth}
\centering
\includegraphics[width=\linewidth]{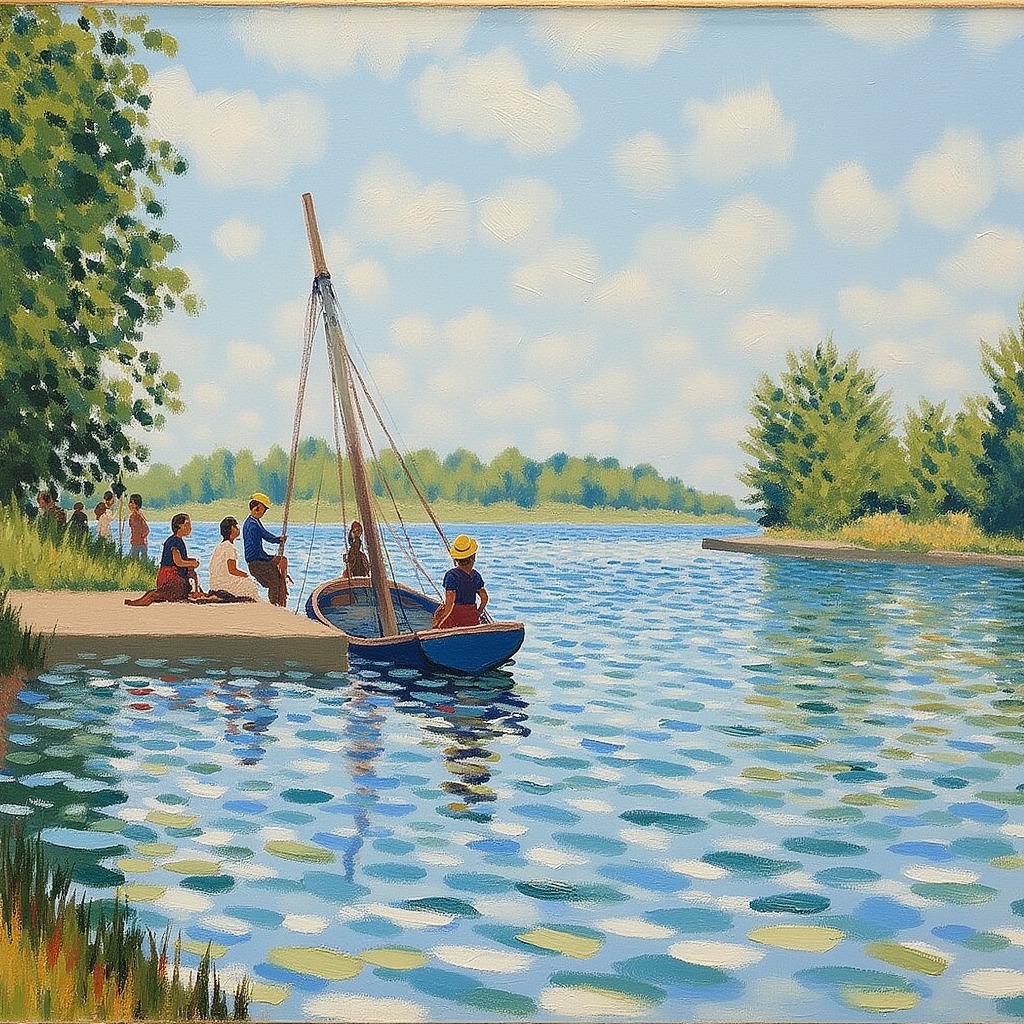}
\\\scriptsize (b) Flux-LoRA, C.~Monet style
\end{minipage}

\vspace{0.5em}

\begin{minipage}[t]{0.27\linewidth}
\centering
\includegraphics[width=\linewidth]{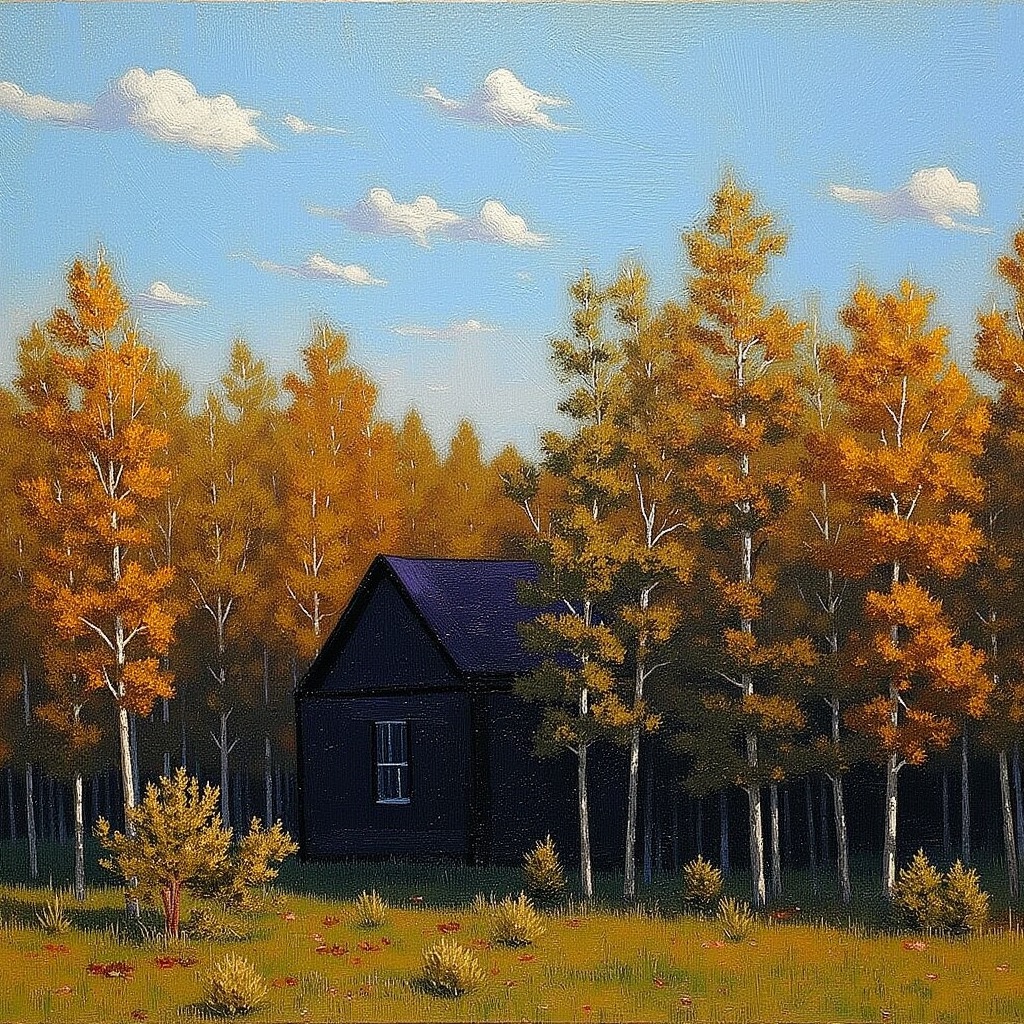}
\\\scriptsize (c) Flux-LoRA, I.~Levitan style
\end{minipage}\hspace{0.02\linewidth}
\begin{minipage}[t]{0.27\linewidth}
\centering
\includegraphics[width=\linewidth]{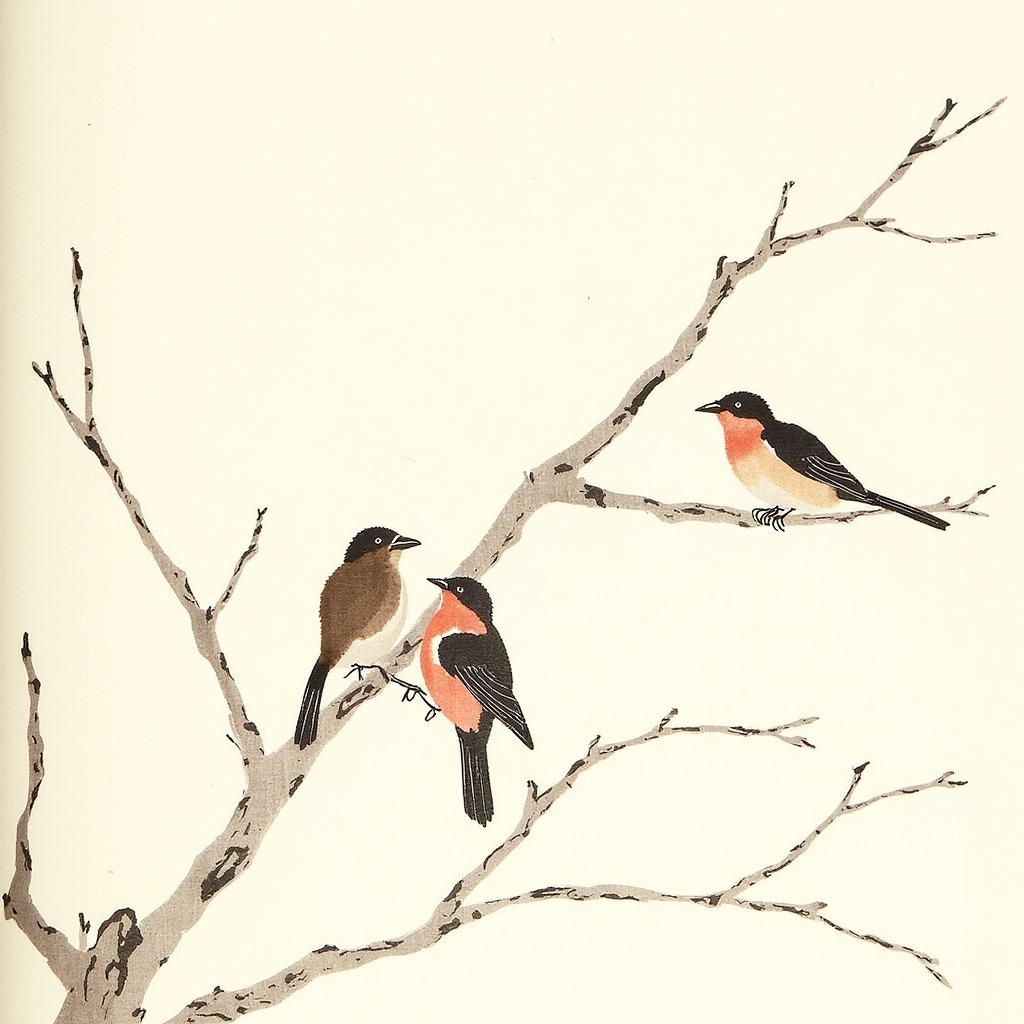}
\\\scriptsize (d) Flux-LoRA, K.~Kyōsai style
\end{minipage}
\caption{LoRA-fidelity exemplars: four Flux-LoRA generations spanning
Edo woodblock print, French Impressionism, Russian Realist landscape
and Japanese \emph{kach\=o-ga}, prompt-conditioned on rendered captions
of authentic strict-clean anchors. None of these images is a work of
the named artist; each is a generation by FLUX.1-dev with a per-artist
LoRA adapter. The text discusses what these galleries can and cannot
show.}
\label{fig:lora-fidelity}
\end{figure}

\paragraph{LoRA fidelity exemplars.}
The negative top-1 and bidir numbers in the main table can mislead
the reader into taking the trained LoRAs themselves for poor style
imitators. They are not. Fig.~\ref{fig:lora-fidelity} shows four LoRA outputs
spanning four traditions (Edo woodblock print, French Impressionism,
Russian Realist landscape, Japanese \emph{kach\=o-ga}). We
deliberately do \emph{not} place authentic anchors next to each
generation. The anchors are public-domain scans with age-related
patina (yellowed paper, darkened varnish, faded pigment), which the
LoRA does not attempt to reproduce. The side-by-side comparison
would therefore unfairly penalise the generations on a dimension
unrelated to style. Seen as a qualitative gallery, the
generations appear to us to carry style-defining properties of the
trained artist (palette, composition, brushwork, line economy). We ran
no blinded preference study, so this is an illustration and not a
measured finding. The bidir column of main-text
Tab.~\ref{tab:lora-stress} therefore measures corpus-internal
discriminability under raw cosine, which is the phenomenon
\S\S\ref{sec:diagnostic}/\ref{sec:generated} are about, and not training
quality.

\paragraph{Caption-template robustness pretest (Sharaku/Kyōsai).}
Before the main LoRA stress test we verified that the
asymmetry in pair discrimination between Sharaku and Ky\=osai
(Tab.~\ref{tab:lora-stress} in \S\ref{sec:generated}) is not an artefact of the caption template.
On the same trained adapters we generated $30$ images per artist with
two templates: the full template
\texttt{"late-Edo Japanese ukiyo-e woodblock print, <trigger>, <subject>"}
and a stripped variant \texttt{"woodblock print, <trigger>, <subject>"}.
The full template raises absolute self-cosines (Sharaku $0.308 \to 0.503$,
Kyōsai $0.225 \to 0.370$), but bidir is
identical, with Sharaku at $100\%$ and Kyōsai at $0\%$ in both. The
asymmetry is therefore a property of the CSD geometry and not of the
inference prompt.

\section{Reproducibility}
\label{app:reproducibility}

\paragraph{Software and precision.}
The CSD ViT-L/14 checkpoint is the community safetensors mirror of the
weights released by \citet{somepalli2024csd} (HuggingFace
\texttt{yuxi-liu-wired/CSD}, revision \texttt{cd2dac3a}, SHA-256 of
\texttt{model.safetensors} beginning \texttt{02d842ed2536e1f8}), used
without modification at the standard $224 \times 224$ input. The official
repository notes that its uploaded weights may differ from the paper's
reported results. Pos-interp
inference uses bilinear interpolation of the trained positional embeddings
to a $24 \times 24$ grid for $336 \times 336$ input. DINOv2-Large
\citep{oquab2024dinov2} is used in its public pretrained configuration.
All embedding computations are in fp32 and deterministic given the seeds
below. Clustering uses UMAP, HDBSCAN, scikit-learn and \texttt{leidenalg},
and the full pipeline runs on a single GPU.

\paragraph{Random seeds.}
The 25 verification splits are drawn with the fixed seeds
$1000 + i$, $i \in \{0, 1, \dots, 24\}$, and split $i$ seeds its pair
sampling from $i$. UMAP and Leiden experiments report
mean $\pm$ standard deviation across seeds $\{0, \dots, 9\}$ where
stated. Every result reported as ``$25/25$'' refers to a paired
comparison across the 25 verification splits. Since the splits overlap
in their artists, such a count says that the ordering is stable under
resampling of the artist split, and nothing stronger
(Tab.~\ref{tab:uncertainty-designs}).

\paragraph{LoRA stress test: what is recorded.}
Per-artist metadata is logged with each adapter and is part of the supplement:
training-image count ($52$ to $120$ across the eight artists), pool
composition, the artwork-kind distribution of the training pool, the
pseudo-token trigger, the caption schema
\texttt{\{movement\} \{artwork\_kind\}, \{trigger\}, \{caption\}} used
identically for all artists, the dropout caption, and the number of
evaluation anchors held out from training ($13$ to $25$). Training adds a
tradition-level regularisation pool in the DreamBooth manner. It consists of
$20$ to $65$ works by other artists of the same tradition, never by either
member of the evaluated pair. They are captioned with the class phrase
but without the trigger and sampled uniformly together with the
artist's own images. Pool directories and sizes are in the run files. Generations
use as prompts the rendered captions of the artist's held-out anchors (one
Monet anchor, a detail crop, was left out), at \texttt{lora\_scale} $1.4$,
$28$ sampling steps, guidance $3.5$ and three fixed seeds each. The count is therefore the number of prompts
times three, which is why the denominators differ between artists. The bare
condition uses one shared list of $22$ subjects, hence a uniform $66$. No
generation was excluded at evaluation time, and the number of files on disk
equals the number scored in every cell.

The optimiser configuration is recorded per adapter in a run file written at
training time, and the training script fixes the rest: rank $32$, alpha $32$,
AdamW at learning rate $10^{-4}$ with $50$ linear warm-up steps and a constant
rate thereafter, $3000$ steps at batch size $1$ without gradient
accumulation, $1024$\,px resolution, caption dropout $0.05$, bf16. The adapter
targets the attention projections of the FLUX transformer. The text encoders
stay frozen, and prompts are encoded once before training. The evaluated
adapter is the final checkpoint; intermediate checkpoints were saved but not
used for selection. Training script and run files ship with the supplement.
Each adapter was trained once. A retrain therefore reproduces the protocol
but, in bf16, not necessarily identical weights. With eight
artists, the LoRA test of \S\ref{sec:generated} remains a case series.

\paragraph{Pair-feature logistic regression.}
Each split uses $4000$ same-artist and $4000$ different-artist training pairs
(balanced by construction) and $1500$ of each for evaluation. Positives cycle
the artist pool taking two distinct works, while negatives draw two distinct
artists and one work each. Seeds are fixed per split (split index for
training, plus $17$ for evaluation). The features are
$[\,|a-b| \,\Vert\, a \odot b \,\Vert\, (a+b)/2\,]$ in
$\mathbb{R}^{2304}$, unscaled. The classifier is scikit-learn's
\texttt{LogisticRegression} with \texttt{C=1.0}, \texttt{max\_iter=2000}
and otherwise library defaults. No hyperparameter search is performed
anywhere, so the same fixed $C$ applies on every split, corpus and
backbone. The learned Mahalanobis comparison of \S\ref{sec:verification}
is likewise untuned. $L$ is initialised at the identity, and we run $200$
full-batch Adam steps at learning rate $0.05$ on the binary cross-entropy of
$\mathrm{bias} - (a-b)^\top L L^\top (a-b)$.

\paragraph{Technique-permutation test.}
The per-artist technique test of \S\ref{sec:tradition} compares each artist's
same-technique within-median against a null in which that artist's technique
labels are permuted at fixed group sizes, so the group-size profile is held and
only the assignment moves. By default we use $1000$ permutations, seed $0$ and
a one-sided test against larger observed shifts, with the add-one correction
$p = (1 + \#\{\text{null} \ge \text{observed}\}) / (1 + B)$. The smallest
reportable value is therefore $1/1001$, and entries shown as $p < 0.001$ are at
that floor. Artists recorded in a single technique have no permutable labels
and are excluded from the test instead of being counted as non-significant.
Groups of size one contribute no same-technique pairs and drop out of the
within-median by construction. Ties in the shift are counted towards the null,
which is the conservative direction. The false-discovery-rate correction is
Benjamini--Hochberg at $q < 0.05$ over the $89$ artists of Commons-89.
On the external corpora $200$ permutations bound the smallest attainable
$p$-value at $1/201$. Their per-artist counts
(Tab.~\ref{tab:category-replication}) are therefore screening counts at
unadjusted $p < 0.05$.

\paragraph{Strict-clean filter.}
The strict-clean anchor set used throughout this paper is the output
of a single-source-of-truth filter applied to the post-attribution
pool. An image is admitted iff its
\texttt{is\_artwork\_by\_artist} channel is true, its
\texttt{artwork\_kind} is in
$\{$\texttt{painting}, \texttt{drawing}, \texttt{print},
\texttt{watercolor}, \texttt{mixed\_media}$\}$,
both \texttt{has\_frame} and \texttt{has\_museum\_context} are
false, and its caption does not match a curated set of
suspect-caption regular expressions targeting reproductions,
postcards, plaques and book covers. Artists with fewer than ten
admissible anchors are dropped from the working corpus. The same
filter applies uniformly to all backbones and all input pipelines
reported in the paper, so cross-backbone and cross-input numbers
reference the same $1645$-image, $89$-artist set. The resulting
embedding NPZ files therefore differ only in the embedding
function applied to identical input pixels.

\paragraph{External corpora.}
The external replication of App.~\ref{app:external} uses the public
parquet distributions of the WikiArt dump and of ArtBench-10. The WikiArt
dump has $81{,}444$ images, of which $39{,}530$ carry one of the $128$ named
artist labels. For ArtBench the train and test splits are pooled,
$60{,}000$ images in total. Images are embedded in batches straight from the
parquet shards. Bit-exact agreement with the single-image reference is
therefore not expected, because batched inference changes the fp32
reduction order. Per-artist gaps, worst-other identities, cross-median
matrices and bootstrap distributions for both corpora are stored alongside
the main-corpus results.

\paragraph{Per-image attribution side-channel.}
The audit records key each anchor's attribution label by filename, and
the attribution sensitivity of \S\ref{sec:diagnostic} is computed through
this join.

\end{document}